\documentclass[twoside,11pt]{article}
\usepackage{jair, theapa, rawfonts}

\usepackage{amsthm}
\usepackage{amsmath}
\usepackage{amsfonts}
\usepackage{amssymb}
\usepackage{mathtools}
\usepackage{units}
\usepackage{graphicx}
\usepackage{color}
\usepackage{subfig} 
\usepackage{url}
\usepackage{booktabs}
\usepackage{multirow}
\usepackage{dcolumn}
\usepackage{bbm}
\usepackage{pifont}

\newcommand{\cmark}{\text{\ding{51}}}%
\newcommand{\xmark}{\text{\ding{55}}}%

\newcommand{\eg}{e.\,g., }
\newcommand{\ie}{i.\,e., }

\newcolumntype{d}[1]{D{.}{.}{#1} }

\newtheorem{definition}{Definition}

\begin{document}

\title{Graph Kernels: A Survey}

\author{\name Giannis Nikolentzos \email nikolentzos@lix.polytechnique.fr \\
		\addr LIX, \'Ecole Polytechnique\\
       	Palaiseau, 91120, France\\ \\
       \name Ioannis Siglidis \email ioannis.siglidis@enpc.fr \\
       \addr LIGM, \'Ecole des Ponts, Universit\'e Gustave Eiffel, CNRS \\
       Marne-la-Vall\'ee, 77420, France\\ \\
       \name Michalis Vazirgiannis \email mvazirg@lix.polytechnique.fr \\
       \addr LIX, \'Ecole Polytechnique\\
       Palaiseau, 91120, France}


\maketitle

\begin{abstract}
Graph kernels have attracted a lot of attention during the last decade, and have evolved into a rapidly developing branch of learning on structured data.
During the past $20$ years, the considerable research activity that occurred in the field resulted in the development of dozens of graph kernels, each focusing on specific structural properties of graphs.
Graph kernels have proven successful in a wide range of domains, ranging from social networks to bioinformatics.
The goal of this survey is to provide a unifying view of the literature on graph kernels.
In particular, we present a comprehensive overview of a wide range of graph kernels.
Furthermore, we perform an experimental evaluation of several of those kernels on publicly available datasets, and provide a comparative study.
Finally, we discuss key applications of graph kernels, and outline some challenges that remain to be addressed. 
\end{abstract}

\section{Introduction}\label{Introduction}
In recent years, the amount of data that can be naturally modeled as graphs has increased significantly.
Such types of data have become ubiquitous in many application domains, ranging from social networks to biology and chemistry.
A large portion of the available graph representations corresponds to data derived from social networks.
These networks represent the interactions between a set of individuals such as friendships in a social website or collaborations in a network of film actors or scientists.
In chemistry, molecular compounds are traditionally modeled as graphs where vertices represent atoms and edges represent chemical bonds.
Biology constitutes another primary source of graph-structured data. Protein-protein interaction networks, metabolic networks, regulatory networks, and phylogenetic networks are all examples of graphs that arise in this domain.
Graphs are also well-suited to representing technological networks.
For example, the World Wide Web can be modeled as a graph where vertices correspond to webpages and edges to hyperlinks between these webpages.
The use of graph representations is not limited to the above application domains.
In fact, most complex systems are usually represented as compositions of entities along with their interactions, and can thus be modeled as graphs.
Interestingly, graphs are very flexible and rich as a means of data representation.
It is not thus surprising that they can also represent data that do not inherently possess an underlying graph structure.
For instance, sequential data such as text can be mapped to graph structures \shortcite{filippova2010multi}.
From the above, it becomes clear that graphs emerge in many real-world applications, and hence, they deserve no less attention than feature vectors which is the dominant representation in data mining and machine learning.

The aforementioned abundance of graph-structured data raised requirements for automated methods that can gain useful insights.
This often requires applying machine learning techniques to graphs.
In chemistry and biology, some experimental methods are very expensive and time-consuming, and machine learning methods can serve as cost-effective alternatives.
For example, identifying experimentally the function of a protein with known sequence and structure is a very expensive and tedious process.
Therefore, it is often desirable to be able to use computational approaches in order to predict the function of a protein.
By representing proteins as graphs, the problem can be formulated as a graph classification problem where the function of a newly discovered protein is predicted based on structural similarity to proteins with known function \shortcite{borgwardt2005protein}.
Besides the need for more efficient methods, there is also a need for automating tasks that were traditionally handled by humans and which involve large amounts of data.
For instance, in cybersecurity, humans used to manually inspect code samples to identify if they contain malicious functionality.
However, due to the rapid increase in the number of malicious applications in the past years, humans are no longer capable of meeting the demands of this task \shortcite{suarez2014evolution}.
Hence, there is a need for methods that can accumulate human knowledge and experience, and that can successfully detect malicious behavior in code samples.
It turns out that machine learning approaches are particularly suited to this task since most of the newly discovered malware samples are variations of existing malware.
By representing code samples as function call graphs, detecting such variations becomes less problematic.
Hence, the problem of detecting malicious software can be formulated as a graph classification problem where unknown code samples are compared against known malware samples and clean code \shortcite{anderson2011graph}.
From the above example, it becomes clear that performing machine learning tasks on graph-structured data is of critical importance for many real-world applications. 

A central issue for machine learning is modelling and computation of similarity among objects.
In the case of graphs, graph kernels have received a lot of attention in the past years, and have been established as one of the major approaches for learning on graph-structured data.
A graph kernel is a symmetric, positive semidefinite function defined on the space of graphs $\mathcal{G}$.
This function can be expressed as an inner product in some Hilbert space.
Specifically, given a kernel $k$, there exists a map $\phi : \mathcal{G} \rightarrow \mathcal{H}$ into a Hilbert space $\mathcal{H}$ such that $k(G_1,G_2) = \langle \phi(G_1), \phi(G_2) \rangle$ for all $G_1,G_2 \in \mathcal{G}$.
Roughly speaking, a graph kernel is a measure of similarity between graphs.
Graph comparison is a fundamental problem with numerous applications in many disciplines \shortcite{conte2004thirty}.
However, the problem is far from trivial and requires considerable computational time.
Graph kernels tackle this problem by trying to both capture as much as possible the semantics inherent in the graph but also to remain computationally efficient.
One of the most important reasons behind the success of graph kernels is that they allow the large family of kernel methods to work directly on graphs.
Therefore, graph kernels can bring to bear several machine learning algorithms to real-world problems on graph-structured data.
The field of graph kernels has been intensively developed recently.
Interestingly, dozens of graph kernels have been proposed in the past $20$ years.
Some of these kernels have achieved state-of-the-art results on several datasets.
Recently, there has been a significant surge of interest in Graph Neural Network (GNN) approaches for graph representation learning.
Most of these models follow a neighborhood aggregation scheme similar to that of many graph kernels, and can be reformulated into a single common framework \shortcite{gilmer2017neural}.
The main advantage of GNNs over graph kernels is that their complexity is linear to the number of samples, while kernels require quadratic time to compute all kernel values.
For a detailed presentation of this important emerging field, the interested reader is referred to \shortciteA{wu2020comprehensive}.

This paper is a survey of graph kernels, that is kernels that operate on graph-structured data.
We present a comprehensive study of these approaches.
We begin with well-known kernels that established the foundations of the field, and we proceed with more recent kernels that are considered the state-of-the-art for many graph-related machine learning tasks.
Besides the detailed description of the kernels, we also provide an extensive experimental evaluation of most of them.
As we show in this survey, graph kernels are powerful tools with a wide range of applications, while their empirical performance is superior to that of graph neural networks for certain types of graphs.
We thus expect these methods to gain soon more attention in a wealth of applications due to their attractive properties.
Importantly, this study aims to assist both practitioners and researchers who are interested in applying machine learning tasks on graphs.
Furthermore, it should be of interest to all researchers who deal with the problems of graph similarity and graph comparison.
The abundance of applications related to the above problems stresses the value of the survey.
We should note that three similar surveys reviewing work on graph kernels became very recently available \shortcite{ghosh2018journey,kriege2020survey,borgwardt2020graph}.
One may thus ask the question: why another survey within such a short period of time?
The answer is that in contrast to the first two above surveys, this survey is much more thorough and covers a larger number of kernels.
Moreover, it presents kernels in a more comprehensive way allowing researchers to identify open problems and areas for further exploration, and practitioners to gain a deeper understanding of kernels so that they can decide which kernel best suits their needs.
Specifically, the above two surveys do not go into sufficient details about the mathematical foundations of the different kernels.
On the other hand, we provide an in-depth discussion of a large number of kernels along with all the mathematical details that are of high importance in this domain.
This survey also provides a much more meaningful taxonomy of graph kernels.
More specifically, kernels are grouped into classes based on different criteria such as the type of data on which they operate, and the design paradigm that they follow.
The third survey \shortcite{borgwardt2020graph} is very detailed and well-written, and there is a considerable intersection with this survey, especially in terms of the articulation of the presentation of the kernels, however, it lags behind in terms of empirical analysis.
To the best of our knowledge, we provide the most complete evaluation in terms of the number of considered graph kernels.
\shortciteA{ghosh2018journey} do not perform original graph classification experiments, but they only report results from the kernels' original papers.
\shortciteA{kriege2020survey} perform original experiments, however, they only evaluate $9$ kernels (and their variants) and $1$ framework in total, while \shortciteA{borgwardt2020graph} evaluate $12$ kernels and $1$ framework.
On the other hand, our list of methods includes $16$ different kernels and $2$ frameworks.
Besides classification performance, we also measure and report running times (not provided by \shortciteR{kriege2020survey} or by \shortciteR{borgwardt2020graph}).
We believe that running times are one of the major reasons behind the choice of a kernel for a practical application.
Also, we need to stress that such a wider, and more extensive experimental comparison of graph kernels can provide useful insights into the strengths and weaknesses of the different kernels.
Furthermore, we compare graph kernels against graph neural networks which we believe that is an important piece of exploration as to the comparison of two worlds (neural networks and kernels) in the context of graphs.
Finally, we empirically compare the expressiveness of the kernels to each other, that is how well the different kernels capture the similarity of graphs, something that is missing from the current literature.

The rest of this manuscript is organized as follows.
In Section~\ref{sec:motivation_challenges}, we discuss why the use of graphs as a means of object representation is vital and necessary in many domain areas, and we also present the challenges of applying learning algorithms on graphs.
In Section~\ref{sec:preliminaries}, we introduce notation and background material that we need for the remainder of the paper, including some fundamental concepts from graph theory and from kernel methods.
In Section~\ref{sec:graph_kernels}, we discuss the core concepts of graph kernels, and we give an overview of the literature on graph kernels.
We begin by describing important kernels that were developed in the early days of the field.
We next present kernels that are based on neighborhood aggregation mechanisms.
We then describe more recent kernels that do not employ neighborhood aggregation mechanisms.
Subsequently, we present kernels that are based on assignment, and methods that can handle continuous node attributes.
Finally, we give details about frameworks that work on top of graph kernels and aim to improve their performance.
The grouping of the reported studies is designed to make it easier for the reader to follow the analysis of the literature, and to obtain a complete picture of the different graph kernels that have been proposed throughout the years.
In Section~\ref{sec:gnns}, we provide a short introduction to graph neural networks, the main competitors of graph kernels, and we discuss how the major family of these models is related to graph kernels.
In Section~\ref{sec:applications}, we present applications of graph kernels in many different domain areas.
In Section~\ref{sec:experiments}, we experimentally evaluate the performance of many graph kernels on several widely-used graph classification benchmark datasets.
Furthermore, we measure the running times of these kernels.
Based on the obtained results, we provide guidelines for the successful application of graph kernels in different classification problems.
We also study the expressive power of graph kernels from an empirical standpoint by comparing the obtained kernel values against the similarities that are produced by a well-accepted but intractable graph similarity function.
Finally, Section~\ref{sec:conclusion} contains the summary of the survey, along with a discussion about future research directions in the field of graph kernels.

\section{Motivation and Challenges}\label{sec:motivation_challenges}
In this Section, we present the main reasons that motivate the use of graphs instead of feature vectors as a means of data representation.
Furthermore, we describe the problem of learning on graphs which arises in many application domains.
We focus on the instance of the problem where each sample is a graph, and highlight its relationship to the graph comparison problem.

\subsection{Why Graphs}
Graphs are a powerful and flexible means of representing structured data.
The power of graphs stems from the fact that they represent both entities, and the relationships between them.
Typically, the vertices of a graph correspond to some entities, and the edges model how these entities interact with each other.
It is important to note that several fundamental structures for representing data can be seen as instances of graphs \shortcite{borgwardt2007graph}.
This highlights the generality of graphs as a form of representation.
For example, a vector can be naturally thought of as a graph where vertices correspond to components of the vector and consecutive components within the vector are joined by an edge.
Associative arrays can be modeled as graphs, with keys and values represented as vertices, and directed edges connecting keys to their corresponding values.
Strings can also be represented as graphs, with one vertex per character and edges between consecutive characters.
Due to the power and the generality of graphs as representational models, in some cases, even data that does not exhibit graph-like structure is mapped to graph representations.
A very common example is that of textual data, where graphs are usually employed to model the relationships between sentences or terms \shortcite{mihalcea2004textrank}.

In data mining and machine learning, observations traditionally come in the form of vectors.
However, vector representations suffer from a series of limitations.
Specifically, vectors have limited capability to model complex objects since they are unable to capture relationships that may exist between different entities of an object.
Furthermore, all the input objects are usually represented as vectors of the same length, despite their size and complexity.
On the other hand, as discussed above, graphs are characterized by increased flexibility which allows them to adequately model a variety of different objects.
Graphs model both the entities and the relationships between them.
Moreover, they are allowed to vary in the number of vertices and in the number of edges.
Therefore, graphs address several of the limitations inherent to vectors.
It is thus clear that the need for methods that perform learning tasks on graphs is intense.

\subsection{Learning on Graphs and Challenges}
Learning on graphs has gained extensive attention in the past years.
This is mainly due to the representational power of graphs which has established them as a major structure for modeling data from various disciplines.
Hence, it is not surprising that a plethora of learning problems have been defined on graphs. 
Most of these learning problems focus either on the node level or on the graph level.
Node classification belongs to the former set of problems, while graph classification belongs to the latter set of problems.
In this survey, we focus exclusively on tasks performed at the graph level.
Therefore, all the kernels that are presented correspond to functions between graphs.

Data representation is a key issue in the fields of data mining and machine learning.
Algorithms are mainly designed to handle data in a specific representation.
Due to the appealing properties of graphs, one would expect that there would be great progress in the development of algorithms that can handle graph-structured data.
However, the combinatorial nature of graphs acts as a ``barrier'' since it is very likely that algorithms that operate directly on graphs will be computationally expensive and will not scale to large datasets.
Thus, research in these areas has mainly focused on algorithms operating on vectors, as vectors possess many desirable mathematical properties and can be dealt with much more efficiently.
Hence, it is not surprising that the most popular learning algorithms are designed for data represented as vectors.
As a consequence, it has become common practice to represent any type of data as feature vectors.
Even in application domains where data is naturally represented as graphs, attempts were made to transform graphs into feature vectors instead of designing algorithms that operate directly on graphs.
Ideally, we would like to have a method that runs in polynomial time and is capable of transforming graphs to feature vectors without sacrificing their representational power.
Unfortunately, such a method does not exist.
Directly representing data as vectors is thus suboptimal since vectors fail to preserve the rich topological information encoded in a graph.
Hence, it would be much more preferable to devise algorithms that operate directly on graphs.

The problem of learning on graphs (at the graph level) is directly related to that of \textit{graph comparison}.
The ability to compute meaningful similarity or distance measures is often a prerequisite to perform machine learning tasks.
Such similarity and distance measures are at the core of many machine learning algorithms.
Examples include the $k$-nearest neighbor classifier, and algorithms that learn decision functions in proximity spaces \shortcite{graepel1999classification}.
These algorithms are very flexible since they require only a distance or similarity function to be defined as the sole mathematical structure on the set of input objects.
Hence, by defining a meaningful distance function $d: \mathcal{G} \times \mathcal{G} \rightarrow \mathbb{R}_+$ between graphs, we can immediately use one of the above algorithms to perform tasks such as graph classification and graph clustering.
However, it turns out that graph comparison is a very complex problem.
Specifically, graphs lack the convenient mathematical context of vector spaces, and many operations on them, though conceptually simple, are either not properly defined or computationally expensive.
Perhaps the most striking example of these operations is to determine if two objects are identical.
In the case of vectors, it requires comparing all their corresponding components, and it can thus be accomplished in linear time with respect to the size of the vectors.
For the analogous operation on graphs, known as \textit{graph isomorphism}, no polynomial-time algorithm has been discovered so far \shortcite{garey1979computers}.
In general, the problem of comparing two objects is much less well-defined on graphs compared to vectors.
For vectors, distance can be computed efficiently using the universally accepted Euclidean distance metric.
Unfortunately, there exists no such metric on graphs.
Several fundamental problems in graph theory related to graph comparison such as the subgraph isomorphism problem and the maximum common subgraph problem are NP-complete \shortcite{garey1979computers}.
Furthermore, identifying common parts in two graphs is computationally infeasible.
Given a graph consisting of $n$ vertices, there are $2^n$ possible subsets of vertices.
Hence, there are exponentially many (in the size of the graphs) pairs of subsets to consider.
It becomes thus clear that although graphs offer a very intuitive way of modeling data from diverse sources, their power and flexibility do not come without a price.

\section{Preliminaries}\label{sec:preliminaries}
Before we delve into the details of graph kernels, we outline some fundamental aspects of graph theory and kernel methods. 
We first introduce basic concepts from graph theory, and define our notation.
We also provide a short introduction to kernel functions and kernel methods in machine learning.

\subsection{Definitions and Notations}
\begin{figure}[t]
    \centering
    \subfloat
    {\includegraphics[width=.25\linewidth]{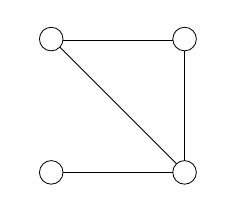}} \qquad \qquad
    \subfloat
    {\includegraphics[width=.25\linewidth]{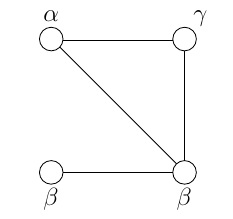}} \qquad \qquad
    \subfloat
    {\includegraphics[width=.25\linewidth]{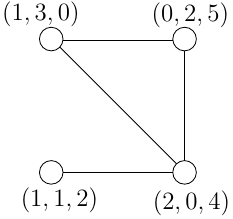}} 
    \caption{Examples of different types of graphs. A simple undirected graph (left), a labeled graph (center), and an attributed graph (right).}
    \label{fig:example_graphs}
\end{figure}

\begin{definition}[Graph]
    A graph is a pair $G=(V,E)$ consisting of a set of vertices (or nodes) $V$ and a set of edges $E \subseteq V \times V$ which connect pairs of vertices. 
\end{definition}
The size of the graph corresponds to its number of vertices denoted by $|V|$ or $n$.
As regards the number of edges of the graph, we will denote it as $|E|$ or $m$.
An example of a graph is given in Figure~\ref{fig:example_graphs} (left).
A graph may have labels on its nodes and edges.
This is often necessary for capturing the semantics of complex objects.
For instance, most graphs derived from chemistry (\eg molecules) are annotated with categorical labels from a finite set.
\begin{definition}[Labeled Graph]
    A labeled graph is a graph $G=(V,E)$ endowed with a function $\ell : V \cup E \rightarrow \Sigma$ that assigns labels to the vertices and edges of the graph from a discrete set of labels $\Sigma$.
\end{definition}
A graph with labels on its vertices is called node-labeled.
Similarly, a graph with labels on edges is called edge-labeled. 
A graph with labels on both the vertices and edges is called fully-labeled.
An example of a node-labeled graph is given in Figure~\ref{fig:example_graphs} (center).
In many settings, vertex and edge annotations are in the form of vectors.
For example, vertices and edges may be annotated with multiple categorical or real-valued properties.
These graphs are known as attributed graphs.
\begin{definition}[Attributed Graph]
    An attributed graph is a graph $G=(V,E)$ endowed with a function $f : V \cup E \rightarrow \mathbb{R}^d$ that assigns real-valued vectors to the vertices and edges of the graph.
\end{definition}
An example of a node-attributed graph is given in Figure~\ref{fig:example_graphs} (right).
Note that labeled graphs are a special case of attributed graphs.
We can represent labeled graphs as attributed graphs if we map the discrete labels to one-hot vector representations.
A graph $G=(V,E)$ can be represented by its adjacency matrix $A$.
\begin{definition}[Adjacency Matrix]
    Let $A_{ij}$ be the element in the $i$-th row and $j$-th column of matrix $A$.
    Then, the adjacency matrix $A$ of a graph $G=(V,E)$ can be defined as follows
    \[A_{ij} = \left\{
      \begin{array}{lr}
        1 & \text{if }(v_i,v_j) \in E,\\
        0 & \text{otherwise}
      \end{array}
    \right.
    \]
\end{definition}
The adjacency matrix $A$ consists of $n$ rows and $n$ columns, that is $A \in \mathbb{R}^{n \times n}$.
The neighborhood $\mathcal{N}(v_i)$ of vertex $v_i$ is the set of all vertices adjacent to $v_i$.
Hence, $\mathcal{N}(v_i) = \{v_j : (v_i,v_j) \in E\}$ where $(v_i, v_j)$ is an edge between vertices $v_i$ and $v_j$ of $V$.
A concept closely related to the neighborhood of a vertex $v_i$ is its degree $deg_G(v_i)$.
\begin{definition}[Degree]
    Given an undirected graph $G=(V,E)$ and a vertex $v_i \in V$, the degree of $v_i$ is the number of edges incident to $v_i$, and is defined as
    \begin{equation}
        deg(v_i) = | \{v_j : (v_i,v_j) \in E \} | = |\mathcal{N}(v_i)|
    \end{equation}
\end{definition}
The maximum of the degrees of the vertices of a graph is denoted by $deg^*$, and $deg^* = \max_{v \in V} deg(v)$.
Besides the adjacency matrix $A$, a graph $G=(V,E)$ can also be represented by its Laplacian matrix $L$.
\begin{definition}[Laplacian Matrix]
    Let $A$ be the adjacency matrix of a graph $G=(V,E)$ and $D$ a diagonal matrix with $D_{ii} = \sum_j A_{ij}$.
    Then, the Laplacian matrix $L$ of a graph $G=(V,E)$ can be defined as follows
    \begin{equation}
      L = D - A
    \end{equation}
\end{definition}
Similarly to the adjacency matrix $A$, the dimensionality of the Laplacian matrix is $n \times n$.
A subgraph of a graph $G$ is a graph whose set of vertices and set of edges are both subsets of those of $G$.
Let $G' \subseteq G$ denote that $G'$ is a subgraph of $G$.
\begin{definition}[Induced Subgraph]
    Given a graph $G=(V,E)$ and a subset of vertices $S \subseteq V$, the subgraph $G(S) = (S, E(S))$ induced by $S$ consists of the set of vertices $S$ and the set of edges $E(S)$ that have both end-points in $S$ defined as follows
    \begin{equation}
        E(S) = \{ (v_i, v_j) \in E : v_i,v_j \in S \}
    \end{equation}
\end{definition}
The degree of a vertex $v_i \in S$, $deg_{G(S)}(v_i)$, is equal to the number of vertices that are adjacent to $v_i$ in $G(S)$.
The \textit{density} of a graph $G$ is $\delta(G) = m/\binom{n}{2}$, the number of edges $m$ over the total possible edges.
A graph $G$ with density $\delta(G) = 1$ is called a \textit{complete} graph.
In a complete graph, every pair of distinct vertices are adjacent.
A \textit{clique} is a subset of vertices such that every pair of them are connected by an edge, that is, their induced subgraph is complete.
\begin{definition}[Walk, Path, Cycle]
    A walk in a graph $G=(V,E)$ is a sequence of vertices $v_1,v_2,\ldots,v_{k+1}$ where $v_i \in V$ for all $1 \le i \le k+1$ and $(v_i, v_{i+1}) \in E$ for all $1 \le i \le k$.
    The length of the walk is equal to the number of edges in the sequence, that is $k$ in the above case.
    A walk in which $v_i \ne v_j \Leftrightarrow i \ne j$ is called a path.
    A cycle is a path with $(v_{k+1}, v_1) \in E$.
\end{definition}
\begin{definition}[Shortest Path]
    A shortest path from vertex $v_i$ to vertex $v_j$ of a graph $G$ is a path from $v_i$ to $v_j$ such that there exist no other path between these two vertices with smaller length.
\end{definition}
The \textit{diameter} of a graph $G$ is the length of the longest shortest path between any pair of vertices of $G$.
The \textit{neighborhood of radius} $r$ (or $r$-hop neighborhood) of vertex $v_i$ is the set of vertices whose shortest path distance from $v_i$ is less than or equal to $r$ and is denoted
by $\mathcal{N}_r(v_i)$.
Table~\ref{tab:symbols} gives a list of the most commonly used symbols along with their definition.
\begin{table}[t]
  \centering
  \footnotesize
  \def\arraystretch{1.1}
  \begin{tabular}{llll} \hline
    \multicolumn{4}{c}{\textbf{List of key symbols}} \\ \hline
    $\mathcal{G}$ & Set of graphs & $G$ & A graph \\
    $V$ & Set of vertices & $E$ & Set of edges \\
    $n$ & Number of vertices & $m$ & Number of edges \\
    $\mathcal{N}(v)$ & Neighbors of $v$ & $deg(v)$ & Degree of vertec $v$ \\
    $G(S)$ & Subgraph of $G$ induced by set of vertices $S$ & $deg^*$ & Maximum degree \\
    $A$ & Adjacency matrix of graph & $L$ & Laplacian matrix of graph \\
    $\ell$ & Function that assigns labels to vertices and edges & $\delta$ & Diameter of graph \\ 
    $f$ & Function that assigns attributes to vertices and edges & $\mathcal{N}_r(v)$ & $r$-hop neighborhood of $v$ \\ \hline
  \end{tabular}
  \caption{Commonly used symbols and notations}
  \label{tab:symbols}
\end{table}

\subsection{Kernel Functions and Kernel Methods}
We next give an introduction to kernel functions and kernel methods.
\begin{definition}[Gram Matrix]
  Given a set of inputs $x_1,\ldots,x_N \in \mathcal{X}$ and a function $k : \mathcal{X} \times \mathcal{X} \rightarrow \mathbb{R}$, the $N \times N$ matrix $K$ defined as
  \begin{equation}
    K_{ij} = k(x_i, x_j)
  \end{equation}
  is called the gram matrix (or kernel matrix) of $k$ with respect to the inputs $x_1,\ldots,x_N$.
\end{definition} 
In what follows, we will refer to gram matrices as kernel matrices.
\begin{definition}[Positive Semidefinite Matrix]
  A real $N \times N$ symmetric matrix $K$ satisfying
  \begin{equation}
    \sum_{i=1}^N \sum_{j=1}^N c_i c_j K_{ij} \geq 0
  \end{equation}
  for all $c_i \in \mathbb{R}$ is called positive semidefinite.
\end{definition}
\begin{definition}[Positive Semidefinite Kernel]
  Let $\mathcal{X}$ be a nonempty set.
  A function $k : \mathcal{X} \times \mathcal{X} \rightarrow \mathbb{R}$ which for all $N \in \mathbb{N}$ and all $x_1,\ldots,x_N \in \mathcal{X}$ gives rise to a positive semidefinite kernel matrix is called a positive semidefinite kernel, or just a kernel.
\end{definition} 
Informally, a kernel function 
measures the similarity between two objects. 
Furthermore, kernel functions can be represented as inner products between the vector representations of these objects. 
Specifically, if we define a kernel $k$ on $\mathcal{X} \times \mathcal{X}$, then there exists a mapping $\phi : \mathcal{X} \rightarrow \mathcal{H}$ into a Hilbert space with inner product $\langle \cdot, \cdot \rangle$, such that:
\begin{equation}
  \forall x_i,x_j \in \mathcal{X} : k(x_i, x_j) = \langle \phi(x_i), \phi(x_j) \rangle
\end{equation}
A Hilbert space is an inner product space which also possesses the completeness property that every Cauchy sequence of points taken from the space converges to a point in the space.
Furthermore, the Hilbert space $\mathcal{H}$ has the following property known as the reproducing property:
\begin{equation}
  \forall f \in \mathcal{H}, \forall x \in \mathcal{X} : f(x) = \langle f, k(x,\cdot) \rangle
\end{equation}
By virtue of this property, $\mathcal{H}$ is called a reproducing kernel Hilbert space (RKHS) associated with kernel $k$.
It is interesting to note that every kernel function on $\mathcal{X} \times \mathcal{X}$ is associated with an RKHS and vice versa \shortcite{aronszajn1950theory}.

Kernel methods are a class of machine learning algorithms which operate on input data after they have been mapped into an implicit feature space using a kernel function.
One of the major advantages of kernel methods is that they can operate on very general types of data \shortcite{scholkopf2002learning}.
The input space $\mathcal{X}$ does not have to be a vector space, but it can represent any structured domain, such as the space of strings or graphs \shortcite{gartner2003survey}.
Kernel methods can still be applied to such types of data, as long as we can find a mapping $\phi : \mathcal{X} \rightarrow \mathcal{H}$, where $\mathcal{H}$ is an RKHS.
This mapping is not neccasary to be explicitly determined. 
These methods implicitly represent data in a feature space and compute inner products between them in that space using a kernel function.
These inner products can be interpreted as the similarities between the corresponding objects.
Machine learning tasks such as classification and clustering can be carried out by using only the inner products computed in that feature space.
Kernel methods are very popular and have been successfully used in a wide variety of applications.
Here, we need to stress that the optimization problem of several kernel methods such as the Support Vector Machines is convex only if the employed function is positive semidefinite.

\section{Graph Kernels}\label{sec:graph_kernels}
In this Section, we give an overview of the graph kernel literature.
Our study is not exhaustive, however, we have tried to cover the most representative approaches that have appeared in the literature of graph kernels.
We first present some fundamental aspects of graph kernels, and we then proceed by discussing the details of several graph kernel instances.

\subsection{Kernels between Graphs}
Kernels on graphs can be divided into two categories: ($1$) those that compare nodes in a graph, and ($2$) those that compare graphs.
As mentioned above, in this survey, we focus on the second category, that is, \textit{kernels between graphs} and thus we exclusively use the term \textit{graph kernel }for describing such kernel functions.
As regards the first category, we refer the interested reader to the work of \shortciteA{kondor2002diffusion} which was later extended by \shortciteA{smola2003kernels}.
Graph kernels have recently emerged as a promising approach for learning on graph-structured data.
These methods exhibit several attractive statistical properties.
They combine the representative power of graphs and the discrimination power of kernel-based methods.
Hence, they constitute powerful tools for tackling the graph similarity and learning tasks at the same time.

\begin{figure}
  \centering
  \includegraphics[width=.7\linewidth]{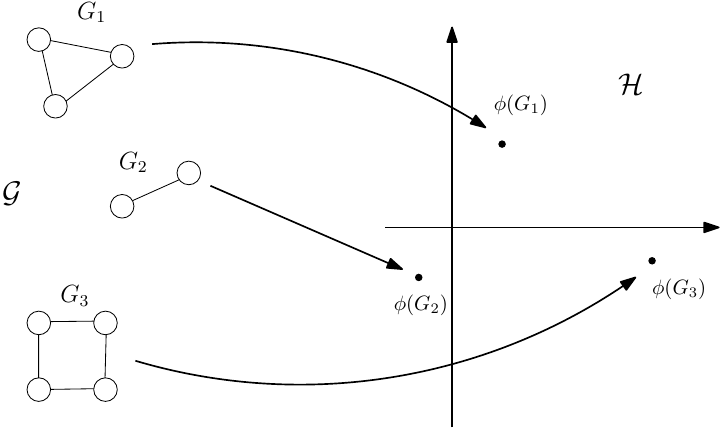}
    \caption{Feature space and map defined by graph kernels.
    Any kernel on a space of graphs $\mathcal{G}$ can be represented as an inner product after graphs are mapped into a Hilbert space $\mathcal{H}$.}
    \label{fig:mapping}
\end{figure}

From the previous Section, it is clear that the application of kernel methods consists of two steps.
First, a kernel function is designed, and based on this function the kernel matrix is constructed.
Second, a learning algorithm is employed to compute the optimal manifold in the feature space (\eg a hyperplane in binary classification problems).
Since several mature kernel-based classifiers are available in the literature, research on graph kernels has focused on the first step.
Hence, the main effort has been devoted to developing expressive and efficient graph kernels capable of accurately measuring the similarity between input graphs.
These kernels implicitly (or explicitly sometimes) project graphs into a feature space $\mathcal{H}$ as illustrated in Figure~\ref{fig:mapping}.
As regards the second step, it is common to employ off-the-shelf algorithms such as the Support Vector Machines classifier \shortcite{boser1992training} or the kernel $k$-means algorithm \shortcite{dhillon2004kernel}, and thus, we will not enter into more details here.
The interested reader is referred to \shortciteA{scholkopf2002learning} or to \shortciteA{shawe2004kernel}.

Concluding, the main challenge in applying kernel methods to graphs is  to \textit{define appropriate positive semidefinte kernel functions on the set of input graphs which are able to reliably assess the similarity among them}.
We next present, for illustration purposes, two very simple kernels that compare node and edge labels of the two involved graphs.

\subsection{Simple Kernels}
The vertex histogram and edge histogram kernels are very simple instances of graph kernels which generate explicit graph representations.

\subsubsection{Vertex Histogram Kernel}
The vertex histogram kernel is a basic linear kernel on vertex label histograms.
The kernel assumes node-labeled graphs.
Let $\Sigma=\{1,\ldots,d\}$ be a set of node labels.
Clearly, there are $d$ node labels in total, that is $d = |\Sigma|$.
Then, the vertex label histogram of a graph $G=(V,E)$ is a vector $f = (f_1, f_2, \ldots, f_d)^\top$, such that $f_i = |\{ v \in V : \ell(v) = i \}|$ for each $i \in \Sigma$.
Let $f, f'$ be the vertex label histograms of two graphs $G, G'$, respectively.
The vertex histogram kernel is then defined as the linear kernel between $f$ and $f'$, that is
\begin{equation}
    k(G, G') = \langle f, f' \rangle
\end{equation}
The complexity of the vertex histogram kernel is linear in the number of vertices of the graphs.

\subsubsection{Edge Histogram Kernel}
The edge histogram kernel is a basic linear kernel on edge label histograms.
The kernel assumes edge-labeled graphs.
Given a set of edge labels $\Sigma=\{1,\ldots,d\}$ ($d$ edge labels in total), the edge label histogram of a graph $G=(V,E)$ is a vector $f = (f_1, f_2, \ldots, f_d)^\top$, such that $f_i = |\{ (v,u) \in E : \ell(v,u) = i \}|$ for each $i \in \Sigma$.
Let $f, f'$ be the edge label histograms of two graphs $G, G'$, respectively.
The edge histogram kernel is then defined as the linear kernel between $f$ and $f'$, that is
\begin{equation}
    k(G, G') = \langle f, f' \rangle
\end{equation}
The complexity of the edge histogram kernel is linear in the number of edges of the graphs.

\subsection{Expressiveness vs Efficiency}
The two kernels defined above are indeed positive semidefinite, but they both correspond to rather naive concepts - as a distribution of values is.
A question that may arise at this point is how expressive can graph kernels be in practice.

Let us first define the class of kernels which are capable of distinguishing between all (non-isomorphic) graphs in the feature space.
Such kernels are called \textit{complete}.
\begin{definition}[Complete Graph Kernel]
  A graph kernel $k(G_i,G_j) = \langle \phi(G_i), \phi(G_j) \rangle$ is complete if $\phi$ is injective.
\end{definition}
\shortciteA{gartner2003graph} showed that computing any complete graph kernel is at least as hard as deciding whether two graphs are isomorphic.
The above result, in effect, prohibits the use of complete graph kernels in practical applications.
Instead, by using kernels that are not complete, it is not further guaranteed that non-isomorphic graphs will not be mapped into the same point in the feature space.
This is a negative result since it implies that to develop expressive kernels, it is necessary to sacrifice some of their efficiency.
More recently, \shortciteA{kriege2018property} showed that several established graph kernels, such as the Weisfeiler-Lehman subtree kernel, cannot distinguish essential graph properties such as connectivity, planarity and bipartiteness.
Considering that the Weisfeiler-Lehman subtree kernel achieves state-of-the-art results on most benchmark datasets, this result blurs even more the already vague issue of choosing a graph kernel a practitioner is faced with when dealing with a particular application.
In fact, devising a good trade-off between efficiency and effectiveness is an issue of vital importance when designing a graph kernel.

\subsection{Taxonomy of Graph Kernels}
\begin{figure}[t]
    \centering
    \includegraphics[width=.8\linewidth]{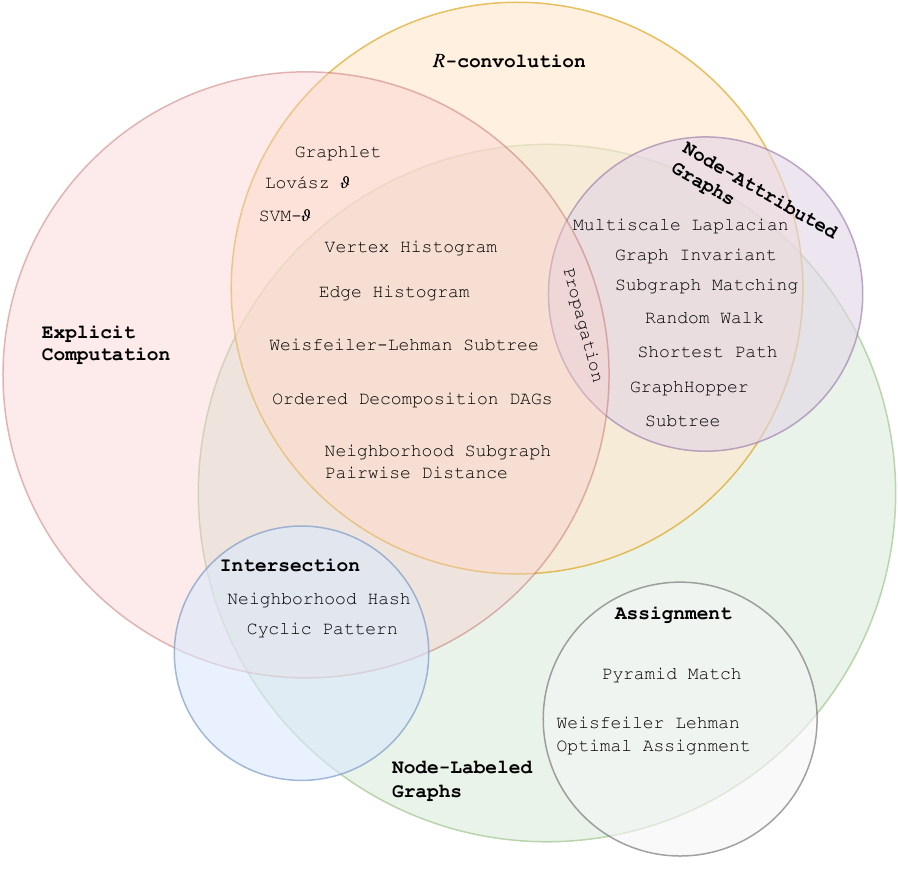}
    \caption{Taxonomy of graph kernels.}
    \label{fig:taxonomy}
\end{figure}

There exist many different criteria we can use to divide the various graph kernels into different categories.
For instance, graph kernels are traditionally grouped into some major families, each focusing on a different structural aspect of graphs such as random walks, subtrees, cycles, paths, and small subgraphs.
Alternatively, graph kernels can be divided into groups according to their ability to handle unlabeled graphs, node-labeled or node-attributed graphs.
Furthermore, graph kernels can be divided into approaches that employ explicit computation schemes and approaches that employ implicit computation schemes \shortcite{kriege2014explicit}.
Graph kernels can also be grouped into categories based on the design paradigm that they follow (\ie if they are $R$-convolution, assignment or intersection kernels).
Note that groups emerging from different criteria may be related to each other.
For instance, graph kernels that can handle node-attributed graphs usually employ implicit computation schemes.
Figure~\ref{fig:taxonomy} illustrates the taxonomy of graph kernels.
The devised taxonomy is based on some of the criteria mentioned above. 
However, in what follows, we do not adopt exclusively any of these criteria.
We begin our treatment with approaches that were proposed in the early days of graph kernels, starting from the well-studied random walk kernel till the very popular Weisfeiler-Lehman subtree kernel.
We next present some approaches that were inspired from the neighborhood aggregation schmeme of the Weisfeiler-Lehman subtree kernel, and then kernels that do not fall into either of the previous two categories.
The subequent subsections are devoted to assignment kernels, and to kernels that can handle continuous node attributes.
The final subsections deals with frameworks and approaches that can be applied on top of existing graph kernels.
An overview of the graph kernels that are presented in this survey and their properties is given in Table~\ref{tab:comparison}.
\begin{table}[t]
  \centering
  \scriptsize
  \def\arraystretch{1.2}
  \begin{tabular}{l|c|c|c|c|c}
    \multirow{2}{*}{Graph Kernel}& \multirow{2}{*}{Exp. $\phi$} & Node & Node & \multirow{2}{*}{Type} & \multirow{2}{*}{Complexity} \\ 
    & & Labels & Attributes & \\ \hline
    Vertex Histogram & $\cmark$ & $\cmark$ & $\xmark$ & $R$-convolution & $\mathcal{O}(n)$ \\ 
    Edge Histogram & $\cmark$ & $\cmark$ & $\xmark$ & $R$-convolution & $\mathcal{O}(m)$ \\ 
    Random Walk & $\xmark^{\dagger}$ & $\cmark$ & $\cmark$ & $R$-convolution & $\mathcal{O}(n^3)$ \\ 
    Subtree & $\xmark$ & $\cmark$ & $\cmark$ & $R$-convolution & $\mathcal{O}(n^2 4^{deg^*} h)$ \\ 
    Cyclic Pattern & $\cmark$ & $\cmark$ & $\xmark$ & intersection & $\mathcal{O}((c+2)n+2m)$ \\ 
    Shortest Path & $\xmark^{\dagger}$ & $\cmark$ & $\cmark$ & $R$-convolution  & $\mathcal{O}(n^4)$ \\ 
    Graphlet & $\cmark$ & $\xmark$ & $\xmark$ & $R$-convolution & $\mathcal{O}(n^k)$ \\ 
    Weisfeiler-Lehman Subtree & $\cmark$ & $\cmark$ & $\xmark$ & $R$-convolution & $\mathcal{O}(hm)$ \\ 
    Neighborhood Hash & $\cmark$ & $\cmark$ & $\xmark$ & intersection & $\mathcal{O}(hm)$ \\ 
    Neighborhood Subgraph Pairwise Distance & $\cmark$ & $\cmark$ & $\xmark$ & $R$-convolution & $\mathcal{O}(n^2 m \log(m))$ \\ 
    Lov\'asz $\vartheta$ & $\cmark$ & $\xmark$ & $\xmark$ & $R$-convolution & $\mathcal{O}(n(s+\frac{nm}{\epsilon})+s^2)$ \\ 
    SVM-$\vartheta$ & $\cmark$ & $\xmark$ & $\xmark$ & $R$-convolution & $\mathcal{O}(n(s+n^2)+s^2)$ \\ 
    Ordered Decomposition DAGs & $\cmark$ & $\cmark$ & $\xmark$ & $R$-convolution & $\mathcal{O}(n \log n)$ \\ 
    Pyramid Match & $\xmark$ & $\cmark$ & $\xmark$ & assignment & $\mathcal{O}(ndL)$ \\ 
    Weisfeiler-Lehman Optimal Assignment & $\xmark$ & $\cmark$ & $\xmark$ & assignment & $\mathcal{O}(hm)$ \\ 
    Subgraph Matching & $\xmark$ & $\cmark$ & $\cmark$ & $R$-convolution & $\mathcal{O}(kn^{k+1})$ \\ 
    GraphHopper & $\xmark$ & $\cmark$ & $\cmark$ & $R$-convolution & $\mathcal{O}(n^4)$ \\ 
    Graph Invariant Kernels & $\xmark$ & $\cmark$ & $\cmark$ & $R$-convolution & $\mathcal{O}(n^6)$ \\ 
    Propagation & $\cmark$ & $\cmark$ & $\cmark$ & $R$-convolution & $\mathcal{O}(hm)$ \\ 
    Multiscale Laplacian & $\xmark$ & $\cmark$ & $\cmark$ & $R$-convolution & $\mathcal{O}(n^5 h)$ \\ \hline
  \end{tabular}
  \caption{Summary of selected graph kernels regarding computation by explicit feature mapping (Exp. $\phi$), support for node-labeled and node-attributed graphs, type, and computational complexity. A dagger ($\dagger$) implies that the kernel admits an explicit feature mapping for certain types of graphs. The complexity refers to the worst-case theoretical complexity for evaluating the kernel between two graphs. In practice, and for certain kinds of graphs, some graph kernels (\eg the shortest-path kernel) can be evaluated much more efficiently. The Table uses notation that has not been introduced yet: $k$: size of largest subgraph considered, $c$: upper bound on the number of cycles, $h$: maximum distance between root of neighborhood subgraph/subtree pattern and its nodes, $s$: number of sampled subgraphs, $\epsilon$: additive error associated with semidefinite programming solvers, $d$: dimensionality of node representations, $L$: number of levels.}
  \label{tab:comparison}
\end{table}

\subsection{Early Days of Graph Kernels}
While early studies on kernel functions and kernel methods focused almost exclusively on input data represented as vectors, it soon became clear that these methods could handle more complex structured objects such as strings, trees and graphs.
One of the most popular methods for defining kernels between such objects is to decompose the objects into their ``parts'', and to compare all pairs of these ``parts'' by applying existing kernels on them.
Kernels constructed using the above framework are called \textit{R-convolution} kernels \shortcite{haussler1999convolution}.
Most graph kernels in the literature are instances of the $R$-convolution framework.
These kernels decompose graphs into their substructures and add up the pairwise similarities between these substructures.

The most intuitive example of an $R$-convolution kernel is probably a kernel that decomposes each graph into the set of all of its subgraphs, and compares them pairwise.
\shortciteA{gartner2003graph} showed that the problem of computing the kernel that compares all the subgraphs of two graphs is NP-hard.
Based on this result, it becomes evident that we need to consider alternative, less powerful graph kernels that can be computed in polynomial time.
However, as discussed above, it is necessary that these kernels provide an expressive measure of similarity on graphs.
Over the years,  several graph kernels have been proposed, each focusing on a \textit{different structural aspect} of graphs.
Such aspects involve comparing graphs based on random walks, subtrees, cycles, paths, and small subgraphs, to name a few.
We next look at some kernels that date back to the early days of this field.
Furthermore, we present kernels that were motivated by problems encountered by the above instances, and were proposed as more advanced alternatives. 

\subsubsection{Random Walk Kernel}
\begin{figure}[t]
    \centering
    \includegraphics[width=.5\linewidth]{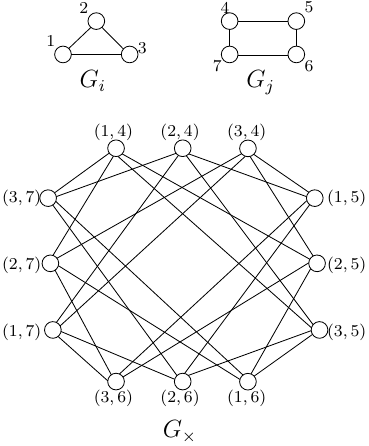}
    \caption{Two graphs (top left and right) and their direct product (bottom). Each vertex of the direct product graph is labeled with a pair of vertices; an edge exists in the direct product if and only if the corresponding vertices are adjacent in both original graphs. For instance, nodes $1-4$ and $3-5$ are adjacent because there is an edge between vertices $1$ and $3$ in the first, and $4$ and $5$ in the second graph.}
    \label{fig:product_graph}
\end{figure}

The random walk kernels are perhaps one of the first successful efforts to design kernels between graphs that can be computed in polynomial time.
The members of this well-studied family of graph kernels quantify the similarity between a pair of graphs based on the number of common walks in the two graphs \shortcite{kashima2003marginalized,gartner2003graph,mahe2004extensions,borgwardt2005protein,vishwanathan2010graph,sugiyama2015halting,zhang2018retgk}.
Kernels belonging to this family have concentrated mainly on counting matching walks in the two input graphs.
There are several variations of random walk kernels.
The $k$-step random walk kernel compares random walks up to length $k$ in the two graphs.
The most widely-used kernel from this family is the geometric random walk kernel \shortcite{gartner2003graph} which compares walks up to infinity assigning a weight $\lambda^k$ ($\lambda < 1$) to walks of length $k$ in order to ensure convergence of the corresponding geometric series.
We next give the formal definition of the geometric random walk kernel.
Given two node-labeled graphs $G=(V,E)$ and $G'=(V',E')$, their direct product $G_\times=(V_\times,E_\times)$ is a graph with vertex set:
\begin{equation}
  V_{\times} = \{(v,v') : v \in V \wedge v' \in V' \wedge \ell(v) = \ell(v') \} 
\end{equation}
and edge set:
\begin{equation}
  E_{\times} = \{\{(v,v'),(u,u')\} : (v,u) \in E \wedge (v',u') \in E'\}
\end{equation}
An example of the product graph of two graphs is illustrated in Figure~\ref{fig:product_graph}.
Performing a random walk on $G_{\times}$ is equivalent to performing a simultaneous random walk on $G_i$ and $G_j$.
The geometric random walk kernel counts common walks (of potentially infinite length) in two graphs and is defined as follows.  
\begin{definition}[Geometric Random Walk Kernel]
  Let $G$ and $G'$ be two graphs, let $A_\times$ denote the adjacency matrix of their product graph $G_\times$, and let $V_\times$ denote the vertex set of the product graph $G_\times$.
  Then, the geometric random walk kernel is defined as
  \begin{equation}
      K_{\times}^{\infty}(G,G') = \sum_{p,q=1}^{|V_{\times}|} \Big[ \sum_{l=0}^{\infty} \lambda^l A_{\times}^l \Big]_{pq} = e^\top(I - \lambda A_{\times})^{-1} e
    \end{equation}
  where $I$ is the identity matrix, $e$ is the all-ones vector, and $\lambda$ is a positive, real-valued weight.
  The geometric random walk kernel converges only if $\lambda < \frac{1}{\lambda_\times}$ where $\lambda_\times$ is the largest eigenvalue of $A_{\times}$.
\end{definition}
Direct computation of the geometric random walk kernel requires $\mathcal{O}(n^6)$ time.
The computational complexity of the method severely limits its applicability to real-world applications.
To account for this, \shortciteA{vishwanathan2010graph} proposed four efficient methods to compute random walk graph kernels which generally reduce the computational complexity from $\mathcal{O}(n^6)$ to $\mathcal{O}(n^3)$.
\shortciteA{mahe2004extensions} proposed some other extensions of random walk kernels.
Specifically, they proposed a label enrichment approach which increases specificity and in most cases also reduces computational complexity.
They also employed a second order Markov random walk to deal with the problem of ``tottering''.
\shortciteA{sugiyama2015halting} focused on a different problem of random walk kernels, a phenomenon referred to as ``halting''.
More recently, \shortciteA{zhang2018retgk} proposed a kernel that capitalizes on the isomorphism-invariance property of the return probabilities of random walks.

\subsubsection{Subtree Kernel}
Due to problems with the expressiveness of the random walk kernels that they identified, \shortciteA{ramon2003expressivity} worked on designing new kernels.
Their research efforts resulted in the development of the subtree kernel, an algorithm that counts the number of common subtree patterns in two graphs.
The kernel is more expressive (in the sense that it can distinguish non-isomorphic graphs which walk-based kernels cannot), but also more computationally expensive than the random walk kernels.

The subtree patterns that the subtree kernel considers correspond to rooted subgraphs.
Every subtree pattern has a tree-structured signature, and the kernel associates each possible subtree pattern signature to a feature.
Given a graph, the value of each feature is the number of times that a subtree of the signature that corresponds to this feature occurs in the graph.
Let $k_h(v,v')$ be a kernel that counts the pairs of subtrees of the same signature of height less than or equal to $h$, where the first subtree is rooted at $v$ and the second one is rooted at $v'$.
The kernel $k_h(v,v')$ is equal to:
\begin{equation}
  k_h(v, v') =
  \begin{cases}
    \delta(\ell(v), \ell(v')) & \text{if } h=1 \\
    \lambda_v \lambda_{v'} \sum_{R \in M(v,v')} \prod_{(u,u') \in R} k_{h-1}(u,u') & \text{if } h > 1
  \end{cases}
\end{equation}
where $\lambda_v$ and $\lambda_{v'}$ are positive values smaller than $1$ to cause higher trees to have a smaller weight in the overall sum, and $\delta$ is the dirac kernel.
Therefore, if $h=1$ and the two nodes share the same label, then it holds that $k_1(v,v')=1$.
If $h=1$ and the two nodes have different labels, we have $k_1(v,v')=0$.
For $h \geq 1$, one can compute $k_h(v,v')$ using a recursive scheme.
Specifically, we define the set of all matchings from $\mathcal{N}(v)$ to $\mathcal{N}(v')$ as follows
\begin{equation}
  \begin{split}
    M(v,v') = \Big\{ R \subseteq \mathcal{N}(v) \times \mathcal{N}(v') &| \big( \forall (u,u'),(w,w') \in R : u=w \Leftrightarrow u'=w' \big) \\
    &\wedge \big( \forall (u,u') \in R : \ell(u) = \ell(u') \big) \Big\}
  \end{split}
\end{equation}
Each element $R$ of $M(v,v')$ is a set of pairs of nodes from the neighborhoods of $v \in V$ and $v' \in V'$, such that nodes in each pair have identical labels and no node is contained in more than one pair.
The subtree kernel compares all pairs of vertices from two graphs by iteratively comparing their neighborhoods.
\begin{definition}[Subtree Kernel]
  Let $G=(V,E)$ and $G'=(V',E')$ be two graphs.
  Then, the subtree kernel is defined as
  \begin{equation}
    k(G, G') = \sum_{v \in V} \sum_{v' \in V'} k_h(v,v')
  \end{equation}
\end{definition}
The computational complexity of the subtree kernel for a pair of graphs is $\mathcal{O}(n^2 4^{deg^*} h)$.
Although in the worst-case scenario, the runtime complexity of the subtree kernel is very high, in practice, it can be quite low if the input graphs are sparse or if there is sufficient diversity in the labels of the vertices.

\subsubsection{Cyclic Pattern Kernel}
The cyclic pattern kernel is also one of the earliest approaches developed in the area of graph kernels.
This kernel decomposes a graph into cyclic and tree patterns, and counts the number of common patterns which occur in two graphs \shortcite{horvath2004cyclic}.
More specifically, let $G=(V,E)$ be a graph.
Let also $\mathcal{S}(G)$ denote the set of cycles of $G$.
Let $C = (v_1, v_2, \ldots, v_k, v_1)$ be a sequence of vertices that forms a cycle in $G$, that is $C \in \mathcal{S}(G)$.
The canonical representation of a cycle $C$ is the lexicographically smallest string $\pi(C)$ among the strings obtained by concatenating the labels along the vertices of the cyclic permutations of $C$ and its reverse.
Formally, denoting by $\rho(s)$ the set of cyclic permutations of a sequence $s$ and its reverse, the canonical representation of $C$ is defined by
\begin{equation}
  \pi(C) = \min\{ w : w \in \rho\big(\ell(v_1), \ell(v_2), \ldots, \ell(v_k) \big)\}
\end{equation}
where $\ell$ is a function that assigns labels to the vertices of the graph.
In case of edge-labeled graphs, edgle labels can also be taken into account.
The set of cyclic patterns of $G$ is then defined by
\begin{equation}
  \mathcal{C}(G) = \{ \pi(C) : C \in \mathcal{S}(G) \}
\end{equation}

The kernel then extracts from $G$ all the edges that do not belong to any cycle (a.k.a bridges) by removing from $G$ all the edges of all cycles.
The set of bridges of $G$ forms a set of trees (each tree is a connected component composed of bridges).
Then, similarly to cycles, the kernel computes the canonical representation $\pi(T)$ of each tree $T$.
The set of tree patterns of $G$ is then defined by
\begin{equation}
  \mathcal{T}(G) = \{ \pi(T) : T \text{ is a tree} \}
\end{equation}
Then, given two graphs, the kernel computes the intersection of their sets of cyclic and tree patters.
\begin{definition}[Cyclic Pattern Kernel]
  Let $G$, $G'$ be two graphs, and $\mathcal{C}(G), \mathcal{C}(G')$ and $\mathcal{T}(G), \mathcal{T}(G')$ be the sets of cyclic patterns and tree patters of the two graphs, respectively.
  Then, the cyclic pattern kernel is defined as
  \begin{equation}
    k(G,G') = |\mathcal{C}(G) \cap \mathcal{C}(G')| + |\mathcal{T}(G) \cap \mathcal{T}(G')|
  \end{equation}
\end{definition}
Unfortunately, computing the cyclic pattern kernel is an NP-hard problem.
The cardinality of the set of cyclic and tree patterns of a graph can be exponential in the number of vertices of the graph. 
However, the cyclic pattern kernel can prove useful for practical problem classes where the number of cycles in the input graphs is bounded.

\subsubsection{Shortest-Path Kernel}
The high computational complexity of graph kernels based on walks, subtrees and cycles renders them impractical for most real-world scenarios.
\shortciteA{borgwardt2005shortest} worked on developing more efficient kernels based on paths.
However, computing all the paths in a graph and computing the longest paths in a graph are both NP-hard problems.
Instead, shortest paths can be computed in polynomial time, and they gave rise to the shortest-path kernel, one of the most popular kernels to this day.

The shortest-path kernel decomposes graphs into shortest paths and compares pairs of shortest paths according to their lengths and to the labels of their endpoints.
The first step of the shortest-path kernel is to transform the input graphs into shortest-paths graphs.
Given an input graph $G=(V,E)$, the algorithm creates a new graph $S=(V,E_s)$ (\ie its shortest-path graph).
The shortest-path graph $S$ contains the same set of vertices as its source graph.
The edge set of the former is a superset of that of the latter, since in the shortest-path graph $S$, there exists an edge between all vertices that are connected by a walk in the original graph $G$.
To complete the transformation, the algorithm assigns labels to all the edges of the shortest-path graph $S$.
The label of each edge is set equal to the shortest distance between its endpoints in the original graph $G$.

\begin{figure}[t]
    \centering
    \includegraphics[width=.9\linewidth]{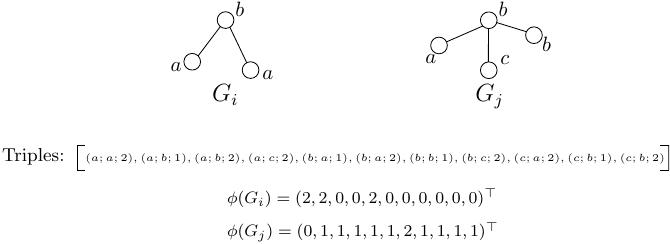}
    \caption{Example of explicit computation of the shortest path kernel. Each triple is a feature and corresponds to: (label of source vertex; label of sink vertex; shortest path length between the two vertices).}
    \label{fig:shortest_path}
\end{figure}

Given the above procedure for transforming a graph into a shortest-path graph, the shortest-path kernel is defined as follows.
\begin{definition}[Shortest-Path Kernel]
  Let $G$, $G'$ be two graphs, and $S=(V,E)$, $S'=(V',E')$ their corresponding shortest-path graphs.
  The shortest-path kernel is then defined as
  \begin{equation}
    k(G,G') = \sum_{e \in E} \sum_{e' \in E'} k_{walk}^{(1)}(e, e')
  \end{equation}
  where $k_{walk}^{(1)}(e, e')$ is a positive semidefinite kernel on edge walks of length $1$.
\end{definition}
In labeled graphs, the $k_{walk}^{(1)}(e, e')$ kernel is designed to compare both the lengths of the shortest paths corresponding to edges $e$ and $e'$, and the labels of their endpoint vertices.
Let $e = (v, u)$ and $e' = (v', u')$.
Then, $k_{walk}^{(1)}(e, e')$ is usually defined as
\begin{equation}
  k_{walk}^{(1)}(e, e') = k_v \big(\ell(v),\ell(v') \big) \ k_e \big(\ell(e),\ell(e') \big) \ k_v \big(\ell(u),\ell(u') \big)
\end{equation}
where $k_v$ is a kernel comparing vertex labels, and $k_e$ a kernel comparing shortest path lengths.
Vertex labels are usually compared via a dirac kernel, while shortest path lengths may also be compared via a dirac kernel or, more rarely, via a brownian bridge kernel \shortcite{borgwardt2005shortest}.
When $k_v$ and $k_e$ both are dirac kernels, an explicit computation scheme can be employed as shown in Figure~\ref{fig:shortest_path}.
In terms of runtime complexity, the shortest-path kernel can be computed in $\mathcal{O}(n^4)$ time.

\subsubsection{Graphlet Kernel}
\begin{figure}[t]
    \centering
    \includegraphics[width=.7\linewidth]{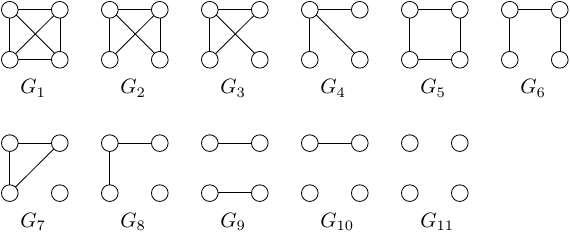}
    \caption{All graphlets of size $4$.}
    \label{fig:graphlets}
\end{figure}

The graphlet kernel decomposes graphs into graphlets (\ie small subgraphs with $k$ vertices where $k \in \{ 3,4,5\}$) \shortcite{prvzulj2007biological} and counts matching graphlets in the input graphs.
For example, the set of graphlets of size $4$ is shown in Figure~\ref{fig:graphlets}.
This kernel was originally designed to address scalability issues experienced by earlier approaches.
In fact, the graphlet kernel was one of the first kernels that could cope with very large graphs using a simple sampling scheme.
However, apart from the scalability issue, the graphlet kernel was also motivated by the graph reconstruction conjecture \shortcite{bondy1977graph}, which states that any graph of size $n$ can be reconstructed from the set of all its subgraphs of size $n-1$.
This could possibly be interpreted as indicating that kernels that compare graphs based on their subgraphs should reflect graph similarity better than approaches that are defined based on random walks, subtrees, cyclic patterns or shortest paths.
However, even if graphs that have similar distributions of graphlets are very likely to be similar themselves, there is no theoretical justification on why such a substructure (\ie graphlets) is better than the others.

As mentioned above, the graphlet kernel computes the distribution of small subgraphs in a graph.
Let $\mathcal{G} = \{ graphlet_1$, $graphlet_2$, $\ldots$, $graphlet_d\}$ be the set of size-$k$ graphlets.
Let also $f_G \in \mathbb{N}^d$ be a vector such that its $i$-th entry is equal to the frequency of occurrence of $graphlet_i$ in $G$, $f_{G,i} = \#(graphlet_i \sqsubseteq G)$.
Then, the graphlet kernel is defined as follows.
\begin{definition}[Graphlet of size $k$ Kernel]
  Let $G$, $G'$ be two graphs of size $n \geq k$, and $f_{G}, f_{G'}$ vectors that count the occurrence of each graphlet of size $k$ (not necessarily connected) in the two graphs. 
  Then the graphlet kernel is defined as
  \begin{equation}
      k(G,G') = f_{G}^\top \ f_{G'}
    \end{equation}
\end{definition}
As is evident from the above definition, the graphlet kernel is computed by explicit feature maps.
First, the representation of each graph in the feature space is computed.
And then, the kernel value is computed as the dot product of the two feature vectors.
The main problem of the graphlet kernel is that an exaustive enumeration of graphlets is very expensive.
Since there are $\binom{n}{k}$ size-$k$ subgraphs in a graph, computing the feature vector for a graph of size $n$ requires $\mathcal{O}(n^k)$ time.
To account for that, \shortciteA{shervashidze2009efficient} resorted to sampling. 
Following \shortciteA{weissman2003inequalities}, they showed that by sampling a fixed number of graphlets the empirical distribution of graphlets will be sufficiently close to their actual distribution in the graph.  
An alternative proposed strategy that reduces the expressivity of the kernel is to enumerate only the connected graphlets of $k$ vertices, and not all the possible graphlets.

\subsubsection{Weisfeiler-Lehman Subtree Kernel}
The Weisfeiler-Lehman subtree kernel is a very popular algorithm, and is considered the state-of-the-art in graph classification.
It belongs to the family of subtree kernels, and was motivated by the need for a fast subtree kernel that scales up to large, labeled graphs.
The kernel is an instance of the Weisfeiler-Lehman framework.
This framework operates on top of existing graph kernels and is inspired by the Weisfeiler-Lehman test of graph isomorphism \shortcite{weisfeiler1968reduction}.
The key idea of the Weisfeiler-Lehman algorithm is to replace the label of each vertex with a multiset label consisting of the original label of the vertex and the sorted set of labels of its neighbors.
The resultant multiset is then compressed into a new, short label.
This relabeling procedure is then repeated for $h$ iterations.
Note that this procedure is performed simultaneously on all input graphs.
Therefore, two vertices from different graphs will get identical new labels if and only if they have identical multiset labels.

More formally, given a graph $G=(V,E)$ endowed with a labeling function $\ell=\ell_0$, the Weisfeiler-Lehman graph of $G$ at height $i$ is a graph $G_i=(V,E)$ endowed with a labeling function $\ell_i$ which has emerged after $i$ iterations of the relabeling procedure described above.
The Weisfeiler-Lehman sequence up to height $h$ of $G$ consists of the Weisfeiler-Lehman graphs of $G$ at heights from $0$ to $h$, $\{ G_0,G_1,\ldots,G_h\}$. 
\begin{definition}[Weisfeiler-Lehman Framework]
  Let $k$ be any kernel for graphs, that we will call the base kernel.
  Then the Weisfeiler-Lehman kernel with $h$ iterations with the base kernel $k$ between two graphs $G$ and $G'$ is defined as
  \begin{equation}
    k_{WL}(G,G') = k(G_0,G_0') + k(G_1,G_1') + \ldots + k(G_h,G_h')
  \end{equation}
  where $h$ is the number of Weisfeiler-Lehman iterations, and $\{ G_0,G_1,\ldots,G_h\}$ and $\{ G_0',G_1'$, $\ldots,G_h'\}$ are the Weisfeiler-Lehman sequences of $G$ and $G'$ respectively.
\end{definition}
From the above definition, it is clear that any graph kernel that takes into account discrete node labels can take advantage of the Weisfeiler-Lehman framework and compare graphs based on the whole Weisfeiler-Lehman sequence.

\begin{figure}[t]
    \centering
    \includegraphics[width=\linewidth]{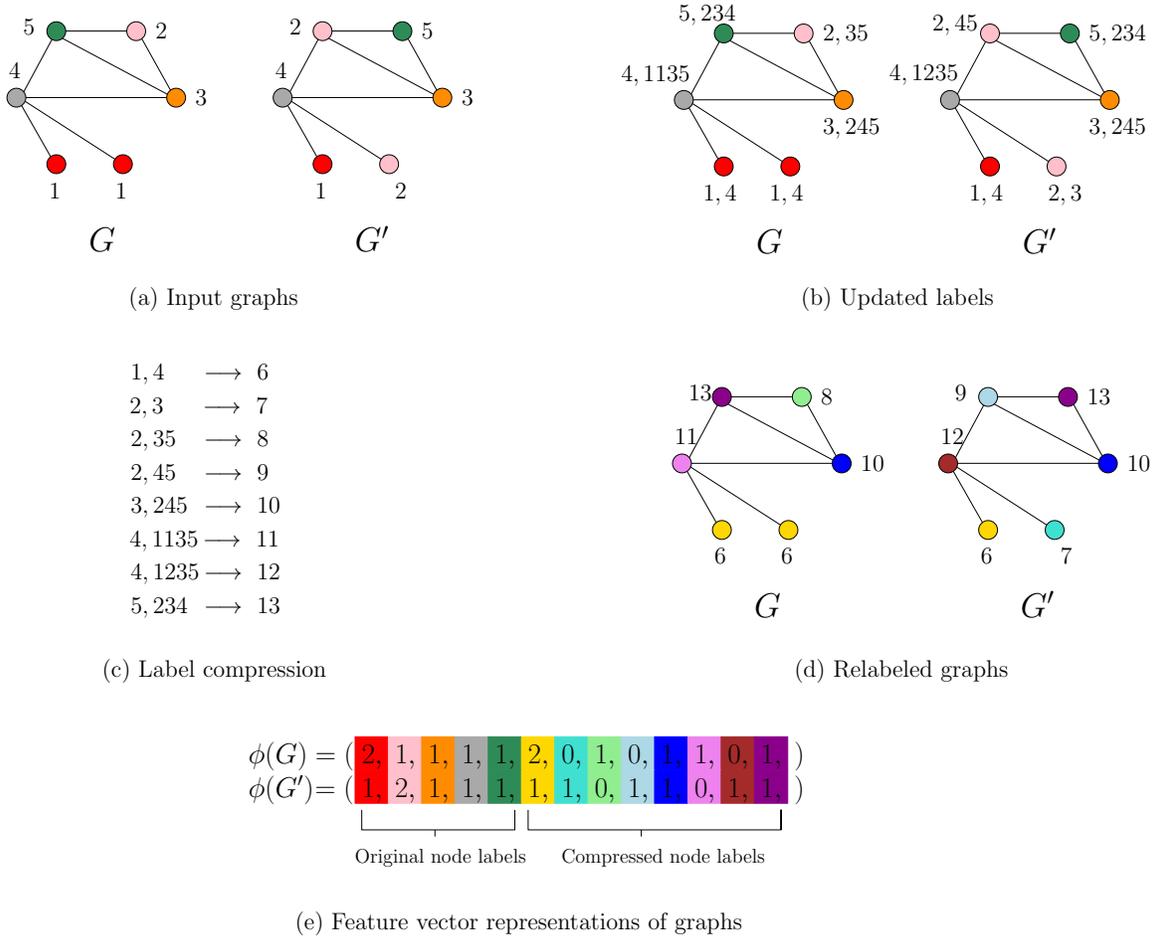}
    \caption{Illustration of the computation of the Weisfeiler-Lehman subtree kernel with $h=1$ for two  graphs $G$ and $G'$. Here, $ 1,2,\ldots,13 \in \Sigma$ are letters that occur as node labels. Compressed labels map to subtree patterns. For example, if a node has label $6$, this means that there is a subtree pattern of height $1$ rooted at this node, where the root has label $1$ and its single neighbor has label $4$.}
    \label{fig:wl_example}
\end{figure}

When the base kernel compares subtrees extracted from two graphs, the computation involves counting the common original and compressed labels in the two graphs.
The emerging Weisfeiler-Lehman subtree kernel is a byproduct of the Weisfeiler-Lehman test of isomorphism.
\begin{definition}[Weisfeiler-Lehman Subtree Kernel]
  Let $G$, $G'$ be two graphs.
  Define $\Sigma_i \subseteq \Sigma$ as the set of letters that occur as node labels at least once in $G$ or $G'$ at the end of the $i$-th iteration of the Weisfeiler-Lehman algorithm.
  Let $\Sigma_0$ be the set of original node labels of $G$ and $G'$.
  Assume all $\Sigma_i$ are pairwise disjoint.
  Without loss of generality, assume that every $\Sigma_i = \{ \sigma_{i1},\ldots,\sigma_{i|\Sigma_i|} \}$ is ordered.
  Define a map $c_i : \{ G,G' \} \times \Sigma_i \rightarrow \mathbb{N}$ such that $c_i(G, \sigma_{ij})$ is the number of occurrences of the letter $\sigma_{ij}$ in the graph $G$.

  The Weisfeiler-Lehman subtree kernel on two graphs $G$ and $G'$ with $h$ iterations is defined as
  \begin{equation}
    k(G,G') = \langle \phi(G),\phi(G') \rangle 
  \end{equation}
  where
  \begin{equation}
    \phi(G) = (c_0(G,\sigma_{01}),\ldots,c_0(G,\sigma_{0|\Sigma_0|}),\ldots,c_h(G,\sigma_{h1}),\ldots,c_h(G,\sigma_{h|\Sigma_h|}))
  \end{equation}
  and
  \begin{equation}
    \phi(G') = (c_0(G',\sigma_{01}),\ldots,c_0(G',\sigma_{0|\Sigma_0|}),\ldots,c_h(G',\sigma_{h1}),\ldots,c_h(G',\sigma_{h|\Sigma_h|}))
  \end{equation}
\end{definition}
An illustration of the Weisfeiler-Lehman subtree kernel is given in Figure~\ref{fig:wl_example}.
It can be shown that the above definition is equivalent to comparing the number of shared subtrees between the two input graphs \shortcite{shervashidze2011weisfeiler}.
In contrast to the subtree kernel that was proposed by Ramon and G{\"a}rtner and was presented above, the Weisfeiler-Lehman subtree kernel considers all subtrees up to height $h$, instead of subtrees of exactly height $h$.
Furthermore, the Weisfeiler-Lehman subtree kernel checks whether the neighborhoods of two vertices match exactly, while the subtree kernel considers all pairs of matching subsets of the neighborhoods of two vertices.
It is interesting to note that the Weisfeiler-Lehman subtree kernel exhibits a very attractive computational complexity since it can be computed in $\mathcal{O}(hm)$ time.

\subsection{Neighborhood Aggregation Approaches}
The Weisfeiler-Lehman subtree kernel triggered a lot of activity in the field of graph kernels.
The relabeling procedure of the Weisfeiler-Lehman algorithm can be viewed as a \textit{neighborhood aggregation} scheme.
The main idea behind neighborhood aggregation algorithms (a.k.a. message-passing algorithms) is that each vertex receives messages from its neighbors and utilizes these messages to update its representation.
Following the success of this kernel, several variations of it were proposed.
All these variations employ a neighborhood aggregation scheme similar to that of the Weisfeiler-Lehman algorithm.
The goal of most of these works is to speed-up the computation time of the Weisfeiler-Lehman subtree kernel \shortcite{hido2009linear,kataoka2016hadamard}.
However, other types of variations were also proposed such as a streaming version of the Weisfeiler-Lehman algorithm \shortcite{li2012nested}, a kernel that uses the $k$-dimensional Weisfeiler-Lehman test of isomorphism \shortcite{morris2017global}, and a method that augments the subtree features with topological information \shortcite{rieck2019persistent}.
We next present the neighborhood hash kernel, a kernel that was born out of these research efforts.

\subsubsection{Neighborhood Hash Kernel}
Similar to the Weisfeiler-Lehman subtree kernel, the neighborhood hash kernel also assumes node-labeled graphs \shortcite{hido2009linear}.
It compares graphs by updating their node labels and counting the number of common labels.
The kernel replaces the discrete node labels with binary arrays of fixed length, and it then employs logical operations to update the labels so that they contain information about the neighborhood structure of each vertex.

Let $\ell : V \rightarrow \Sigma$ be a function that maps vertices to an alphabet $\Sigma$ which is the set of possible discrete node labels.
Hence, given a vertex $v$, $\ell(v) \in \Sigma$ is the label of vertex $v$.
The algorithm first transforms each discrete node label to a bit label.
A bit label is a binary array consisting of $d$ bits as
\begin{equation}
    s = (b_1, b_2, \ldots, b_d)
\end{equation}
where the constant $d$ satisfies $2^d - 1 \gg |\Sigma|$ and $b_1, b_2, \ldots, b_d \in \{0, 1\}$.

The most important step of the algorithm involves a procedure that updates the labels of the vertices.
To achieve that, the kernel makes use of two very common bit operations: ($1$) the exclusive or ($XOR$) operation, and ($2$) the bit rotation ($ROT$) operation.
Let $XOR(s_i, s_j) = s_i \oplus s_j$ denote the $XOR$ operation between two bit labels $s_i$ and $s_j$ (\ie the $XOR$ operation is applied to all their components).
The output of the operation is a new binary array whose components represent the $XOR$ value between the corresponding components of the $s_i$ and $s_j$ arrays.
The $ROT_o$ operation takes as input a bit array and shifts its last $o$ bits to the left by $o$ bits and moves the first $o$ bits to the right end as shown below  
\begin{equation}
    ROT_o(s) = \{ b_{o+1}, b_{o+2}, \ldots, b_d, b_1, \ldots, b_o \}
\end{equation}
Below, we present in detail two procedures for updating the labels of the vertices: ($1$) the simple neighborhood hash, and ($2$) the count-sensitive neighborhood hash.

\begin{figure}[t]
  \centering
  \includegraphics[width=0.55\textwidth]{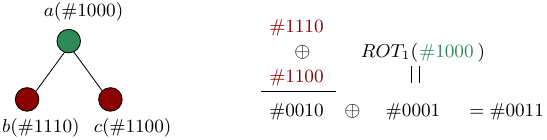}
  \caption{Example of computation of the simple neighborhood hash for a vertex (in green). The vertex has two adjacent vertices (in red). The three vertices have different labels from each other. The algorithm uses XOR and ROT operations to compute the neighborhood hash of the vertex ($\#0011$).}
  \label{fig:simple_nh}
\end{figure}

\paragraph{Simple Neighborhood Hash.}
Given a graph $G=(V,E)$ with bit labels, the simple neighborhood hash update procedure computes a neighborhood hash for each vertex using the logical operations $XOR$ and $ROT$.
More specifically, given a vertex $v \in V$, let $\mathcal{N}(v)=\{ u_1,\ldots,u_d \}$ be the set of neighbors of $v$.
Then, the kernel computes the neighborhood hash as
\begin{equation}
    NH(v) = ROT_1 \big( \ell(v) \big) \oplus \big( \ell(u_1) \oplus \ldots \oplus \ell(u_d) \big)
\end{equation}
The resulting hash $NH(v)$ is still a bit array of length $d$, and we regard it as the new label of $v$.
This new label represents the distribution of the node labels around $v$.
Hence, if $v_i$ and $v_j$ are two vertices that have the same label (\ie $\ell(v_i) = \ell(v_j)$) and the label sets of their neighborhors are also identical, their hash values will be the same (\ie $NH(v_i) = NH(v_j))$.
Otherwise, they will be different except for accidental hash collisions.
The main idea behind this update procedure is that the hash value is independent of the order of the neighborhood values due to the properties of the $XOR$ operation.
Hence, one can check whether or not the distributions of neighborhood labels of two vertices are equivalent without sorting or matching these two label sets.
Figure~\ref{fig:simple_nh} illustrates how the simple neighborhood hash is computed for a given vertex.

\begin{figure}[t]
  \centering
  \includegraphics[width=\textwidth]{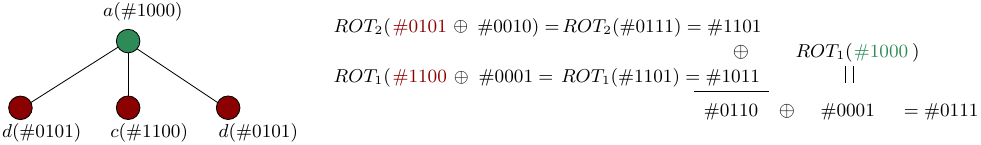}
  \caption{Example of computation of the count-sensitive neighborhood hash for a vertex (in green). The vertex has three adjacent vertices (in red). Two of these three vertices have identical labels. The algorithm uses XOR and ROT operations to compute the count-sensitive neighborhood hash of the vertex ($\#0111$).}
  \label{fig:count_sensitive_nh}
\end{figure}

\paragraph{Count-sensitive Neighborhood Hash.}
The simple neighborhood hash update procedure described above suffers from some problematic hash collisions.
Specifically, the neighborhood hash values for two independent nodes have a small probability of being the same even if there is no accidental hash collision.
Such problematic hash collisions may affect the positive semidefiniteness of the kernel.
To address that problem, the count-sensitive neighborhood hash update procedure counts the number of occurences of each label in the set.
More specifically, it first uses a sorting algorithm (\eg radix sort) to align the bit labels of the neighbors, and then, it extracts the unique labels (set $\{ \ell_1, \ldots, \ell_l \}$ in the case of $l$ unique labels) and for each label counts its number of occurences.
Then, it updates each unique label based on its number of occurences as follows
\begin{equation}
    \ell'_i = ROT_o \big( \ell_i \oplus o \big)
\end{equation}
where $\ell_i, \ell'_i$ is the initial and updated label respectively, and $o$ is the number of occurences of that label in the set of neighbors.
The above operation makes the hash values unique by depending on the number of label occurrences.
Then, the count-sensitive neighborhood hash is computed as
\begin{equation}
    CSNH(v) = ROT_1 \big( \ell(v) \big) \oplus \big( \ell'_1 \oplus \ldots \oplus \ell'_l \big)
\end{equation}
Figure~\ref{fig:count_sensitive_nh} illustrates the operations of the count-sensitive neighborhood hash for a given vertex.
Both the simple and the count-sensitive neighborhood hash can be seen as general approaches for enriching the labels of vertices based on the label distribution of their neighborhood vertices.

\paragraph{Kernel Calculation.}
The neighborhood hash update procedures presented above aggregate the information of the neighborhood vertices to each vertex.
Then, given two graphs $G$ and $G'$, the updated labels of their vertices are compared using the following function
\begin{equation}
    \kappa(G, G') = \frac{c}{|V| + |V'| - c}
\end{equation}
where $c$ is the number of labels the two graphs have in common.
This function is equivalent to the Tanimoto coefficent which is commonly used as a similarity measure between sets of discrete values and which has been proven to be positive semidefinite \shortcite{gower1971general}.

The label-update procedures is not necessary to be applied once, but they can be applied iteratively.
By updating the bit labels several times, the new labels can capture high-order relationships between vertices.
For instance, if the procedure is performed $h$ times in total, the updated label $\ell(v)$ of a vertex $v$ represents the label distribution of its $h$-neighbors.
Hence, two vertices $v_i, v_j$ with identical labels and connections among their $r$-neighbors will be assigned the same label.

\begin{definition}[Neighborhood Hash Kernel]
	Let $G$ and $G'$ be two graphs, and let $G_1, \ldots, G_h$ and $G_1', \ldots, G_h'$ denote their updated graphs where the node labels have been updated $1,\ldots,h$ times based on one of the two procedures presented above, respectively.
	Then, the neighborhood hash kernel is defined as
	\begin{equation}
		k(G, G') = \frac{1}{h} \sum_{i=1}^h \kappa(G_i, G'_i)
	\end{equation}
\end{definition}
The computational complexity of the neighborhood hash kernel is $\mathcal{O}(\overline{deg} \, nhd)$ where $n=|V|$ is the number of vertices of the graphs and $\overline{deg}$ is the average degree of their vertices.

\subsection{Other Approaches}
Recently, several kernels were proposed that belong to the $R$-convolution framework, but do not perform neighborhood aggregation.
There are, for instance, kernels specially designed for graphs with ordered neighborhoods \shortcite{draief2018kong}, kernels that compare pairs of rooted subgraphs containing vertices up to a certain distance from the root \shortcite{costa2010fast}, kernels that extract directed acyclic graphs from the input graphs \shortcite{da2012tree}, and kernels that use the orthonormal representations of vertices introduced by Lov\'asz \shortcite{johansson2014global}.
We next present some of these kernels in detail.

\subsubsection{Neighborhood Subgraph Pairwise Distance Kernel}
The neighborhood subgraph pairwise distance kernel extracts pairs of rooted subgraphs from each graph whose roots are located at a certain distance from each other, and which contain vertices up to a certain distance from the root.
It then compares graphs based on these pairs of rooted subgraphs.
To avoid isomorphism checking, graph invariants are employed to encode each rooted subgraph \shortcite{costa2010fast}.

Let $G=(V,E)$ be a graph.
The distance between two vertices $u,v \in V$, denoted $D(u,v)$, is the length of the shortest path between them.
The neighborhood of radius $r$ of a vertex $v$ is the set of vertices at a distance less than or equal to $r$ from $v$, that is $\{ u \in V : D(u,v) \leq r\}$.
Given a subset of vertices $S \subseteq V$, let $E(S)$ be the set of edges that have both end-points in $S$.
Then, the subgraph with vertex set $S$ and edge set $E(S)$ is known as the subgraph induced by $S$.
The neighborhood subgraph of radius $r$ of vertex $v$ is the subgraph induced by the neighborhood of radius $r$ of $v$ and is denoted by $\mathcal{N}_r(v)$.
Let also $R_{r,d}(A_v,B_u,G)$ be a relation between two rooted graphs $A_v$, $B_u$ and a graph $G=(V,E)$ that is true if and only if both $A_v$ and $B_u$ are in $\{\mathcal{N}_r(v) : v \in V \}$, where we require $A_v, B_u$ to be isomorphic to some $\mathcal{N}_r(v)$ to verify the set inclusion, and that $D(u,v) = d$.
We denote with $R^{-1}(G)$ the inverse relation that yields all the pairs of rooted graphs $A_v$, $B_u$ satisfying the above constraints.
Hence, $R^{-1}(G)$ selects all pairs of neighborhood graphs of radius $r$ whose roots are at distance $d$ in a given graph $G$.

\begin{definition}[Neighborhood Subgraph Pairwise Distance Kernel]
	Let $G,G'$ be two graphs.
	The neighborhood subgraph pairwise distance kernel extracts from the two graphs pairs of rooted subgraphs of radius $r$ whose roots are located at distance $d$ from each other.
	It then utilizes the following kernel to compare them
	\begin{equation}
	    k_{r,d}(G, G') = \sum_{A_v, B_v \in R_{r,d}^{-1}(G)} \ \sum_{A'_{v'}, B'_{v'} \in R_{r,d}^{-1}(G')} \delta(A_v, A'_{v'}) \ \delta(B_v, B'_{v'})
	\end{equation}
	where $\delta$ is $1$ if its input subgraphs are isomorphic, and $0$ otherwise.
	The above kernel counts the number of identical pairs of neighboring graphs of radius $r$ at distance $d$ between two graphs.
	Then, the neighborhood subgraph pairwise distance kernel is defined as
	\begin{equation}
	    k(G, G') = \sum_{r=0}^{r^*} \sum_{d=0}^{d^*} \hat{k}_{r,d}(G, G')
	\end{equation}
	where $\hat{k}_{r,d}$ is a normalized version of $k_{r,d}$, that is
	\begin{equation}
	    \hat{k}_{r,d}(G,G') = \frac{k_{r,d}(G,G')}{\sqrt{k_{r,d}(G,G) k_{r,d}(G',G')}}
	\end{equation}
\end{definition}
The above version ensures that relations of all orders are equally weighted regardless of the size of the induced part sets.
The neighborhood subgraph pairwise distance kernel includes an exact matching kernel over two graphs (\ie the $\delta$ kernel) which is equivalent to solving the graph isomorphism problem.
Solving the graph isomorphism problem is not feasible.
Therefore, the kernel produces an approximate solution to it instead.
Given a subgraph $G_S$ induced by the set of vertices $S$, the kernel computes a graph invariant encoding for the subgraph via a label function $\ell^g : \mathcal{G} \rightarrow \Sigma^*$, where $\mathcal{G}$ is the set of rooted graphs and $\Sigma^*$ is the set of strings over a finite alphabet $\Sigma$.
The function $\ell^g$ makes use of two other label functions: ($1$) a function $\ell^n$ for vertices, and ($2$) a function $\ell^e$ for edges.
The $\ell^n$ function assigns to vertex $v$ the concatenation of the lexicographically sorted list of triplets $\langle D(v,u), D(v,h), \ell(u) \rangle$ for all $u \in S$, where $h$ is the root of the subgraph and $\ell$ is a function that maps vertices/edges to their label symbol.
Hence, the above function relabels each vertex with a string that encodes the initial label of the vertex, the vertex distance from all other labeled vertices, and the distance from the root vertex.
The $\ell^e \big( (u,v) \big)$ function assigns to edge $(u,v)$ the label $\langle \ell^n(u)$, $\ell^n(v)$, $\ell \big( (u,v) \big) \rangle$.
The $\ell^e \big( (u,v) \big)$ function thus annotates each edge based on the new labels of its endpoints, and its initial label, if any.
Finally, the function $\ell^g(G_S)$ assigns to the rooted graph induced by $S$ the concatenation of the lexicographically sorted list of $\ell^e \big( (u,v) \big)$ for all $\{u,v\} \in E(S)$.
The kernel then employs a hashing function from strings to natural numbers $H : \Sigma^* \rightarrow \mathbb{N}$ to obtain a unique identifier for each subgraph.
Hence, instead of testing pairs of subgraphs for isomorphism, the kernel just checks if the subgraphs share the same identifier.

The computational complexity of the neighborhood subgraph pairwise distance kernel is $\mathcal{O}(n |S| |E(S)| \log |E(S)|)$ and is dominated by the repeated computation of the graph invariant for each vertex of the graph.
Since this is a constant time procedure, for small values of $d^*$ and $r^*$, the complexity of the kernel is in practice linear in the size of the graph.

\subsubsection{Lov\'asz $\vartheta$ Kernel}
The Lov\'asz number $\vartheta(G)$ of a graph $G=(V,E)$ is a real number that is an upper bound on the Shannon capacity of the graph.
It was introduced by L\'aszl\'o Lov\'asz in $1979$ \shortcite{lovasz1979shannon}.
The Lov\'asz number is intimately connected with the notion of orthonormal representations of graphs.
An orthonormal representation of a graph $G$ consists of a set of unit vectors $U_G = \{ u_i \in \mathbb{R}^d : || u_i || = 1 \}_{i \in V}$ where each vertex $i$ is assigned a unit vector $u_i$ such that $(i,j) \not \in E \implies u_i^\top u_j = 0$.
Specifically, the Lov\'asz number of a graph $G$ is defined as
\begin{equation}
    \vartheta(G) = \min_{c, U_G} \max_{i \in V} \frac{1}{(c^\top u_i)^2}
\end{equation}
where $c \in \mathbb{R}^d$ is a unit vector and $U_G$ is an orthonormal representation of $G$. 
Geometrically, $\vartheta(G)$ is defined by the smallest cone enclosing a valid orthonormal representation $U_G$.
The Lov\'asz number $\vartheta(G)$ of a graph $G$ can be computed to arbitrary precision in polynomial time by solving a semidefinite program.

The Lov\'asz $\vartheta$ kernel utilizes the orthonormal representations associated with the Lov\'asz number to compare graphs \shortcite{johansson2014global}.
The kernel is applicable only to unlabeled graphs.
Given a collection of graphs, it first generates orthonormal representations for the vertices of each graph by computing the Lov\'asz $\vartheta$ number.
Hence, $U_G$ is a set that contains the orthonormal representations of $G$.
Let $S \subseteq V$ be a subset of the vertex set of $G$.
Then, the Lov\'asz value of the set of vertices $S$ is defined as
\begin{equation}
    \vartheta_S(G) = \min_{c} \max_{i \in S} \frac{1}{(c^\top u_i)^2}
\end{equation}
where $c \in \mathbb{R}^d$ is a unit vector and $u_i$ is the representation of vertex $i$ obtained by computing the Lov\'asz number $\vartheta(G)$ of $G$.
The Lov\'asz value of a set of vertices $S$ represents the angle of the smallest cone enclosing the set of orthonormal representations of these vertices (\ie subset of $U_G$ defined as $\{ u_i : u_i \in U_G, i \in S \}$).

\begin{definition}[Lov\'asz $\vartheta$ Kernel]
	Let $G=(V,E)$ and $G'=(V',E')$ be two graphs.
	The Lov\'asz $\vartheta$ kernel between the two graphs is defined as follows
	\begin{equation}
	    k(G, G') = \sum_{S \subseteq V} \sum_{S' \subseteq V'} \delta(|S|, |S'|) \ \frac{1}{Z_{|S|}} \ k \big( \vartheta_S(G), \vartheta_{S'}(G') \big)
	\end{equation}
	where $Z_{|S|} = \binom{|V|}{|S|} \binom{|V'|}{|S|}$, $\delta(|S|, |S'|)$ is a delta kernel (equal to $1$ if $|S|=|S'|$, and $0$ otherwise), and $k$ is a positive semi-definite kernel between Lov\'asz values (\eg linear kernel, gaussian kernel).
\end{definition}
The Lov\'asz $\vartheta$ kernel consists of two main steps: ($1$) computing the Lov\'asz number $\vartheta$ of each graph and obtaining the associated orthonormal representations, and ($2$) computing the Lov\'asz value for all subgraphs (\ie subsets of vertices $S \subseteq V$) of each graph.
Exact computation of the Lov\'asz $\vartheta$ kernel is in most real settings infeasible since it requires computing the minimum enclosing cones of $2^n$ sets of vertices.

When dealing with large graphs, it is thus necessary to resort to sampling.
Given a graph $G$, instead of evaluating the Lov\'asz value on all $2^n$ sets of vertices, the algorithm evaluates it in on a smaller number of subgraphs induced by sets of vertices contained in $\mathcal{L} \subset 2^V$.
Then, the Lov\'asz $\vartheta$ kernel is defined as follows
\begin{equation}
    \hat{k}(G, G') = \sum_{S \in \mathcal{L}} \sum_{S' \in \mathcal{L}'} \delta(|S|, |S'|) \ \frac{1}{\hat{Z}_{|S|}} \ k \big( \vartheta_S(G), \vartheta_{S'}(G') \big)
\end{equation}
where $\hat{Z}_{|S|} = |\mathcal{L}_{|S|}| |\mathcal{L}'_{|S|}|$ and $\mathcal{L}_{|S|}$ denotes the subset of $\mathcal{L}$ consisting of all sets of cardinality $|S|$, that is $\mathcal{L}_{|S|} = \{ B \in \mathcal{L} : |B| = |S| \}$.

The time complexity of computing $\hat{k}(G, G')$ is $\mathcal{O}(n^2 m \epsilon^{-1} + s^2 T(k) + sn)$ where $T(k)$ is the complexity of computing the base kernel $k$, $n = |V|$, $m = |E|$ and $s = \max(|\mathcal{L}|, |\mathcal{L}'|)$.
The first term represents the cost of solving the semi-definite program that computes the Lov\'asz number $\vartheta$.
The second term corresponds to the worst-case complexity of computing the sum of the Lov\'asz values.
And finally, the third term is the cost of computing the Lov\'asz values of the sampled subsets of vertices.

\subsubsection{SVM-$\vartheta$ Kernel}
The SVM-$\vartheta$ kernel is closely related to the Lov\'asz $\vartheta$ kernel \shortcite{johansson2014global}.
The Lov\'asz $\vartheta$ kernel suffers from high computational complexity, and the SVM-$\vartheta$ kernel was developed as a more efficient alternative. 
Similar to the Lov\'asz $\vartheta$ kernel, this kernel also assumes unlabeled graphs.

Given a graph $G=(V,E)$ such that $|V| = n$, the Lov\'asz number of $G$ can be defined as
\begin{equation}
    \vartheta(G) = \min_{K \in L} \omega(K)
\end{equation}
where $\omega(K)$ is the one-class SVM given by
\begin{equation}
    \label{eq:oneclass_svm}
    \omega(K) = \max_{\alpha_i > 0} 2\sum_{i=1}^{n} \alpha_i - \sum_{i=1}^{n} \sum_{j=1}^{n} \alpha_i \alpha_j K_{ij}
\end{equation}
and $L$ is a set of positive semidefinite matrices defined as
\begin{equation}
    L = \{ K \in S_{n}^+ : K_{ii} = 1, K_{ij}=0 \: \forall (i,j) \not \in E \}
\end{equation}
where $S_{n}^+$ is the set of all $n \times n$ positive semidefinite matrices.

The SVM-$\vartheta$ kernel first computes the matrix $K_{LS}$ which is equal to
\begin{equation}
    K_{LS} = \frac{A}{\rho} + I
\end{equation}
where $A$ is the adjacency matrix of $G$, $I$ is the $n \times n$ identity matrix, and $\rho \geq -\lambda_n$ with $\lambda_n$ the minimum eigenvalue of $A$.
The matrix $K_{LS}$ is positive semidefinite by construction and it has been shown in \shortcite{jethava2013lovasz} that
\begin{equation}
    \omega(K_{LS}) = \sum_{i=1}^n \alpha_i
\end{equation}
where $\alpha_i$ are the maximizers of Equation~\eqref{eq:oneclass_svm}. 
Furthermore, it was shown that on certain families of graphs (\eg Erd{\"o}s R{\'e}nyi random graphs), $\omega(K_{LS})$ is with high probability a constant factor approximation to $\vartheta(G)$.

\begin{definition}[SVM-$\vartheta$ Kernel]
	Let $G=(V,E)$ and $G'=(V',E')$ be two graphs.
	Then, the SVM-$\vartheta$ kernel is defined as follows
	\begin{equation}
	    k(G, G') = \sum_{S \subseteq V} \sum_{S' \subseteq V'} \delta(|S|, |S'|) \ \frac{1}{Z_{|S|}} \ k \Big(\sum_{i \in S} \alpha_i, \sum_{j \in S'} \alpha_j \Big)
	\end{equation}
	where $Z_{|S|} = \binom{|V|}{|S|} \binom{|V'|}{|S|}$, $\delta(|S|, |S'|)$ is a delta kernel (equal to $1$ if $|S|=|S'|$, and $0$ otherwise), and $k$ is a positive semi-definite kernel between real values (\eg linear kernel, gaussian kernel).
\end{definition}

The SVM-$\vartheta$ kernel consists of three main steps: ($1$) constructing matrix $K_{LS}$ of $G$ which takes $\mathcal{O}(n^3)$ time ($2$) solving the one-class SVM problem in $\mathcal{O}(n^2)$ time to obtain the $\alpha_i$ values, and ($3$) computing the sum of the $\alpha_i$ values for all subgraphs (\ie subsets of vertices $S \subseteq V$) of each graph.
Computing the above quantity for all $2^n$ sets of vertices is not feasible in real-world scenarios.

To address the above issue, the SVM-$\vartheta$ kernel employs sampling schemes.
Given a graph $G$, the kernel samples a specific number of subgraphs induced by sets of vertices contained in $\mathcal{L} \in 2^V$.
Then, the SVM-$\vartheta$ kernel is defined as follows
\begin{equation}
    \hat{k}(G, G') = \sum_{S \in \mathcal{L}} \sum_{S' \in \mathcal{L}'} \delta(|S|, |S'|) \ \frac{1}{\hat{Z}_{|S|}} \ k \Big(\sum_{i \in S} \alpha_i, \sum_{j \in S'} \alpha_j \Big)
\end{equation}
where $\hat{Z}_{|S|} = |\mathcal{L}_{|S|}| |\mathcal{L}'_{|S|}|$ and $\mathcal{L}_{|S|}$ denotes the subset of $\mathcal{L}$ consisting of all sets of cardinality $|S|$, that is $\mathcal{L}_{|S|} = \{ B \in \mathcal{L} : |B| = |S| \}$.

The time complexity of computing $\hat{k}(G, G')$ is $\mathcal{O}(n^3 + s^2 T(k) + sn)$ where $T(k)$ is the complexity of computing the base kernel $k$ and $s = \max(|\mathcal{L}|, |\mathcal{L}'|)$.
The first term represents the cost of computing $K_{LS}$ (dominated by the eigenvalue decomposition).
The second term corresponds to the worst-case complexity of comparing the sums of the $\alpha_i$ values.
And finally, the third term is the cost of computing the sum of the $\alpha_i$ values for the sampled subsets of vertices.

\subsubsection{Ordered Decomposition DAGs Kernel}
In contrast to the above two kernels, the ordered decomposition DAGs kernel can handle node-labeled graphs.
The kernel decomposes graphs into multisets of directed acyclic graphs (DAGs), and then uses existing tree kernels to compare these DAGs \shortcite{da2012tree}.

\begin{figure}[t]
  \centering
  \includegraphics[width=\textwidth]{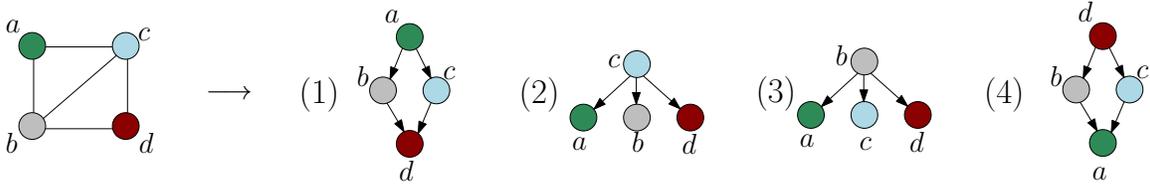}
  \caption{Example of decomposition of a graph into its four DAGs (one for each vertex).}
  \label{fig:odd_1}
\end{figure}

\begin{figure}[t]
  \centering
  \includegraphics[width=0.55\textwidth]{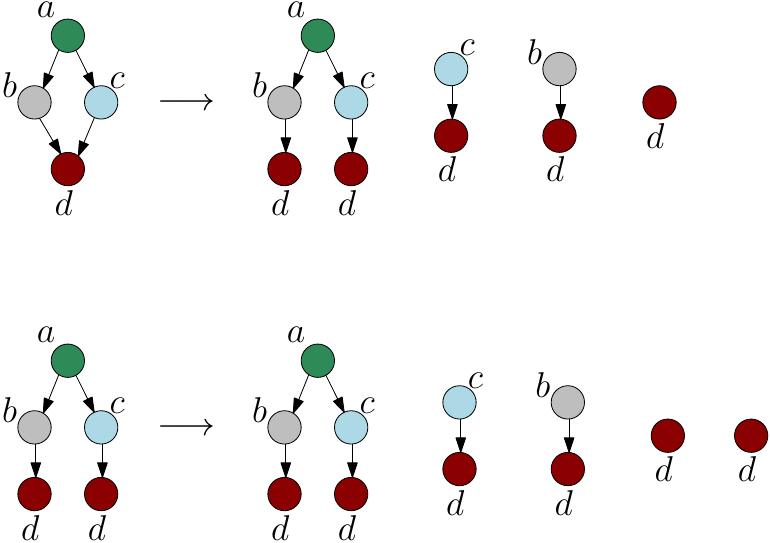}
  \caption{Two DAGs (left) and their associated tree visits $T(u)$ starting from each vertex $u$ (right).}
  \label{fig:odd_2}
\end{figure}

Given a graph $G=(V,E)$, the kernel generates one unordered rooted DAG, say $DD_v$, for each vertex $v \in V$.
To generate the DAG, the kernel keeps only those edges belonging to the shortest paths between $v$ and any vertex $u \in V \setminus \{v\}$.
Furthermore, a direction is given to each edge, while edges connecting vertices visited at level $l$ to vertices visited at level $l' < l$ are also removed.
Figure~\ref{fig:odd_1} gives an example of the decomposition of a graph into a set of DAGs.
\begin{definition}[Ordered Decomposition DAGs Kernel]
  Let $G=(V,E)$ and $G'=(V',E')$ be two graphs.
  Let also $DD(G)$ and $DD(G')$ be multisets defined as $\{ DD_v : v \in V\}$ and $\{ DD_{v'} : {v'} \in V'\}$, respectively.
  Then, the ordered decomposition DAGs kernel is defined as
  \begin{equation}
    k(G, G') = \sum_{D \in DD(G)} \ \sum_{D' \in DD(G')} k_{DAG}(D, D')
  \end{equation}
  where $k_{DAG}$ is a kernel between DAGs.
\end{definition}
The kernel is thus defined as the sum of the computation of a local kernel for DAGs, over all pairs of DAGs in the multiset. 
Note that these DAGs are unordered.
Moreover, there is a large literature on kernels for ordered trees, but only a few kernel functions for unordered trees.
Hence, the ordered decomposition DAGs kernel transforms the unordered DAGs to ordered DAGs, and then applies a kernel for ordered trees.
More specifically, the kernel defines a strict partial order among the vertices of each DAG.
This partial order takes into account the labels of the vertices, the outdegrees of the vertices (in case of identical node labels), and the relation between the sequence of successors of each vertex (in case of identical node labels and equal outdegrees).
Let $ODD_v$ denote the DAG of $v \in V$ ordered according to the above relation.
Let a tree visit be a function $T(u)$ that, given a vertex $u$ of a $ODD_v$, returns the tree resulting from the visit of the DAG starting in $u$.
Figure~\ref{fig:odd_2} gives an example of tree visits.
Then, the ordered decomposition DAGs kernel uses tree visits to project sub-DAGs to a tree space and applies tree kernels on the visits
\begin{equation}
  k_{DAG}(D, D') = \sum_{v \in V_D} \sum_{v' \in V_{D'}} k_{tree} \big( root(T(v)),root(T(v')) \big)
\end{equation}
where $V_D$, $V_{D'}$ are the set of vertices of $D$ and $D'$, respectivevly, and $k_{tree}$ is a kernel between ordered trees.
The time complexity of the ordered decomposition DAGs kernel depends on the employed tree kernel $k_{tree}$.
For instance, using the subtree and subset tree kernel leads to a time complexity of $\mathcal{O}(n^3 \log n)$ and $\mathcal{O}(n^4)$, respectively.
To reduce the time complexity, the kernel employs a strategy that allows it to compute $k_{tree}$ once for each unique pair of subtrees appearing in different DAGs.
Furthermore, in case the subtree kernel is employed, some other strategies can be applied to speed up the computation such as for instance limiting the depth of the visits during the generation of the multiset of DAGs

\subsection{Assignment Kernels}
The majority of kernels presented so far belong to the family of $R$-convolution kernels.
Besides this family of kernels, another family that has received a lot of attention recently is that of \textit{assignment kernels}.
In general, these kernels compute a matching between substructures of one object and substructures of a second object such that the overall similarity of the two objects is maximized \shortcite{frohlich2005optimal,schiavinato2015transitive,bai2015graph,bai2015aligned,kriege2016valid,nikolentzos2017matching,togninalli2019wasserstein}.
Such a matching can reveal structural correspondences between the two objects.
However, defining valid graph kernels that follow this design paradigm is not trivial.
For example, an optimal assignment kernel that was proposed in the early days of graph kernels to compute a correpondence between the atoms of molecules \shortcite{frohlich2005optimal} was later proven not to always be positive semidefinite \shortcite{vert2008optimal}.
Despite these design difficulties, there is a handful of valid assignment graph kernels.
For instance, there is a method that capitalizes on the well-known pyramid match kernel to match the node embeddings of graphs \shortcite{nikolentzos2017matching}, while another approach uses multi-graph matching techniques to obtain valid assignment kernels \shortcite{schiavinato2015transitive}.
More importantly, it was recently shown that there exists a class of base kernels used to compare substructures that guarantees positive semidefinite optimal assignment kernels \shortcite{kriege2016valid}.
We next present some of the above instances of assignment kernels in detail.

\subsubsection{Pyramid Match Graph Kernel}
The pyramid match kernel is a very popular algorithm in Computer Vision, and has proven useful for many applications including object recognition and image retrieval \shortcite{grauman2007pyramid,lazebnik2006beyond}.
The pyramid match graph kernel extends its applicability to graph-structured data \shortcite{nikolentzos2017matching}.
The kernel can handle unlabeled graphs as well as graphs that contain discrete node labels.

The pyramid match graph kernel first embeds the vertices of each graph into a low-dimensional vector space using the eigenvectors of the $d$ largest in magnitude eigenvalues of the graph's adjacency matrix.
Since the signs of these eigenvectors are arbitrary, it replaces all their components by their absolute values.
Each vertex is thus a point in the $d$-dimensional unit hypercube.
To find an approximate correspondence between the sets of vertices of two graphs, the kernel maps these points to multi-resolution histograms, and compares the emerging histograms with a weighted histogram intersection function.

Initially, the kernel partitions the feature space into regions of increasingly larger size and takes a weighted sum of the matches that occur at each level.
Two points match with each other if they fall into the same region.
Matches made within larger regions are weighted less than those found in smaller regions.
The kernel repeatedly fits a grid with cells of increasing size to the $d$-dimensional unit hypercube.
Each cell is related only to a specific dimension and its size along that dimension is doubled at each iteration, while its size along the other dimensions stays constant and equal to $1$.
Given a sequence of levels from $0$ to $L$, then at level $l$, the $d$-dimensional unit hypercube has $2^l$ cells along each dimension and $D = 2^{l}d$ cells in total.
Given a pair of graphs $G,G'$, let $H_G^l$ and $H_{G'}^l$ denote the histograms of $G$ and $G'$ at level $l$, and $H_G^l(i)$, $H_{G'}^l(i)$, the number of vertices of $G$, $G'$ that lie in the $i$-th cell.
The number of points in two sets which match at level $l$ is then computed using the histogram intersection function
\begin{equation}
  I(H_G^l,H_{G'}^l) = \sum_{i=1}^D \min\big(H_G^l(i),H_{G'}^l(i)\big)
\end{equation}
The matches that occur at level $l$ also occur at levels $0, \ldots, l-1$.
The algorithm takes into account only the new matches found at each level which is given by $I(H_{G_1}^l,H_{G_2}^l) - I(H_{G_1}^{l+1},H_{G_2}^{l+1})$ for $l=0,\ldots,L-1$.
Furthermore, the number of new matches found at each level in the pyramid is weighted according to the size of that level's cells.
Matches found within smaller cells are weighted more than those that occur in larger cells.
Specifically, the weight for level $l$ is set equal to $\nicefrac{1}{2^{L-l}}$.
Hence, the weights are inversely proportional to the length of the side of the cells that varies in size as the levels increase.

\begin{definition}[Pyramid Match Graph Kernel]
  Let $G=(V,E)$ and $G'=(V',E')$ be two graphs.
	The pyramid match kernel is defined as follows
	\begin{equation}
	  k(G,G') = I(H_G^L,H_{G'}^L) + \sum_{l=0}^{L-1} \frac{1}{2^{L-l}}\big(I(H_G^l,H_{G'}^l) - I(H_G^{l+1},H_{G'}^{l+1})\big)
	\end{equation}
	where $L$ is the number of different levels.
\end{definition}
The complexity of the pyramid match kernel is $\mathcal{O}(dnL)$ where $n$ is the number of vertices of the graphs under comparison.

In the case of labeled graphs, the kernel restricts matchings to occur only between vertices that share same labels.
It represents each graph as a set of sets of vectors, and matches pairs of sets of two graphs corresponding to the same label using the pyramid match kernel.
The emerging kernel for labeled graphs corresponds to the sum of the separate kernels
\begin{equation}
    k(G, G') = \sum_{i=1}^{|\Sigma|} k^i(G,G')
\end{equation}
where $|\Sigma|$ is the number of distinct labels and $k^i(G,G')$ is the pyramid match kernel between the sets of vertices of the two graphs which are assigned the label $i$.

\subsubsection{Weisfeiler-Lehman Optimal Assignment Kernel}
The Weisfeiler-Lehman optimal assignment kernel is currently a state-of-the-art approach for learning on graphs \shortcite{kriege2016valid}.
The kernel capitalizes on the theory of valid assignment kernels to improve the performance of the Weisfeiler-Lehman subtree kernel.
Before we delve into the details of the kernel, it is necessary to introduce the theory of valid optimal assignment kernels.

Let $\mathcal{X}$ be a set, and $[\mathcal{X}]^n$ denote the set of all $n$-element subsets of $\mathcal{X}$.
Let also $X,X' \in [\mathcal{X}]^n$ for $n \in \mathbb{N}$, and $\mathfrak{B}(X,X')$ denote the set of all bijections between $X$ and $X'$.
The optimal assignment kernel on $[\mathcal{X}]^n$ is defined as
\begin{equation}\label{eq:valid_assignment_kernel}
  K_\mathfrak{B}^k(X,X') = \max_{B \in \mathfrak{B}(X,X')} \sum_{(x,x') \in B} k(x,x')
\end{equation}
where $k$ is a kernel between the elements of $X$ and $X'$.
\shortciteA{kriege2016valid} showed that the above function $K_\mathfrak{B}(\mathcal{X},\mathcal{X}')$ is a valid kernel only if the base kernel $k$ is strong.
\begin{definition}[Strong Kernel]
  A function $k : \mathcal{X} \times \mathcal{X} \rightarrow \mathbb{R}_{\geq 0}$ is called strong kernel if $k(x,y) \geq \min\{ k(x,z),k(z,y) \}$ for all $x,y,z \in \mathcal{X}$.
\end{definition}
Strong kernels are equivalent to kernels obtained from a hierarchy defined on set $\mathcal{X}$.
More specifically, let $T$ be a rooted tree such that the leaves of $T$ are the elements of $\mathcal{X}$.
Let $V(T)$ be the set of vertices of $T$.
Each inner vertex $v \in T$ corresponds to a subset of $\mathcal{X}$ comprising all leaves of the subtree rooted at $v$.
Let $w : V(T) \rightarrow \mathbb{R}_{\geq 0}$ be a weight function such that $w(v) \geq w(p(v))$ for all $v$ in $T$ where $p(v)$ is the parent of vertex $v$.
Then, the tuple $(T,w)$ defines a hierarchy.
Let $LCA(u,v)$ be the lowest common ancestor of vertices $u$ and $v$, that is, the unique vertex with maximum depth that is an ancestor of both $u$ and $v$.
\begin{definition}[Hierarchy-induced Kernel]
  Let $H = (T,w)$ be a hierarchy on $\mathcal{X}$, then the function defined as $k(x,y) = w(LCA(x,y))$ for all $x,y$ in $\mathcal{X}$ is the kernel on $\mathcal{X}$ induced by $H$.
\end{definition}
Interestingly, strong kernels are equivalent to kernels obtained from a hierarchical partition of the domain of the kernel. 
Hence, by constructing a hierarchy on $\mathcal{X}$, we can derive a strong kernel $k$ and ensure that the emerging assignment function is a valid kernel.

\begin{figure}[t]
  \centering
  \includegraphics[width=\linewidth]{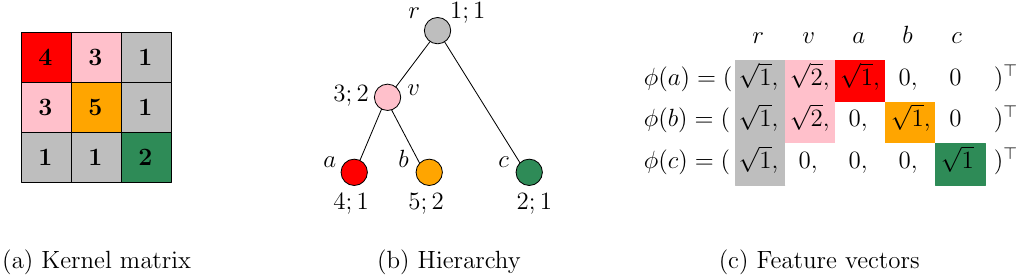}
    \caption{The matrix of a strong kernel on objects $a,b$ and $c$ (a) induced by the hierarchy (b) and the derived feature vectors (c). A vertex $v$ in (b) is annotated by its weights $w(v);\omega(v)$.}
    \label{fig:optimal_assignment_example}
\end{figure}

Based on the property that every strong kernel is induced by a hierarchy, we can derive explicit feature maps for strong kernels.
Let $\omega : V(T) \rightarrow \mathbb{R}_{\geq 0}$ be an additive weight function defined as $\omega(v) = w(v) - w(p(v))$ and $\omega(r) = w(r)$ for the root $r$.
Note that the property of a hierarchy assures that the values of the $\omega$ function are nonnegative.
For $v \in V(T)$, let $P(v) \subseteq V(T)$ denote the vertices on the path from $v$ to the root $r$.
The strong kernel $k$ induced by the hierarchy $H$ can be defined using the mapping $\phi : \mathcal{X} \rightarrow \mathbb{R}^n$, where $n = |V(T)|$ and the components indexed by $v \in V(T)$ are
\begin{equation}
  \phi(v) = \left\{
  \begin{array}{lr}
    \sqrt{\omega(u)} & \text{if }u \in P(v),\\
    \quad 0 & \text{otherwise}
  \end{array}
\right.
\end{equation}
Figure~\ref{fig:optimal_assignment_example} shows an example of a strong kernel, an associated hierarchy and the derived feature vectors.

\begin{figure}[t]
  \centering
  \includegraphics[width=.8\linewidth]{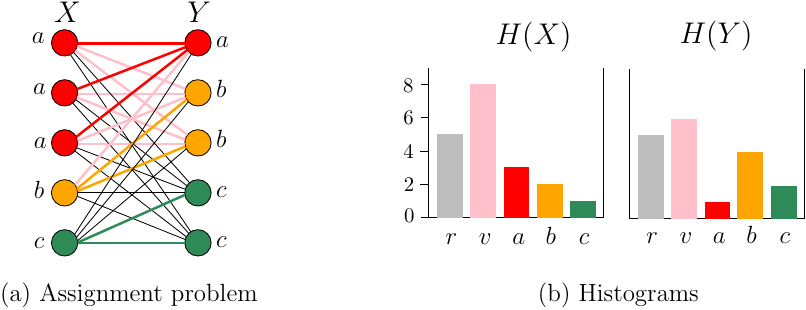}
  \caption{An assignment instance (a) for $X,Y \in [\mathcal{X}]^5$ and the derived histograms (b). The set $X$ contains three distinct vertices labeled $a$ and the set $Y$ two distinct vertices labeled $b$ and $c$. Taking the multiplicities into account the histograms are obtained from the hierarchy of the base kernel $k$ depicted in Figure~\ref{fig:optimal_assignment_example}.  The optimal assignment yields a value of $K_\mathfrak{B}^k(X, Y) = \sum_{i=1}^n \min\big(H_{X}(i),H_{Y}(i)\big) = \min\{5,5\}+ \min\{8,6\}+ \min\{3,1\}+ \min\{2,4\}+ \min\{1,2\}=15$.}
  \label{fig:optimal_assignment_histograms}
\end{figure}

Let $H = (T,w)$ be a hierarchy on $\mathcal{X}$.
As mentioned above, the hierarchy $H$ induces a strong kernel $k$.
Since $k$ is strong, the function $K_\mathfrak{B}^k$ defined in Equation~\ref{eq:valid_assignment_kernel} is a valid kernel.
The kernel $K_\mathfrak{B}^k$ can be computed in linear time in the number of vertices $n$ of the tree $T$ using the histogram intersection kernel \shortcite{swain1991color} as follows
\begin{equation}
  K_\mathfrak{B}^k(X, X') = \sum_{i=1}^n \min\big(H_{X}(i),H_{X'}(i)\big)
\end{equation}
which is known to be a valid kernel on $\mathbb{R}^n$ \shortcite{barla2003histogram}.
Hence, the complexity of the proposed kernel depends on the size of the tree $T$.
Figure~\ref{fig:optimal_assignment_histograms} illustrates the relation between the optimal assignment kernel employing a strong base kernel and the histogram intersection kernel.

\begin{figure}[t]
  \centering
  \includegraphics[width=.9\linewidth]{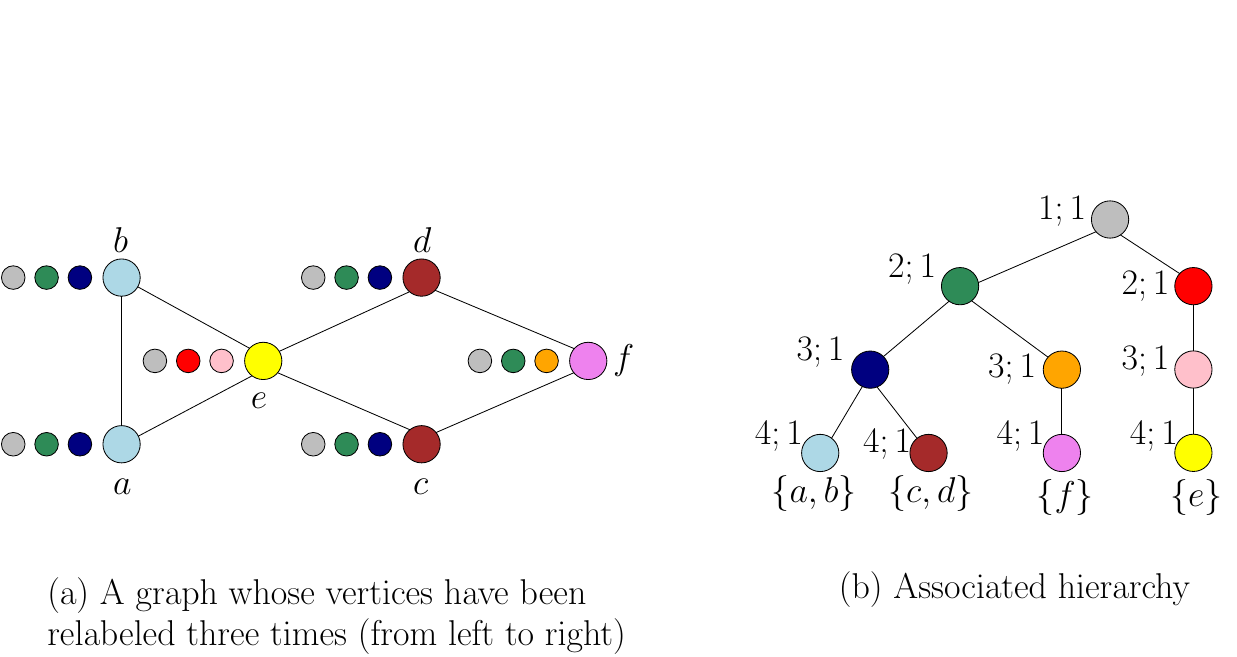}
    \caption{A graph $G$ with uniform initial labels $\tau_0$ and refined labels $\tau_i$ for $i \in \{1,\ldots,3\}$ (a), and the associated hierarchy (b).}
    \label{fig:wl_optimal_assignment}
\end{figure}

We next present the Weisfeiler-Lehman optimal assignment kernel.
\begin{definition}[Weisfeiler-Lehman Optimal Assignment Kernel]
  Let $G=(V,E)$ and $G'=(V',E')$ be two graphs.
  The Weisfeiler-Lehman optimal assignment kernel is defined as
  \begin{equation}
    k(G,G') = K_\mathfrak{B}^k(V,V')
  \end{equation}
  where $k$ is the following base kernel
  \begin{equation}
    k(v,v') = \sum_{i=0}^h \delta(\tau_i(v), \tau_i(v'))
  \end{equation}
  where $\tau_i(v)$ is the label of node $v$ at the end of the $i$-th iteration of the Weisfeiler-Lehman relabeling procedure.
\end{definition}
The base kernel value reflects to what extent two vertices $v$ and $v'$ have a similar neighborhood.
It can be shown that the colour refinement process of the Weisfeiler-Lehman algorithm defines a hierarchy on the set of all vertices of the input graphs.
Specifically, the sequence $(\tau_i)_{0\leq i \leq h}$ gives rise to a family of nested subsets, which can naturally be represented by a hierarchy $(T,w)$.
When assuming $\omega(v) = 1$ for all vertices $v \in V(T)$, the hierarchy induces the kernel defined above.
Such a hierarchy for a graph on six vertices is illustrated in Figure~\ref{fig:wl_optimal_assignment}.

\subsection{Kernels for Graphs with Continuous Attributes}
Most existing graph kernels are designed to operate on both unlabeled and node-labeled graphs.
However, many real-world graphs contain continuous real-valued node attributes.
One example comes from the field of cybersecurity where the function call graphs extracted from the source code of application programs typically contain multi-dimensional node labels. 
Such types of graphs do not appear only in cybersecurity, but also in computer vision \shortcite{harchaoui2007image} or even in bioinformatics \shortcite{borgwardt2005protein}, where labels may represent RGB values of colors or physical properties of protein secondary structure elements, respectively.
Research in graph kernels has achieved a remarkable progress in recent years.
However, it has focused mainly on unlabeled graphs and graphs with discrete node labels.
For such kind of graphs, there are several highly scalable graph kernels available which can handle graphs with thousands of vertices (\eg the Weisfeiler-Lehman subtree kernel).
However, the same does not happen in the case of datasets where node labels correspond to vectors.
Some of the existing graph kernels for node-labeled graphs such as the shortest-path kernel can be extended to handle continuous labels.
Unfortunately, by taking into account these labels, their computational complexity becomes prohibitive.
Designing graph kernels for graphs with continuous node labels is a much less well studied problem which started to gain some attention recently \shortcite{kriege2012subgraph,feragen2013scalable,orsini2015graph,neumann2016propagation,morris2016faster,su2016fast,kondor2016multiscale}.
There are mainly two categories of approaches for graphs with continuous node labels: ($1$) those that directly handle continuous node labels, and ($2$) those that first discretize the node labels and then employ existing kernels that operate on graphs with discrete node labels.
We will next present some kernels belonging to the first category.
With regards to the second category, worthy of mention is the work of \shortciteA{morris2016faster} that proposes the hash graph kernel framework which iteratively transforms continuous attributes into discrete labels using randomized hash functions, thus allowing kernels that support discrete node labels to handle node-attributed graphs.

\subsubsection{Subgraph Matching Kernel}
The subgraph matching kernel counts the number of matchings between subgraphs of bounded size in two graphs \shortcite{kriege2012subgraph}.
The kernel is very general since it can be applied to graphs that contain node labels, edge labels, node attributes or edge attributes.

Let $\mathcal{G}$ be a set of graphs.
We assume that the graphs that are contained in the set are labeled or attributed.
Specifically, let $\ell$ be a labeling function that assigns either discrete labels or continuous attributes to vertices and edges.
A graph isomorphism between two labeled/attributed graphs $G=(V,E)$ and $G'=(V',E')$ is a bijection $\phi : V \rightarrow V'$ that preserves adjacencies, that is $\forall v,u \in V : (v,u) \in E \Leftrightarrow (\phi(v), \phi(u)) \in E'$, and labels, that is if $\psi \in V \times V \rightarrow V' \times V'$ is the mapping of vertex pairs implicated by the bijection $\phi$ such that $\psi((v,u)) = (\phi(v), \phi(u))$, then, the conditions $\forall v \in V : \ell(v) \equiv \ell(\phi(v))$ and $\forall e \in E : \ell(e) \equiv \ell(\psi(e))$ must hold, where $\equiv$ denotes that two labels are considered equivalent.

\begin{definition}[Subgraph Matching Kernel]
  Given two graphs $G=(V,E)$ and $G'=(V',E')$, let $\mathcal{B}(G,G')$ denote the set of all bijections between sets $S \subseteq V$ and $S' \subseteq V'$, and let $\lambda : \mathcal{B}(G,G') \rightarrow \mathbb{R}^+$ be a weight function.
  The subgraph matching kernel is defined as
  \begin{equation}
      k(G, G') = \sum_{\phi \in \mathcal{B}(G,G')} \lambda(\phi) \prod_{v \in S} \kappa_V(v, \phi(v)) \prod_{e \in S \times S} \kappa_E(e, \psi(e))
  \end{equation}
  where $S = dom(\phi)$ and $\kappa_V, \kappa_E$ are kernel functions defined on vertices and edges, respectively.
\end{definition}

The instance of the subgraph matching kernel that is obtained if we set the $\kappa_V, \kappa_E$ functions as follows
\begin{equation}
    \begin{split}
        \kappa_V(v,v') &= \begin{cases}
        1, & \text{if $\ell(v) \equiv \ell(v')$},\\
        0, & \text{otherwise and} 
        \end{cases}\\
        \kappa_E(e,e') &= \begin{cases}
        1, & \text{if $e \in E \wedge e' \in E' \wedge \ell(e) \equiv \ell(e')$ or $e \not \in E \wedge e' \not \in E'$},\\
        0, & \text{otherwise.}
        \end{cases}
    \end{split}
\end{equation}
is known as the common subgraph isomorphism kernel.
This kernel counts the number of isomorphic subgraphs contained in two graphs.

To count the number of isomorphisms between subgraphs, the kernel capitalizes on a classical result of \shortciteA{levi1973note} which makes a connection between common subgraphs of two graphs and cliques in their product graph.
More specifically, each maximum clique in the product graph is associated with a maximum common subgraph of the factor graphs.
This allows someone to compute the common subgraph isomorphism kernel by enumerating the cliques of the product graph.

The general subgraph matching kernel extends the theory of Levi and builds a weighted product graph to allow a more flexible scoring of bijections.
Given two graphs $G=(V,E)$, $G'=(V',E')$, and vertex and edge kernels $\kappa_V$ and $\kappa_E$, the weighted product graph $G_P=(V_P, E_P)$ of $G$ and $G'$ is defined as
\begin{equation}
    \begin{split}
        V_P &= \{ (v,v') \in V \times V' : \kappa_V(v,v') > 0 \} \\
        E_P &= \{ \{(v,v'),(u,u')\} \in V_P \times V_P : v \neq u \wedge v' \neq u' \wedge \kappa_E((v,u),(v',u')) > 0 \} \\
        c(u) &= \kappa_V(v,v') \quad \forall u=(v,v') \in V_P \\
        c(e) &= \kappa_E((v,u),(v',u')) \quad \forall e \in E_P, \\
        \text{where } &e=((v,u),(v',u')) 
    \end{split}
\end{equation}
After creating the weighted product graph, the kernel enumerates its cliques.
The kernel starts from an empty clique and extends it stepwise by all vertices preserving the clique property.
Let $w$ be the weight of a clique $C$.
Whenever the clique $C$ is extended by a new vertex $v$, the weight of the clique is updated as follows: first it is multiplied by the weight of the vertex $w' = w \, c(v)$, and then, it is multiplied by all the edges connecting $v$ to a vertex in $C$, that is $w' = \sum_{u \in C} w \, c((v,u))$.
The algorithm effectively avoids duplicates by removing a vertex from the candidate set after all cliques containing it have been exhaustively explored.

The runtime of the subgraph matching kernel depends on the number of cliques in the product graph.
The worst-case runtime complexity of the kernel when considering subgraphs of size up to $k$ is $\mathcal{O}(kn^{k+1})$, where $n=|V|+|V'|$ is the sum of the number of vertices of the two graphs.

\subsubsection{GraphHopper Kernel}
The GraphHopper kernel is closely related to the shortest path kernel.
In the case of graphs with discrete node labels, the kernels $k_v$ and $k_e$ of the shortest-path kernel which compare vertex labels and path lengths correspond typically to dirac kernels.
Hence, nodes and shortest path lengths are considered similar if they are completely identical.
That specific instance of the shortest path kernel allows the use of an explicit computation scheme which is very efficient, even for larger datasets.
However, for attributed graphs, such an explicit mapping is no longer possible.
This has a large impact on the runtime of the algorithm which is generally $\mathcal{O}(n^4)$, and makes the kernel unfeasible for many real-world applications.
GraphHopper is a kernel which also compares shortest paths between node pairs from the two graphs, but with a different path kernel
\shortcite{feragen2013scalable}.
The kernel takes into account both path lengths and the vertices encountered while ``hopping'' along shortest paths.
The kernel is equivalent to a weighted sum of node kernels.
Moreover, it can handle both labeled and attributed graphs, and is much more efficient than the shortest-path kernel.

Let $G=(V,E)$ be a graph.
The graph contains discrete node labels, continuous node attributes or both.
Let $\ell$ be a labeling function that assigns either discrete labels or continuous attributes to vertices.
The kernel compares node labels/attributes using a kernel $k_n$ (\eg delta kernel in the case of node labels, and linear or gaussian kernel in the case of node attributes).
Given two vertices $v,u \in V$, a path $\pi$ from $v$ to $u$ in $G$ is defined as a sequence of vertices
\begin{equation}
    \pi = [v_1, v_2, v_3, \ldots, v_l]
\end{equation}
where $v_1 = v$, $v_l = u$ and $(v_i, v_{i+1}) \in E$ for all $i=1,\ldots,l-1$.
Let $\pi(i) = v_i$ denote the $i$-th vertex encountered when ``hopping'' along the path.
Denote by $l(\pi)$ the weighted length of $\pi$ and by $|\pi|$ its discrete length, defined as the number of vertices in $\pi$.
The shortest path $\pi_{ij}$ from $v_i$ to $v_j$ is defined in terms of weighted length.
The diameter $\delta(G)$ of $G$ is the maximal number of nodes in a shortest path in $G$, with respect to the weighted path length.

\begin{definition}[GraphHopper Kernel]
  The GraphHopper kernel is defined as a sum of path kernels $k_p$ over the families $P, P'$ of shortest
  paths in $G,G'$
  \begin{equation}
      k(G,G') = \sum_{\pi \in P} \sum_{\pi' \in P'} k_p(\pi, \pi')
  \end{equation}
  The path kernel $k_p(\pi, \pi')$ is a sum of node kernels $k_n$ on vertices simultaneously encountered while simultaneously hopping along paths $\pi$ and $\pi'$ of equal discrete length, that is
  \begin{equation}
      k_p(\pi, \pi') = \begin{cases}
          \sum_{j=1}^{|\pi|} k_n(\pi(j), \pi'(j)), & \text{if $|\pi| = |\pi'|$},\\
          0, & \text{otherwise.} 
          \end{cases}
  \end{equation}
\end{definition}

The $k(G,G')$ kernel can be decomposed into a weighted sum of node kernels
\begin{equation}
    k(G,G') = \sum_{v \in V} \sum_{v' \in V'} w(v,v') k_n(v, v')
\end{equation}
where $w(v,v')$ counts the number of times $v$ and $v'$ appear at the same hop, or coordinate, $i$ of shortest paths $\pi,\pi'$ of equal discrete length $|\pi| = |\pi'|$.
We can decompose the weight $w(v,v')$ as
\begin{equation}
    w(v,v') = \sum_{j=1}^\delta \sum_{i=1}^\delta | \{ (\pi,\pi') : \pi(i)=v, \pi'(i)=v', |\pi|=|\pi'|=j \} | = \sum_{j=1}^\delta \sum_{i=1}^\delta M_{ij}^v M_{ij}^{v'}
\end{equation}
where $M^v$ is a $\delta \times \delta$ matrix whose entry $M_{ij}^v$ counts how many times $v$ appears at the $i$-th coordinate of a shortest path in $G$ of discrete length $j$, and $\delta = \max(\delta(G), \delta(G'))$.
The components of these matrices can be computed efficiently using recursive message-passing algorithms. 
The total complexity of computing the GraphHopper kernel is $\mathcal{O}(n^2(m + \log n + d + \delta^2))$ where $n$ is the number of vertices, $m$ is the number of edges and $d$ is the dimensionality of the node attributes ($d=1$ in the case of discrete node labels).

\subsubsection{Graph Invariant Kernels}
Kernels for attributed graphs have received increased attention recently, and research efforts have focused not only on new kernels, but also on frameworks for building kernels that can handle such continuous node attributes.
Graph invariant kernels are instances of such a framework \shortcite{orsini2015graph}.
These kernels decompose graphs into sets of vertices, and compare them to each other using a kernel that measures their similarity both in terms of their attributes and in terms of their structural roles.

Let $G$ be a graph.
Let $R$ be a decomposition relation that specifies a decomposition of $G$ into its parts.
Then, we denote by $R^{-1}(G)$ the multiset of all patterns in $G$.
An example of such a decomposition relation is the one that generates neighborhood subgraphs.
Graph invariant kernels compare vertices of graphs based on their attributes, but also based on their  structural role in subgraphs obtained using a decomposition relation.
\begin{definition}[Graph Invariant Kernel]
  Given two attributed graphs $G=(V,E)$ and $G'=(V',E')$, the graph invariant kernels compare the attributes of all pairs of vertices of the two graphs using a kernel
  \begin{equation}
    k(G, G') = \sum_{v \in V} \sum_{v' \in V'} w(v, v') \ k_{attr}(v, v')
  \end{equation}
  where $k_{attr}$ is a kernel between vertex attributes, and $w(v, v')$ is a weight function defined as follows
  \begin{equation}
    w(v, v') = \sum_{g \in R^{-1}(G)} \ \sum_{g' \in R^{-1}(G')} k_{inv}(v, v') \ \frac{\delta_m(g, g')}{|V_{g}| |V_{g'}|} \ \mathbbm{1} \{ v \in V_{g} \wedge v' \in V_{g'} \}
  \end{equation}
  where $\delta_m$ is a dirac function that determines whether two patterns match, $V_{g},V_{g'}$ are the set of vertices of patterns $g, g'$, and $\mathbbm{1}$ is an indicator function. 
\end{definition}
If $g, g'$ are subgraphs of $G, G'$, $\delta_m$ can be a dirac function that compares the canonical representations of the subgraphs obtained by applying a labeling function which produces efficient string encodings of the subgraphs along with a hash function from strings to natural numbers.  
The indicator function $\mathbbm{1} \{ v \in V_{g} \wedge v' \in V_{g'} \}$ from all the subgraphs extracted from the two graphs selects only those in which vertices $v$ and $v'$ are involved into.
The kernel function $k_{inv}$ is used to measure the similarity between the colors produced by a vertex invariant $\mathcal{L}$ and encodes the extent to which the vertices play the same structural role in the two subgraphs.
By employing different graph invariants $\mathcal{L}$, different instances of graph invariant kernels emerge.
Some common graph invariants include the Weisfeiler-Lehman relabeling procedure and coloring methods that capitalize on diffusion updates.
For kernels that decompose graphs into sets of subgraphs, their complexity is $\mathcal{O} \big( n^2(d_{attr}+d_{inv}n^2|V_g|^2) \big)$

\subsubsection{Propagation Kernel}
The propagation kernel is another instance of the neighborhood aggregation framework, and in contrast to most other instances, it can handle continuous node attributes \shortcite{neumann2016propagation}.
The kernel leverages quantization in order to transform continuous node attributes to discrete labels.
Similarly to the Weisfeiler-Lehman subtree kernel, the propagation kernel applies an iterative procedure which updates the node attributes, places the nodes into bins based on their attributes, and counts nodes that fall into the same bins in two graphs. 

Let $G=(V,E)$ be a node-attributed graph.
Let also $P_0$ be a matrix whose $i$-th row contains the intial attribute of vertex $v_i \in V$.
The propagation kernel first uses a hash function that maps the node attributes to integer-valued bins, such that vertices with similar attributes end up in the same bin.
Hence, this function maps each row of matrix $P_0$ to an integer.
Then, the kernel employs a propagation scheme to update the attributes of the vertices.
Different schemes can be employed.
A common scheme updates node attributes as follows
\begin{equation}
  P_{t+1} = D^{-1} A P_t
\end{equation}
where $D$ is a diagonal matrix with $D_{ii} = \sum_j A_{ij}$, and $D^{-1} A$ corresponds to the transition matrix, that is the row-normalized adjacency matrix.
The above two steps (hashing and update of node attributes) are performed for $T$ iterations.

\begin{definition}[Propagation Kernel]
  Let $G$, $G'$ be two node-attributed graphs.
  Define $n_i$ as the number of integer bins occupied by nodes of $G$ and $G'$ after applying the hashing function to the node attributes at the $i$-th iteration of the algorithm.
  Let also $c_t(G, i)$ be the number of nodes of $G$ placed into bin $i$ at the $t$-th iteration of the algorithm.
  Then, the propagation kernel on two graphs $G$ and $G'$ with $T$ iterations is defined as
  \begin{equation}
    k(G,G') = \langle \phi(G),\phi(G') \rangle 
  \end{equation}
  where
  \begin{equation}
    \phi(G) = (c_0(G, 1),\ldots,c_0(G, n_0),\ldots,c_T(G, 1),\ldots,c_T(G, n_T))
  \end{equation}
  and
  \begin{equation}
    \phi(G') = (c_0(G', 1),\ldots,c_0(G', n_0),\ldots,c_T(G', 1),\ldots,c_T(G', n_T))
  \end{equation}
\end{definition}
An illustration of the propagation kernel is given in Figure~\ref{fig:propagation_kernel}.
\begin{figure}[t]
  \centering
  \includegraphics[width=\textwidth]{propagation_kernel}
  \caption{Propagation kernel computation. Distributions, bins, count features, and kernel contributions for two graphs $G$ and $G'$ with binary node labels and one iteration of label propagation. Node-label distributions are decoded by color.}
  \label{fig:propagation_kernel}
\end{figure}
The total runtime complexity of the kernel is $\mathcal{O}\big((T-1)m+Tn\big)$.

\subsubsection{Multiscale Laplacian Graph Kernel}
The multiscale Laplacian graph kernel can handle unlabeled graphs, graphs with discrete node labels, but also graphs with continuous node attributes \shortcite{kondor2016multiscale}.
It takes into account structure in graphs at a range of different scales by building a hierarchy of nested subgraphs.
These subgraphs are compared to each other using another graph kernel, called the feature space Laplacian graph kernel.
This kernel is capable of lifting a base kernel defined on the vertices of two graphs to a kernel between the graphs themselves.
Since exact computation of the multiscale Laplacian graph kernel is a very expensive operation, the kernel uses a randomized projection procedure  similar to the popular Nystr{\"o}m approximation for kernel matrices \shortcite{williams2001using}.

Let $G=(V,E)$ be an undirected graph such that $n = |V|$ and let $L$ be the Laplacian of $G$.
Given two graphs $G_1$ and $G_2$ of $n$ vertices, we can define the kernel between them to be a kernel between the corresponding normal distributions $p_1 = \mathcal{N}(0, L_1^{-1})$ and $p_2 = \mathcal{N}(0, L_2^{-1})$ where $0$ is the $n$-dimensional all-zeros vector.
Note that the Laplacian matrices of the two graphs have a zero eigenvalue eigenvector.
Hence, in order to be able to invert them, the algorithm adds a small constant ``regularizer'' $\eta I$ to them.
In the following, we denote the regularized Laplacians of $G_1$ and $G_2$ by $L_1$ and $L_2$, respectively.
More specifically, given two graphs $G_1$ and $G_2$ of $n$ vertices with regularized Laplacians $L_1$ and $L_2$ respectively, the Laplacian graph kernel with parameter $\gamma$ between the two graphs is
\begin{equation}
    k_{LG}(G_1, G_2) = \frac{| (\frac{1}{2} S_1^{-1} + \frac{1}{2} S_2^{-1} )^{-1} |^{1/2}}{|S_1|^{1/4} |S_2|^{1/4}} 
\end{equation}
where $S_1 = L_1^{-1} + \gamma I$, $S_2 = L_2^{-1} + \gamma I$ and $I$ is the $n \times n$ identity matrix.
The Laplacian graph kernel captures similarity between the overall shapes of the two graphs.
However, it assumes that both graphs have the same size, and it is not invariant to permutations of the vertices.

To achieve permutation invariance, the multiscale Laplacian graph kernel represents each vertex as a $d$-dimensional vector whose components correspond to local and permutation invariant vertex features.
Such features may include for instance the degree of the vertex or the number of triangles in which it participates.
Then, it performs a linear transformation and represents each graph as a distribution of the considered features instead of a distribution of its vertices.
Let $U_1, U_2 \in \mathbb{R}^{d \times n}$ be the feature mapping matrices of the two graphs, that is the matrices whose columns contain the vector representations of the vertices of the two graphs. 
Then, the feature space Laplacian graph kernel is defined as
\begin{equation}
    k_{FLG}(G_1, G_2) = \frac{| (\frac{1}{2} S_1^{-1} + \frac{1}{2} S_2^{-1} )^{-1} |^{1/2}}{|S_1|^{1/4} |S_2|^{1/4}} 
\end{equation}
where $S_1 = U_1 L_1^{-1} U_1^\top + \gamma I$, $S_2 = U_2 L_2^{-1} U_2^\top + \gamma I$ and $I$ is the $d \times d$ identity matrix.
Since the vertex features are local and invariant to vertex reordering, the feature space Laplacian graph kernel is permutation invariant.
Furthermore, since the distributions now live in the space of features rather than the space of vertices, the feature space Laplacian graph kernel can be applied to graphs of different sizes.

Let $\phi(v)$ be the representation of vertex $v$ constructed from local vertex features as described above.
The base kernel $\kappa$ between two vertices $v_1$ and $v_2$ corresponds to the dot product of their feature vectors
\begin{equation}
    \kappa(v_1, v_2) = \phi(v_1)^\top \phi(v_2) 
\end{equation}
Let $G_1$ and $G_2$ be two graphs with vertex sets $V_1 = \{ v_1, \ldots, v_{n_1}\}$ and $V_2 = \{ u_1, \ldots, u_{n_2} \}$ respectively, and let $\bar{V} = \{ \bar{v}_1, \ldots, \bar{v}_{n_1+n_2} \}$ be the union of the two vertex sets.
Let also $K \in \mathbb{R}^{(n_1+n_2) \times (n_1+n_2)}$ be the kernel matrix defined as
\begin{equation}
    K_{ij} = \kappa(\bar{v}_i, \bar{v}_j) = \phi(\bar{v}_i)^\top \phi(\bar{v}_j)
\end{equation}
Let $u_1, \ldots, u_p$ be a maximal orthonormal set of the non-zero eigenvalue eigenvectors of $K$
with corresponding eigenvalues $\lambda_1, \ldots, \lambda_p$.
Then the vectors
\begin{equation}
    \xi_i = \frac{1}{\sqrt{\lambda_i}} \sum_{l=1}^{n_1+n_2} [u_i]_l \phi(\bar{v}_l)
\end{equation}
where $[u_i]_l$ is the $l$-th component of vector $u_i$ form an orthonormal basis for the subspace $\{ \phi(\bar{v}_1), \ldots, \phi(\bar{v}_{n_1+n_2}) \}$.
Moreover, let $Q = [ \lambda_1^{1/2} u_1, \ldots,\lambda_p^{1/2} u_p ] \in \mathbb{R}^{p \times p}$ and $Q_1, Q_2$ denote the first $n_1$ and last $n_2 $ rows of matrix $Q$ respectively.
Then, the generalized feature space Laplacian graph kernel induced from the base kernel $\kappa$ is defined as
\begin{equation}
    k_{FLG}^\kappa(G_1, G_2) = \frac{| (\frac{1}{2} S_1^{-1} + \frac{1}{2} S_2^{-1} )^{-1} |^{1/2}}{|S_1|^{1/4} |S_2|^{1/4}} 
\end{equation}
where $S_1 = Q_1 L_1^{-1} Q_1^\top + \gamma I$ and $S_2 = Q_2 L_2^{-1} Q_2^\top + \gamma I$ where $I$ is the $p \times p$ identity matrix.

The multiscale Laplacian graph kernel builds a hierarchy of nested subgraphs, where each subgraph is centered around a vertex and computes the generalized feature space Laplacian graph kernel between every pair of these subgraphs.
Let $G$ be a graph with vertex set $V$, and $\kappa$ a positive semi-definite kernel on $V$.
Assume that for each $v \in V$, we have a nested sequence of $L$ neighborhoods
\begin{equation}
    v \in \mathcal{N}_1(v) \subseteq \mathcal{N}_2(v) \subseteq \ldots \subseteq \mathcal{N}_{l_{max}}(v)
\end{equation}
and for each $\mathcal{N}_l(v)$, let $G_l(v)$ be the corresponding induced subgraph of $G$.
The multiscale Laplacian subgraph kernels are defined as $\mathfrak{K}_1, \ldots, \mathfrak{K}_{l_{max}} : V \times V \rightarrow \mathbb{R}$ as follows
\begin{enumerate}
    \item $\mathfrak{K}_1$ is just the generalized feature space Laplacian graph kernel $k_{FLG}^\kappa$ induced from the base kernel $\kappa$ between the lowest level subgraphs (\ie the vertices)
    \begin{equation}
        \mathfrak{K}_1(v,u) = k_{FLG}^\kappa(v, u)
    \end{equation}
    \item For $l=2,3,\ldots,l_{max}$, $\mathfrak{K}_l$ is the the generalized feature space Laplacian graph kernel induced from $\mathfrak{K}_{l-1}$ between $G_l(v)$ and $G_l(u)$
    \begin{equation}
        \mathfrak{K}_l(v,u) = k_{FLG}^{\mathfrak{K}_{l-1}} \big( G_l(v), G_l(u) \big)
    \end{equation}
\end{enumerate}

\begin{definition}[Multiscale Laplacian Graph Kernel]
	Let $G_1, G_2$ be two graphs.
	The multiscale Laplacian graph kernel between the two graphs is defined as follows
	\begin{equation}
	    k(G_1, G_2) = k_{FLG}^{\mathfrak{K}_{l_{max}}}(G_1, G_2)
	\end{equation}
\end{definition}
The multiscale Laplacian graph kernel computes $\mathfrak{K}_1$ for all pairs of vertices, then computes $\mathfrak{K}_2$ for all pairs of vertices, and so on.
Hence, it requires $\mathcal{O}(n^2 l_{max})$ kernel evaluations.
At the top levels of the hierarchy each subgraph centered around a vertex $G_l(v)$ may have as many as $n$ vertices.
Therefore, the cost of a single evaluation of the generalized feature space Laplacian graph kernel may take $\mathcal{O}(n^3)$ time.
This means that in the worst case, the overall cost of computing $k$ is $\mathcal{O}(n^5 l_{max})$.
Given a dataset of $N$ graphs, computing the kernel matrix requires repeating this for all pairs of graphs, which takes $\mathcal{O}(N^2n^5l_{max})$ time and is clearly problematic for real-world settings.

The solution to this issue is to compute for each level $l=1,2,\ldots,l_{max}+1$ a single joint basis for all subgraphs at the given level across all graphs.
Let $G_1, G_2, \ldots, G_N$ be a collection of graphs, $V_1, V_2, \ldots, V_N$ their vertex sets, and assume that $V_1, V_2, \ldots, V_N \subseteq \mathcal{V}$ for some general vertex space $\mathcal{V}$.
The joint vertex feature space of the whole graph collection is $W = span \big\{ \bigcup_{i=1}^N \bigcup_{v \in V_i} \{ \phi(v) \} \big\}$.
Let $c = \sum_{i=1}^N |V_i|$ be the total number of vertices and $\bar{V} = (\bar{v}_1, \ldots, \bar{v}_c)$ be the concatenation of the vertex sets of all graphs.
Let $K$ be the corresponding joint kernel matrix and $u_1, \ldots, u_p$ be a maximal orthonormal set of non-zero eigenvalue eigenvectors of $K$ with corresponding eigenvalues $\lambda_1,\ldots,\lambda_p$ and $p=dim(W)$.
Then the vectors
\begin{equation}
    \xi_i = \frac{1}{\sqrt{\lambda_i}} \sum_{l=1}^c [u_i]_l \phi(\bar{v}_l) \qquad i=1,\ldots,p
\end{equation}
form an orthonormal basis for $W$.
Moreover, let $Q = [ \lambda_1^{1/2} u_1, \ldots, \lambda_p^{1/2} u_p ] \in \mathbb{R}^{p \times p}$ and $Q_1$ denote the first $n_1$ rows of matrix $Q$, $Q_2$ denote the next $n_2 $ rows of matrix $Q$ and so on.
For any pair of graphs $G_i, G_j$ of the collection, the generalized feature space Laplacian graph kernel induced from $\kappa$ can be expressed as
\begin{equation}
    k_{FLG}^\kappa(G_i, G_j) = \frac{| (\frac{1}{2} \bar{S}_i^{-1} + \frac{1}{2} \bar{S}_j^{-1} )^{-1} |^{1/2}}{|\bar{S}_i|^{1/4} |\bar{S}_j|^{1/4}} 
\end{equation}
where $\bar{S}_i = Q_i L_i^{-1} Q_i^\top + \gamma I$, $\bar{S}_j = Q_j L_j^{-1} Q_j^\top + \gamma I$ and $I$ is the $p \times p$ identity matrix.

Computing the kernel matrix between all vertices of all graphs ($c$ vertices in total) and storing it is a very costly procedure.
Computing its eigendecomposition is even worse in terms of the required runtime.
Morever, $p$ is also very large.
Hence, managing the $\bar{S}_1, \ldots, \bar{S}_N$ matrices (each of which is of size $p \times p$) becomes infeasible.
Hence, the multiscale Laplacian graph kernel replaces $W$ with a smaller, approximate joint features space.
Let $\tilde{V} = (\tilde{v}_1, \ldots, \tilde{v}_{\tilde{c}})$ be $\tilde{c} \ll c$ vertices sampled from the joint vertex set.
Then, the corresponding subsampled vertex feature space is $\tilde{W} = span \{ \phi(v) : v \in \tilde{V} \}$.
Let $\tilde{p} = dim(\tilde{W})$.
Similarly to before, the kernel constructs an orthonormal basis $\{ \xi_1, \ldots, \xi_{\tilde{p}} \}$ for $\tilde{W}$ by forming the (now much smaller) kernel matrix $K_{ij} = \kappa(\tilde{v}_i, \tilde{v}_j)$, computing its eigenvalues and eigenvectors, and setting $\xi_i = \frac{1}{\sqrt{\lambda_i}} \sum_{l=1}^{\tilde{c}} [u_i]_l \phi(\tilde{v}_l)$. 
The resulting approximate generalized feature space Laplacian graph kernel is
\begin{equation}
    k_{FLG}^\kappa(G_1, G_2) = \frac{| (\frac{1}{2} \tilde{S}_1^{-1} + \frac{1}{2} \tilde{S}_2^{-1} )^{-1} |^{1/2}}{|\tilde{S}_1|^{1/4} |\tilde{S}_2|^{1/4}} 
\end{equation}
where $\tilde{S}_1 = \tilde{Q}_1 L_1^{-1} \tilde{Q}_1^\top + \gamma I$, $\tilde{S}_2 = \tilde{Q}_2 L_2^{-1} \tilde{Q}_2^\top + \gamma I$ are the projections of $\bar{S}_1$ and $\bar{S}_2$ to $\tilde{W}$ and $I$ is the $\tilde{p} \times \tilde{p}$ identity matrix.
Finally, the kernel introduces a further layer of approximation by restricting $\tilde{W}$ to be the space spanned by the first $\hat{p} < \tilde{p}$ basis vectors (ordered by descending eigenvalue), effectively doing kernel PCA on $\{ \phi(\tilde{v}) \}_{\tilde{v} \in \tilde{V}}$.
The combination of these two factors makes computing the entire stack of kernels feasible, reducing the complexity of computing the kernel matrix for a dataset of $N$ graphs to $\mathcal{O}(N \tilde{c}^2 \hat{p}^3 l_{max} + N \tilde{c}^3 l_{max} + N^2 \hat{p}^3)$.

\subsection{Frameworks}
Besides designing kernels, research on graph kernels has also focused on \textit{frameworks} and approaches that can be applied to existing graph kernels and increase their performance.
The most popular of all frameworks is perhaps the Weisfeiler-Lehman framework which has been already presented \shortcite{shervashidze2011weisfeiler}.
Interestingly, any kernel that can handle discrete node labels can be plugged into that framework.
Recently, two other frameworks were presented for deriving variants of popular $R$-convolution graph kernels \shortcite{yanardag2015deep,yanardag2015structural}.
Inspired by recent advances in NLP, these frameworks offer a way to take into account similarity between substructures.
In addition, a method that combines several kernels using the multiple kernel learning framework was also recently proposed \shortcite{aiolli2015multiple}.
Another recently proposed framework generates a hierarchy of subgraphs and compares the corresponding according to the hierarchy subgraphs using graph kernels \shortcite{nikolentzos2018}.
Moreover, a recent approach employs graph kernels and performs a series of successive embeddings in order to derive more expressive kernels \shortcite{nikolentzos2018enhancing}.
Some of these frameworks are described in more detail below.

\subsubsection{Frameworks Dealing with Diagonal Dominance}
We next present two frameworks that are inspired by recent advances in natural language processing, namely the deep graph kernels framework \shortcite{yanardag2015deep} and the structural smoothing framework \shortcite{yanardag2015structural}. 
These two frameworks were developed to address the problem of diagonal dominance which is inherent to $R$-convolution kernels.
The feature space of these kernels is usually large (\ie grows exponentially) and we encounter the sparsity problem: only a few substructures will be common across graphs, and therefore each graph is similar to itself, but not to any other graph in the dataset.
However, the substructures used to define a graph kernel are often related to each other, but commonly-used $R$-convolution kernels respect only exact matchings. 
For example, when the features correspond to large graphlets (\eg $k \geq 5$), two graphs may be composed of many similar graphlets, but not any identical.
As a consequence, the kernel value between the two graphs (\ie inner product of their feature representations) will be equal to $0$ even though the two graphs are similar to each other.

Ideally, we would like the kernels to output large values for pairs of graphs that belong to the class, and lower values for pairs of graphs that belong to different classes.
To deal with the aforementioned problem, the deep graph kernels framework computes the kernel between two graphs $G$ and $G'$ as follows
\begin{equation}
    k(G,G') = \phi(G)^\top \ M \ \phi(G')
\end{equation}
where $M$ represents a positive semidefinite matrix that encodes the relationship between substructures and $\phi(G), \phi(G')$ are the representations of graphs $G, G'$ according to a graph kernel which contains counts of atomic substructures.
Therefore, one can design an $M$ matrix that respects the similarity of the substructure space.
Clearly, the deep graph kernels framework can be applied only to graph kernels whose feature maps $\phi$ can be computed explicitly.

Matrix $M$ can be generated by manually defining functions to compare substructures or alternatively, it can be learned using techniques inspired from the field of natural language processing.
When substructures exhibit a clear mathematical relationship, one can define a function to measure the similarities between them (\eg edit distance in the case of graphlets).
However, the above approach requires manually designing the similarity functions.
Furthermore, in many cases, it becomes prohibitively expensive to compare all pairs of substructures.
On the other hand, learning the latent representations of substructures is more efficient and does not involve any manual intervention.
Matrix $M$ can then be computed based on the learned representations.
To learn a latent representation for each substructure, the framework utilizes recent approaches for generating word embeddings such as the continuous bag-of-words (CBOW) and Skip-gram models \shortcite{mikolov2013distributed}.
These models generate semantic representations from word co-occurrence statistics derived from large text corpora.
However, unlike words in a traditional text corpora, substructures of graphs do not have a linear co-occurrence relationship.
Hence, these co-occurrence relationships need to be manually defined. 
\shortciteA{yanardag2015deep} proposed a methodology on how to generate corpora where co-occurrence relationship is meaningful on three popular kernels, namely the Weisfeiler-Lehman subtree kernel, the graphlet kernel, and the shortest path kernel.

The structural smoothing framework is inspired by recent smoothing techniques in natural language processing.
Similar to the deep graph kernels framework, this framework can also only be applied to graph kernels whose feature maps $\phi$ can be computed explicitly.
The framework takes structural similarity into account by constructing a directed acyclic graph (DAG) that encodes the relationships between lower and higher order substructures.
Each vertex of the DAG corresponds to a substructure (and also to a feature in the explicit graph representation).
For each substructure $s$ of size $k$, the framework determines all possible substructures of size $k-1$ into which $s$ can be reduced.
These substructures are the parents of $s$, and a weighted directed edge is drawn from each parent to its children vertices.
Since all descendants of a given substructure at depth $k-1$ are at depth $k$, the emerging graph is indeed a DAG.
\shortciteA{yanardag2015deep} present how such a DAG can be constructed for three popular graph kernels, namely the Weisfeiler-Lehman subtree kernel, the graphlet kernel, and the shortest path kernel.
Given the DAG, the structural smoothing for a substructure $s$ at level $k$ is defined as
\begin{equation}
    P_{SS}^k(s) = \frac{\max(c_s-d,0)}{m} + \frac{d m_d}{m} \sum_{p \in \mathcal{P}_s} P_{SS}^{k-1}(p) \frac{w_{ps}}{\sum_{c \in \mathcal{C}_p} w_{pc}}
\end{equation}
where $c_i$ denotes the number of times substructure $i$ appears in the graph, $m = \sum_i c_i$ denotes the total number of substructures present in the graph, $d > 0$ is a discount factor, $m_d = |\{i : c_i > d\}|$ is the number of substructures whose counts are larger than $d$, $w_{ij}$ denotes the weight of the edge connecting vertex $i$ to vertex $j$, $\mathcal{P}_s$ denotes the parents of vertex $s$, and $\mathcal{C}_p$ the children of vertex $p$.
The above equation subtracts a fixed discount factor $d$ from every substructure that appears in the graph, and accumulates it to a total mass of $d m_d$.
Each substructure $s$ receives some portion of this accumulated probability mass from its parents.
The proportion of the mass that a parent $p$ at level $k-1$ transmits to a given child a depends on the weight $w_{ps}$ between the parent and the child, and the probability mass $P_{SS}^{k-1}(p)$ that is assigned to the parent.
It is thus clear that, even if a graph does not contain a substructure $s$ (\ie $c_s = 0$), its value in the feature vector may become greater than $0$ (\ie $P_{SS}(s) > 0$).

\subsubsection{Core Framework}
The core framework is another tool for improving the performance of graph kernels \shortcite{nikolentzos2018}.
This framework is not restricted to graph kernels, but can be applied to any graph comparison algorithm.
It capitalizes on the $k$-core decomposition which is capable of uncovering topological and hierarchical properties of graphs.
Specifically, the $k$-core decomposition is a powerful tool for network analysis and it is commonly used as a measure of importance and well connectedness of vertices in a broad spectrum of applications.
The notion of $k$-core was first introduced by Seidman to study the cohesion of social networks \shortcite{seidman1983network}.
In recent years, the $k$-core decomposition has been established as a standard tool in many application domains such as in network visualization \shortcite{alvarez2006large}.

\paragraph{Core Decomposition.}
Let $G = (V,E)$ be an undirected and unweighted graph.
Given a subset of vertices $S \subseteq V$, let $E(S)$ be the set of edges that have both end-points in $S$.
Then, $G'=(S,E(S))$ is the subgraph induced by $S$.
We use $G' \subseteq G$ to denote that $G'$ is a subgraph of $G$.
Let $G$ be a graph and $G'$ a subgraph of $G$ induced by a set of vertices $S$.
Then, $G'$ is defined to be a $k$-core of $G$, denoted by $C_k$, if it is a maximal subgraph of $G$ in which all vertices have degree at least $k$.
Hence, if $G'$ is a $k$-core of $G$, then $\forall v \in S$, $deg_{G'}(v) \geq k$.
Each $k$-core is a unique subgraph of $G$, and it is not necessarily connected.
The core number $c(v)$ of a vertex $v$ is equal to the highest-order core that $v$ belongs to.
In other words, $v$ has core number $c(v) = k$, if it belongs to the $k$-core but not to the $(k+1)$-core.
The degeneracy $\delta^*(G)$ of a graph $G$ is defined as the maximum $k$ for which graph $G$ contains a non-empty $k$-core subgraph, $\delta^*(G) = \max_{v \in V}c(v)$.
Furthermore, assuming that $\mathcal{C} = \{  C_0, C_1, \ldots, C_{\delta^*(G)} \}$ is the set of all $k$-cores, then $\mathcal{C}$ forms a nested chain
\begin{equation}
    C_{\delta^*(G)} \subseteq \ldots \subseteq C_1 \subseteq C_0 = G
\end{equation}
Therefore, the $k$-core decomposition is a very useful tool for discovering the hierarchical structure of graphs.
The $k$-core decomposition of a graph can be computed in $\mathcal{O}(n+m)$ time \shortcite{matula1983smallest,batagelj2011fast}. 
The underlying idea is that we can obtain the $i$-core of a graph if we recursively remove all vertices with degree less than $i$ and their incident edges from the graph until no other vertex can be removed.

\paragraph{Core Kernels.}
The $k$-core decomposition builds a hierarchy of nested subgraphs, each having stronger connectedness properties compared to the previous ones.
The core framework measures the similarity between the corresponding according to the hierarchy subgraphs and aggregates the results.
Let $G=(V,E)$ and $G'=(V',E')$ be two graphs.
Let also $k$ be any kernel for graphs.
Then, the core variant of the base kernel $k$ is defined as
\begin{equation}
  k_c(G, G') = k(C_0,C'_0) + k(C_1,C'_1) + \ldots + k(C_{\delta^*_{min}},C'_{\delta^*_{min}}) 
\end{equation}
where $\delta^*_{min}$ is the minimum of the degeneracies of the two graphs, and $C_0,C_1,\ldots,C_{\delta^*_{min}}$ and $C'_0,C'_1,\ldots,C'_{\delta^*_{min}}$ are the $0$-core, $1$-core,$\ldots$, $\delta^*_{min}$-core subgraphs of $G$ and $G'$, respectively.
By decomposing graphs into subgraphs of increasing importance, the algorithm is capable of more accurately capturing their underlying structure.

The computational complexity of the core framework depends on the complexity of the base kernel and the degeneracy of the graphs under comparison.
Given a pair of graphs $G, G'$ and an algorithm $A$ for comparing the two graphs, let $\mathcal{O}_A$ be the time complexity of algorithm $A$.
Let also $\delta^*_{min} = \min \big( \delta^*(G),\delta^*(G') \big)$ be the minimum of the degeneracies of the two graphs.
Then, the complexity of computing the core variant of algorithm $A$ is $\mathcal{O}_{c}=\delta^*_{min}\mathcal{O}_A$.

\subsection{Tree Kernels}\label{sec:tree_kernels}
Before delving into the connection between graph neural networks and graph kernels, it is important to stress that graph kernels are also very related to tree kernels which have been extensively studied mainly in the field of natural language processing \shortcite{collins2001convolution}, but also in other fields \shortcite{vert2002tree}.
In fact, tree kernels were introduced prior to graph kernels.
A tree is an undirected graph in which any two vertices are connected by exactly one path, and thus a tree is a special case of a graph.
Therefore, tree kernels can be thought of as instances of graph kernels specifically designed for trees.
Note that any graph kernel can be applied to trees.
However, the opposite does not hold.
Tree kernels cannot be directly applied to general graphs.
In should be mentioned that certain graph kernels such as the subtree kernel build on ideas from tree kernels.

As already mentioned, tree kernels have found applications mainly in the field of natural language processing.
Examples of applications include semantic role labeling \shortcite{moschitti2004study,moschitti2006efficient,moschitti2008tree,croce2011structured}, relation extraction \shortcite{zelenko2003kernel,culotta2004dependency,bunescu2005shortest}, syntactic parsing re-ranking \shortcite{collins2001convolution} and question classification \shortcite{moschitti2006efficient,croce2011structured}.
In those tasks, an approach that has proven to be effective is to use a set of manually designed features that can capture the syntactic and semantic information encoded into the input data.
However, this set of meaningful features is usually determined by some domain expert, while the whole process is in most cases very expensive and time-consuming.
Instead of computing such handcrafted features, previous studies have capitalized on structured representations of text (\eg dependency parse trees) that might take into account syntactic and semantic aspects of the input data, and have introduced kernels that operate on these representations (\eg tree kernels).
Therefore, tree kernels are very useful since they eliminate the need for the design of new features in the context of several natural language tasks.
Most tree kernels represent trees in terms of their substructures.
And then, they compute the number of common substructures between two trees.
The most common substructures are the subtrees, the subset trees, and the partial trees which give rise to the subtree kernel \shortcite{smola2003fast}, the subset tree kernel \shortcite{collins2001convolution}, and the partial tree kernel \shortcite{moschitti2006efficient}, respectively.
A subtree is defined as a subgraph of the tree rooted at any non-leaf vertex along with all its descendants.
A subset tree is a more flexible structure since its leaves may correspond to non-leaf vertices of the input tree.
Subset trees satisfy the constraint that grammatical rules cannot be broken.
On the other hand, partial trees relax the above constraint and can be generated by the application of partial production rules of the grammar.

It is interesting to mention that graph kernels and tree kernels suffer from common limitations.
For instance, they both fix a set of features in advance, and they thus decouple data representation from learning.
Furthermore, there is no justification on why certain tree kernels perform better than others in a given task \shortcite{moschitti2006efficient,moschitti2006making}.

\section{Link to Graph Neural Networks}\label{sec:gnns}
In the past years, graph kernels have been largely overshadowed by a family of neural network architectures which operate on graphs, known as graph neural networks (GNNs).
The field of graph neural networks has seen an explosion of interest in recent years, with dozens of models developed which have been applied to various tasks such as to drug design \shortcite{kearnes2016molecular} and to modeling physical systems \shortcite{battaglia2016interaction}

The first instances of GNNs were proposed several years ago \shortcite{sperduti1997supervised,micheli2009neural,scarselli2009graph}, however, these models have only recently received a great deal of attention, following the advent of
deep learning.
More specifically, GNNs were initially categorized into spectral and spatial approaches \shortcite{bruna2014spectral}.
The first family of models operates on the spectral domain and draws on the properties
of convolutions in the Fourier domain, while the second family of models operates on the spatial domain where the weights of the edges determine locality.
Later, it became clear that all these models are special cases of a simple message passing framework (MPNNs) \shortcite{gilmer2017neural}.
Most of the recently proposed GNNs fit into this framework \shortcite{bruna2014spectral,duvenaud2015convolutional,li2015gated,defferrard2016convolutional,zhang2018end,xu2019powerful,murphy2019relational}.
Specifically, MPNNs employ a message passing procedure, where each vertex updates its feature vector by aggregating the feature vectors of its neighbors.
After $k$ iterations of the message passing procedure, each vertex obtains a feature vector which captures the structural information within its $k$-hop neighborhood.
MPNNs then compute a feature vector for the entire graph using some permutation invariant readout function such as summing the feature vectors of all the vertices of the graph.
In fact, the family of MPNNs is closely related to the Weisfeiler-Lehman test of isomorphism, and thus also to the Weisfeiler-Lehman subtree kernel \shortcite{shervashidze2011weisfeiler}.
Specifically, these models generalize the relabeling procedure of the Weisfeiler-Lehman subtree kernel to the case where vertices are associated with continuous feature vectors.
Standard MPNNs have been shown to be at most as powerful as the Weisfeiler-Lehman subtree kernel in distinguishing non-isomorphic graphs \shortcite{xu2019powerful,morris2019weisfeiler}.

It is interesting to mention that GNNs address some of the major limitations of graph kernels.
More specifically, as already discussed, graph kernels typically fix a set of features in advance.
This is one of the main limitations of graph kernels since data representation and learning are independent from each other.
The input samples are first implicitly or explicitly transformed into feature vector representations using a user-defined kernel.
Then, learning is perfomed based on these representations regardless of their quality.
Thus, the feature generation scheme is fixed and it does not adapt to the given data distribution.
Another limitation of graph kernels is that they cannot efficiently handle graphs whose vertices are annotated with continuous multi-dimensional attributes.
Indeed, while for unlabeled and node-labeled graphs, there are now available very efficient kernels which can handle graphs containing up to thousands of nodes, unfortunately, the same does not hold for graphs with continuous node attributes. 
Such attributes play an important role in different fields such as in bioinformatics and chemoinformatics, and this limitation renders kernels infeasible for application to these domains.
GNNs, on the other hand, have emerged as a machine learning framework addressing the above two challenges.

\subsection{Message Passing Models and the Weisfeiler-Lehman Test of Isomorphism}
As mentioned above, the majority of existing GNNs belongs to the family of MPNNs.
Suppose we have a MPNN model that contains $T$ neighborhood aggregation layers.
In the $t$-th neighborhood aggregation layer ($t > 0$), the hidden state $h_v^{(t)}$ of a vertex $v$ is updated as follows
\begin{equation}
    \begin{split}
        m_v^{(t)} &= \text{AGGREGATE}^{(t)}  \Bigl( \Bigl\{ h_u^{(t-1)} :  u \in \mathcal{N}(v) \Bigr\} \Bigr) \\
        h_v^{(t)} &= \text{COMBINE}^{(t)}  \Bigl( h_v^{(t-1)}, m_v^{(t)}  \Bigr)
    \end{split}
    \label{eq:gnn_general}
\end{equation}
By defining different $\text{AGGREGATE}^{(t)}$ and $\text{COMBINE}^{(t)}$ functions, we obtain a different GNN variant.
For the GNN to be end-to-end trainable, both functions need to be differentiable.
Furthermore, since there is no natural ordering of the neighbors of a vertex, the $\text{AGGREGATE}^{(t)}$ function must be permutation invariant.
Note that the neighborhood aggregation procedure is closely related to the Weisfeiler-Lehman test of isomorphism and the Weisfeiler-Lehman framework.
More specifically, the number of neighborhood aggregation layers is analogous to the number of iterations of the Weisfeiler-Lehman framework.
Furthermore, in the case of the Weisfeiler-Lehman framework, the employed $\text{AGGREGATE}$ function computes the sorted set of labels of vertex $v$'s neighbors, while the $\text{COMBINE}$ function adds the label of the vertex $v$ itself as the first element of the above set.

To compute a representation for the entire graph, GNNs apply a $\text{READOUT}$ function to vertex representations generated by the final neighborhood aggregation layer to obtain a vector representation over the whole graph
\begin{equation}
    h_G = \text{READOUT} \Bigl( \Bigl\{ h_v^{(T)} :  v \in V \Bigr\} \Bigr)
    \label{eq:readout}
\end{equation}
The $\text{READOUT}$ function needs also to be differentiable and permutation invariant.
Common readout functions include the sum, mean and max aggregators.
These aggregators are different than the one employed by the Weisfeiler-Lehman subtree kernel which produces the histogram of the labels encountered during the different iterations of the algorithm.

Note that most standard GNNs are less powerful than the Weisfeiler-Lehman test of isomorphism in terms of distinguishing non-isomorphic graphs.
In fact, it has been shown that if the $\text{AGGREGATE}^{(t)}, \text{COMBINE}^{(t)}$ and $\text{READOUT}$ functions are injective, then the emerging GNN model maps two graphs that the Weisfeiler-Lehman test of isomorphism decides as non-isomorphic, to different embeddings \shortcite{xu2019powerful}.
To achieve greater expressive power, some models have capitalized on high-order variants of the Weisfeiler-Lehman test of isomorphism \shortcite{morris2019weisfeiler,morris2020weisfeiler}.

We next provide more details about four models which we employ in our experimental evaluation, namely Deep Graph Convolutional Neural Network (DGCNN) \shortcite{zhang2018end}, GraphSAGE \shortcite{hamilton2017inductive}, Differentiable Graph Pooling (DiffPool) \shortcite{ying2018hierarchical}, and Graph Isomorphism Network (GIN) \shortcite{xu2019powerful}.
Note that for clarity of presentation, in what follows, we omit biases.

\subsubsection{Deep Graph Convolutional Neural Network}
This model integrates the $\text{AGGREGATE}^{(t)}$ and $\text{COMBINE}^{(t)}$ functions into a single function as follows
\begin{equation}
  h_v^{(t+1)} = f \left( \sum_{u \in \mathcal{N}(v) \cup \{v\}} \frac{h_u^{(t)}}{1+deg(v)} W^{(t)} \right)
\end{equation}
where $f$ is a nonlinear activation function.
Thus, the model aggregates vertex information in local neighborhoods to extract local substructure information.
After $T$ iterations, the model concatenates the outputs $h_v^{(t)}$, for $t=1,\ldots,T$ horizontally to form a concatenated output
\begin{equation}
  h_v = \big( h_v^{(1)}, h_v^{(2)}, \ldots, h_v^{(T)} \big)
\end{equation}
To generate a representation for the entire graph, the model uses a SortPooling layer which imposes an order on the vertices of the graph.
More specifically, vertices are sorted in a descending order based on the last component of their representations (\ie $h_v$ for vertex $v$), while vertices that have the same value in the last component are compared based on the second to last component, and so on.
Furthermore, to allow the model to handle graphs with different numbers of vertices, this layer unifies the sizes of the outputs for different graphs by truncating/extending the output tensor in the first dimension from $n$ to $k$.
Output is then passed on to a traditional convolutional neural network.

\subsubsection{GraphSAGE}
The GraphSAGE  model can deal with very large graphs since it does not take into account all neighbors of a vertex, but uniformly samples a fixed-size set of neighbors.
Let $\mathcal{N}^k(v)$ be a uniformly drawn subset (of size $k$) from the set $\mathcal{N}(v)$ of a vertex $v$.
The neighborhood aggregation scheme of GraphSAGE is defined as follows
\begin{equation}
  \begin{split}
      m_v^{(t)} &= \text{AGGREGATE}^{(t)}\Big( \Big\{h_u^{(t)} \big| u \in \mathcal{N}^k(v) \Big\} \Big) \\
      h_v^{(t+1)} &= \sigma \Big( W^{(t)} \big( h_v^{(t)} , m_v^{(t)} \big)\Big) \\
      h_v^{(t+1)} &= \frac{h_v^{(t+1)}}{\big|\big|h_v^{(t+1)}\big|\big|_2}
  \end{split}
\end{equation}
where $\big( h_v^{(t)} , m_v^{(t)} \big)$ denotes the concatenation of the two input vectors.
The model draws different uniform samples at each iteration, while it uses one of the following aggregation functions:\\
\noindent($1$) Mean aggregator: the mean operator computes the elementwise mean of the representations of the neighbors and the vertex itself (the concatenation step shown above is skipped)
\begin{equation}
  h_v^{(t+1)} = \sigma \left( W^{(t)} \frac{\sum_{u \in \mathcal{N}^k(v) \cup \{v\}}h_u^{(t)}}{deg(v)+1} \right)
\end{equation}
\noindent($2$) Long short-term memory aggregator: the representations of the neighbors are passed on to an long short-term memory (LSTM) architecture.
However, LSTMs are not permutation invariant.\\
\noindent($3$) Pooling aggregator: an elementwise max-pooling operation is applied to aggregate information across the neighbor set
\begin{equation}
  \text{AGGREGATE}^{(t)} = \max \Big( \Big\{ \sigma \big(W_{\text{pool}}^{(t)} h_u^{(t)} \big) \big| u \in \mathcal{N}^k(v) \Big\} \Big)
\end{equation}
where $\max$ denotes the elementwise max operator.

\subsubsection{Differentiable Graph Pooling}
This model aggregates information in a hierarchical way to capture the structure of the entire graph.
More specifically, for each layer, the model learns a soft assignment of the vertices of that layer to those of the next layer.
This soft assignment considers both topological and feature information.
Formally, a matrix $S^{(t)} \in \mathbb{R}^{n_t \times n_{t+1}}$ is associated with each layer of the model which corresponds to the learned cluster assignment matrix at layer $t$.
Each row corresponds to one of the $n_t$ vertices (or clusters) at layer $t$ and each column to one of the $n_{t+1}$ clusters of the next layer $t+1$.
Matrix $S^{(t)}$ provides a soft assignment of each vertex at layer $t$ to a cluster in the next coarsened layer $t+1$.
Each layer coarsens the input graph as follows
\begin{equation}
  \begin{split}
    X^{(t+1)} &= {S^{(t)}}^{\!\!\top} Z^{(t)} \\
    A^{(t+1)} &= {S^{(t)}}^{\!\!\top} A^{(t)} S^{(t)}
  \end{split}
\end{equation}
where $A^{(t+1)}$ is the coarsened adjacency matrix, and $X^{(t+1)}$ is a matrix of embeddings for each vertex/cluster.
To generate the assignment matrix $S^{(t)}$ and matrix $Z^{(t)}$, the model utilizes two separate message passing neural networks.
Both are applied to the input cluster vertex features $X^{(t)}$ and coarsened adjacency matrix $A^{(t)}$ as follows
\begin{equation}
    \begin{split}
      Z^{(t)} &= \text{GNN}_{\text{embed}}^{(t)} \big( A^{(t)}, X^{(t)} \big) \\
      S^{(t)} &= \text{softmax} \big( \text{GNN}_{\text{pool}}^{(t)} (A^{(t)}, X^{(t)}) \big)
    \end{split}
\end{equation}
where the softmax function is applied in a row-wise fashion.
$\text{GNN}_{\text{embed}}^{(t)}$ generates new representations for the input vertices, while $\text{GNN}_{\text{pool}}^{(t)}$ generates a probabilistic assignment of the input vertices to $n_{t+1}$ clusters.
To generate a final embedding vector corresponding to the entire graph, the model sets the final assignment matrix equal to a vector of ones, that is all vertices at the final layer $T$ are assigned to a single cluster. 

\subsubsection{Graph Isomorphism Network}
The neighborhood aggregation operation in MPNNs can be thought of as an aggregation function over the multiset that contains the representations of the neighbors of a given vertex. 
Specifically, a multiset is a generalized concept of a set that allows multiple instances for its elements.
When node features are from a countable universe, both the representations of all vertices of a graph and the representations of the neighbors of a vertex can be thought of as multisets \shortcite{xu2019powerful}.
Furthemore, the representations of vertices that emerge at deeper layers of a model are also from a countable universe \shortcite{xu2019powerful}.
Importantly, an MPNN can map two graphs that the Weisfeiler-Lehman test of isomorphism decides as non-isomorphic to different embeddings if the $\text{AGGREGATE}$, $\text{COMBINE}$ and $\text{READOUT}$ functions of the model are all injective \shortcite{xu2019powerful}.
It turns out that the sum aggregator is an injective multiset function.
Based on the above result, the graph isomorphism network utilizes the sum aggregator to model injective multiset functions for the neighborhood and vertex aggregation, and has thus the same power as the Weisfeiler-Lehman test of isomorphism.
Each neighborhood aggregation layer is defined as
\begin{equation}
    h_v^{(t+1)} = \text{MLP}^{(t)} \Big( \big( 1 + \epsilon^{(t)} \big) h_v^{(t)} + \sum_{u \in \mathcal{N}(v)} h_u^{(t)} \Big)
\end{equation}
where $\epsilon^{(t)}$ is an irrational number of layer $t$ and $\text{MLP}^{(t)}$ is a multi-layer perceptron of layer $t$.
The model also uses the sum aggregator as its readout function.
Let $h_G^{(t)} = \sum_{v \in G} h_v^{(t)}$ denote the sum of vertex representations at layer $t$.
To produce a graph-level representation, the model utilizes the following readout function which concatenates information from all neighborhood aggregation layers
\begin{equation}
    h_G = \Big( h_G^{(0)}, h_G^{(1)}, \ldots, h_G^{(T)} \Big)
\end{equation}

\subsection{Other Models}
While graph kernels focus on several different structural aspects of graphs (\eg walks, subgraphs, cycles, etc.), the same does not hold for GNNs since most of these models are members of the family of MPNNs.
However, there are exceptions to this ``rule'', and some of these models draw inspiration from graph kernels.
For instance, \shortciteA{lei2017deriving} proposed a class of GNNs and characterized their associated kernel spaces which were found to be associated with either the random walk kernel or the Weisfeiler-Lehman subtree kernel.
Specifically, the hidden states of these models live in the reproducing kernel Hilbert space of these kernels.
In another study, \shortciteA{chen2020convolutional} generated finite-dimensional vertex representations using the Nystr{\"o}m method to approximate a kernel that compares a set of local patterns centered at vertices.
These representations can be learned without supervision by extracting a set of anchor points, or can be modeled as parameters of a neural network and be learned end-to-end.

In the past years, several approaches have been proposed that combine graph kernels with neural networks.
For instance, \shortciteA{navarin2018pre} used graph kernels to pre-train GNNs, while \shortciteA{nikolentzos2018kernel} used graph kernels to extract features that are then fed to convolutional neural networks.
\shortciteA{du2019graph} followed the opposite direction and proposed a new graph kernel which corresponds to infinitely wide multi-layer GNNs trained by gradient descent, while \shortciteA{al2019ddgk} proposed an unsupervised method for learning graph representations by comparing the input graphs against a set of source graphs.
Finally, \shortciteA{nikolentzos2020random} proposed a neural network model whose first layer consists of a number of latent graphs which are compared against the input graphs using a random walk kernel.
The emerging kernel values are fed into a fully-connected neural network which acts as the classifier or regressor.

\section{Applications of Graph Kernels}\label{sec:applications}
In the past years, graph kernels have been applied successfully to a series of real-world problems.
Most of these problems come from the fields of bioinformatics and chemoinformatics.
However, graph kernels are not limited only to these two fields, but they have been applied to problems arising in other domains as well.
We list below some examples of such fields of application.

\subsection{Chemoinformatics}
Traditionally, chemistry is one of the richest sources of graph-structured data.
A common problem in this field is to find chemical compounds with a specific property or activity.
The experimental characterization of molecules is often an expensive and time-consuming process, and thus people usually resort to computational methods.
Specifically, they model chemical compounds as graphs where vertices correspond to atoms and edges to bonds, and then they apply computational methods to identify a small set of potentially interesting molecules for a given property or activity, which are then tested experimentally.
Graph kernels have been used extensively for predicting the mutagenicity, toxicity and anti-cancer activity of small molecules \shortcite{swamidass2005kernels,ralaivola2005graph,mahe2005graph,ceroni2007classification,mahe2009graph,smalter2009graph}
Furthermore, graph kernels have been applied to other problems such as the prediction of the atomization energies of organic molecules \shortcite{ferre2017learning}, the predicition of the boiling points of molecules \shortcite{gauzere2011two}, the prediction of the activity against HIV \shortcite{gauzere2011two}, and the prediction of properties of stereoisomers \shortcite{brown2010compound,grenier2017chemoinformatics}.
A review of the applications of graph kernels in chemoinformatics is provided by \shortciteA{rupp2010graph}.

\subsection{Bioinformatics}
Bioinformatics is also one of the major application domains of graph representations and therefore, of graph kernels.
Recent advances in technology have delivered a step change in our ability to sequence genomes, measure gene expression levels, and test large numbers of potential regulatory interactions between genes.
Despite these advancements, some problems of high interest such as the experimental determination of the function of a protein still remain both expensive and time-consuming.
Interestingly, the above-mentioned processes produce large volumes of data which can give rise to various types of graphs, such as protein structures, protein and gene co-expression networks, or protein-protein interaction networks.
These graphs can then be processed by computational approaches such as graph kernels, and provide solutions to some of these challenging problems. 
Among others, in the field of bioinformatics, graph kernels have been applied to the prediction of the function of proteins with known sequence and structure \shortcite{borgwardt2005protein,schietgat2015predicting}, to the identification of the interactions that are involved in disease outbreak and progression \shortcite{borgwardt2007graphkernels}, to the analysis of functional non-coding RNA sequences \shortcite{sato2008directed}, to the identification of temporally localized relationships among genes \shortcite{antoniotti2010application}, and to the prediction of domain-peptide interactions \shortcite{kundu2013graph}.

\subsection{Computer Vision}
Graph representations have been investigated a lot in the fields of image processing and computer vision.
There exist many different approaches to represent images as graphs.
For instance, vertices usually correspond to pixels or to segmented regions, while edges join neighboring pixels or neighboring regions with each other.
Graph kernels have served as an effective tool for many computer vision tasks such as for classifying images \shortcite{harchaoui2007image,mahboubi2010object,antanas2012relational,zhang2013fast}, for detecting objects represented as point clouds \shortcite{bach2008graph,neumann2013graph}, for achieving place recognition \shortcite{stumm2016robust}, for achieving action recognition \shortcite{wang2013directed,wu2014human,li20163d}, for scene  modeling \shortcite{fisher2011characterizing}, and for matching observations of persons across different cameras \shortcite{brun2011people}.

Besides the above applications, graphs are also used increasingly often in biomedical imaging.
Different types of graphs such as connectivity graphs are usually extracted from functional magnetic resonance imaging (fMRI) data.
Then, graph kernels capitalize on these graphs to address various tasks such as to distinguish between different brain states \shortcite{shahnazian2012method,mokhtari2013decoding,vega2013brain,vega2014classification}, to determine whether a subject is cocaine-addicted or not \shortcite{gkirtzou2016pyramid}, or to predict mild cognitive impairment, a prodromal stage of Alzheimer's disease \shortcite{jie2014topological,jie2016sub}.
There have also been proposed kernels that can handle inter-subject variability in fMRI data \shortcite{takerkart2014graph}.

\subsection{Cybersecurity and Software Verification}
The number of malicious applications targeting desktop and mobile devices has increased daramatically in the past few years.
Due to this unprecedented increase in the number of malicious applications, malware detection has recently become a very active area of research.
It has been observed that most newly discovered malware samples are variations of existing malware. 
Furthermore, it has been shown that it is easier to detect these variations if high-level code representations, such as function call graphs or control flow graphs, are employed.
It should be mentioned that these graphs can prove useful not only for detecting malware, but also for retrieving similar application programs.
Therefore, graph kernels can be applied to such graphs, and have served as a common tool for detecting malware \shortcite{anderson2011graph,gascon2013structural,narayanan2016contextual}, but also for analyzing execution traces obtained from dynamic analysis \shortcite{wagner2009malware}, for measuring the similarity between programs \shortcite{li2016detecting}, and for predicting metamorphic relations \shortcite{kanewala2016predicting}.

\subsection{Natural Language Processing}
Although textual documents do not exhibit an underlying graph structure, in many cases, they are also modeled as graphs.
A vertex corresponds to some meaningful linguistic unit such as a sentence, a word, or even a character, while an edge corresponds to some relationship between two vertices which can be statistical, syntactic, or semantic among others.
A common representation is the word co-occurence network, where vertices correspond to terms and edges represent co-occurrences between the terms within a fixed-size sliding window.
This representation addresses some of the limitations of the bag-of-words representation which treats terms as independent of one another.
Graph kernels have proven useful for several text mining applications such as for recognizing identical real-world events modeled as event graphs \shortcite{glavavs2013recognizing}, for classifying biomedical text documents represented as concept graphs \shortcite{bleik2013text}, for extracting protein-protein interactions from scientific literature \shortcite{airola2008graph,airola2008all}, and for measuring the similarity between documents represented as word co-occurence networks \shortcite{nikolentzos2017shortest}.
Besides the above applications, tree kernels (which can be considered as instances of graph kernels) have been heavily applied to different problems in the field of natural language processing (see subsection~\ref{sec:tree_kernels} for more details).

\subsection{Other Applications}
Graph kernels have been applied to many other practical problems involving graph representations such as for classifying Resource Description Framework (RDF) data \shortcite{losch2012graph,de2015substructure}, for entity disambiguation in anonymized graphs \shortcite{hermansson2013entity}, for classifying architectural designs into architectural styles \shortcite{strobbe2016automatic}, and for estimating the similarity of relational states in relational reinforcement learning \shortcite{driessens2006graph,halbritter2007learning}.

\section{Experimental Comparison}\label{sec:experiments}
In this Section, we experimentally evaluate many of the graph kernels presented above and compare them to each other.
Although there are approaches that measure the expressiveness of graph kernels by using recent results from the field of statistical learning theory \shortcite{oneto2017measuring}, empirically evaluating the graph kernels can provide insights into their utility in real-world scenarios.
We first present the problem of graph classification, and we evaluate the kernels in this task.
We describe the datasets that we used for our experiments, and give details about the experimental settings.
We then report on the performance and running time of the different kernels.
We finally experiment with a synthetic dataset, and compare the emerging kernel values against the similarities produced by an expressive but intractable graph similarity function.

\subsection{Graph Classification}
Classification is perhaps the most frequently encountered machine learning problem.
In classification, the goal is to learn a mapping from input objects to their class labels, given a training set.
When the input objects are graphs, the problem is called \textit{graph classification}.
More formally, in this setting, we are given a training set $\mathcal{D} = \{ (G_i, y_i)\}_{i=1}^N$ consisting of $N$ graphs along with their class labels.
The goal is to learn a function $f : \mathcal{G} \rightarrow \mathcal{Y}$, where $\mathcal{G}$ is the input space of graphs and $\mathcal{Y}$ the set of graph labels.
This function can then be used to assign class labels to new previously unseen graphs, such as those contained in the test set.

The problem of graph classification has become a popular area of research in recent years because it finds numerous applications in a wide variety of fields.
Several of these application have already been discussed above.
For example, graph classification arises in applications which range from predicting the mutagenicity of a chemical compound \shortcite{swamidass2005kernels}, and predicting the function of a protein given its amino acid sequence \shortcite{borgwardt2005protein}, to detecting if a software object is infected with malware \shortcite{wagner2009malware}.

\subsubsection{Datasets}
We next briefly describe the graph datasets used in our experiments.
We have considered data from different domains, including chemoinformatics, bioinformatics and social networks.
All graphs are undirected.
Furtermore, the graphs contained in the chemoinformatics and bioinformatics datasets are node-labeled, node-attributed or both.
All datasets are publicly available \shortcite{KKMMN2016}.
Table~\ref{tab:dataset_statistics} provides a summary of the employed datasets.
\begin{table}[t]
\centering
\def\arraystretch{1.2}
\resizebox{\textwidth}{!} {
\begin{tabular}{|l|c|c|d{3.2}|d{3.2}|d{5.2}|cc|cc|} \hline
\multirow{4}{*}{Dataset} & \multicolumn{5}{c|}{\multirow{2}{*}{Statistics}} & \multicolumn{4}{c|}{Node Labels/} \\ 
& \multicolumn{5}{c|}{}  & \multicolumn{4}{c|}{Attributes} \\ \cline{2-10} 
& \multirow{2}{*}{\#Graphs} & \multirow{2}{*}{\#Classes} & \multicolumn{1}{c|}{Max Class} & \multicolumn{1}{c|}{Avg.} & \multicolumn{1}{c|}{Avg.} & \multicolumn{2}{c|}{Labels} & \multicolumn{2}{c|}{Attributes} \\
& & & \multicolumn{1}{c|}{Imbalance} & \multicolumn{1}{c|}{\#Nodes} & \multicolumn{1}{c|}{\#Edges} & \multicolumn{2}{c|}{(Num.)} & \multicolumn{2}{c|}{(Dim.)} \\ \hline
AIDS                          & 2,000           & 2      & 1:4.0   & 15.69                & 16.20                & + & (38)      & + &(4)                             \\ \hline
BZR                           & 405             & 2      & 1:3.70  & 35.75                & 38.36                & + & (10)      & + &(3)                             \\ \hline
COLLAB                        & 5,000           & 3      & 1:3.35  & 74.49                & 2,457.78             & --&          & --&                                 \\ \hline
D\&D                            & 1,178           & 2      & 1:1.41  & 284.32               & 715.66               & + & (82)      & --&                                 \\ \hline
ENZYMES                       & 600             & 6      & 1:1     & 32.63                & 62.14                & + & (3)       & + &(18)                            \\ \hline
IMDB-BINARY                   & 1,000           & 2      & 1:1     & 19.77                & 96.53                & --&          & --&                                 \\ \hline
IMDB-MULTI                    & 1,500           & 3      & 1:1     & 13.00                & 65.94                & --&          & --&                                 \\ \hline
MUTAG                         & 188             & 2      & 1:1.98  & 17.93                & 19.79                & + & (7)       & --&                                 \\ \hline
NCI1                          & 4,110           & 2      & 1:1     & 29.87                & 32.30                & + & (37)      & --&                                 \\ \hline
PROTEINS                      & 1,113           & 2      & 1:1.47  & 39.06                & 72.82                & + &(3)       & + &(1)                             \\ \hline
PROTEINS\_full                & 1,113           & 2      & 1:1.47  & 39.06                & 72.82                & + &(3)       & + &(29)                            \\ \hline
PTC-MR                        & 344             & 2      & 1:1.26  & 14.29                & 14.69                & + &(19)      & --&                                 \\ \hline
REDDIT-BINARY                 & 2,000           & 2      & 1:1     & 429.63               & 497.75               & --&          & --&                                 \\ \hline
REDDIT-MULTI-5K               & 4,999           & 5      & 1:1     & 508.52               & 594.87               & --&         & -- &                               \\ \hline
REDDIT-MULTI-12K              & 11,929          & 11     & 1:5.05  & 391.41               & 456.89               & --&          & -- &                                \\ \hline
SYNTHETICnew                  & 300             & 2      & 1:1     & 100.00               & 196.25               & --&          & + &(1)                             \\ \hline
Synthie                       & 400             & 4      & 1:1.22  & 95.00                & 172.93               & --&          & + &(15)                            \\ \hline
\end{tabular}
}
\caption{Summary of the $17$ datasets used in our experiments. The ``Max Class Imbalance'' column indicates the ratio of the size of the smallest class of the dataset to the size of its largest class.}
\label{tab:dataset_statistics}
\end{table}

\paragraph{AIDS.} It consists of molecular compounds represented as graphs.
The compounds were obtained from the AIDS Antiviral Screen Database of Active Compounds.
Vertices correspond to atoms and edges to covalent bonds.
Vertices are labeled with the corresponding chemical symbol and edges with the valence of the linkage.
The task is to predict whether or not each compound is active against HIV \shortcite{riesen2008iam}.
\paragraph{BZR.} It contains $405$ chemical compounds (ligands for the benzodiazepine receptor) which are modeled as graphs.
The task is to predict whether a compound is active or inactive \shortcite{sutherland2003spline}.
\paragraph{COLLAB.} This is a scientific collaboration dataset consisting of the ego-networks of several researchers from three subfields of Physics (High Energy Physics, Condensed Matter Physics and Astro Physics).
The task is to determine the subfield of Physics to which the ego-network of each researcher belongs \shortcite{yanardag2015deep}. 
\paragraph{D\&D.} This dataset contains over a thousand protein structures.
Each protein is a graph whose vertices correspond to amino acids and a pair of amino acids are connected by an edge if they are less than $6$ \AA ngstroms apart.
The task is to predict if a protein is an enzyme or not \shortcite{dobson2003distinguishing}.
\paragraph{ENZYMES.} It comprises of $600$ protein tertiary structures obtained from the BRENDA enzyme database.
Each enzyme is a member of one of the Enzyme Commission top level enzyme classes (EC classes) and the task is to correctly assign the enzymes to their classes \shortcite{borgwardt2005protein}.
\paragraph{IMDB-BINARY and IMDB-MULTI.} These datasets were created from IMDb (\url{www.imdb.com}), an online database of information related to movies and television programs. 
The graphs contained in the two datasets correspond to movie collaborations.
The vertices of each graph represent actors/actresses and two vertices are connected by an edge if the corresponding actors/actresses appear in the same movie.
Each graph is the ego-network of an actor/actress, and the task is to predict which genre an ego-network belongs to \shortcite{yanardag2015deep}.
\paragraph{MUTAG.} This dataset consists of $188$ mutagenic aromatic and heteroaromatic nitro compounds.
The task is to predict whether or not each chemical compound has mutagenic effect on the Gram-negative bacterium {\it Salmonella typhimurium} \shortcite{debnath1991structure}.
\paragraph{NCI1.} This dataset contains a few thousand chemical compounds screened for activity against non-small cell lung cancer and ovarian cancer cell lines \shortcite{wale2008comparison}.
\paragraph{PROTEINS, PROTEINS\_full.} They contain proteins represented as graphs where vertices are secondary structure elements and there is an edge between two vertices if they are neighbors in the amino-acid sequence or in $3$D space.
The task is to classify proteins into enzymes and non-enzymes \shortcite{borgwardt2005protein}. 
\paragraph{PTC-MR.} This dataset contains $344$ organic molecules represented as graphs.
The task is to predict their carcinogenic effects on male rats \shortcite{toivonen2003statistical}. 
\paragraph{REDDIT-BINARY, REDDIT-MULTI-5K, REDDIT-MULTI-12K.} The graphs contained in these three datasets represent social interaction between users of Reddit (\url{www.reddit.com}), one of the most popular social media websites.
Each graph represents an online discussion thread.
Specifically, each vertex corresponds to a user, and two users are connected by an edge if one of them responded to at least one of the other's comments.
The task is to classify graphs into either communities or subreddits \shortcite{yanardag2015deep}.
\paragraph{SYNTHETICnew.} It comprises of $300$ synthetic graphs divided into two classes of equal size.
Each graph is obtained by adding noise to a random graph with $100$ vertices and $196$ edges, whose vertices are endowed with normally distributed scalar attributes sampled from $\mathcal{N}(0,1)$.
The graphs of the first class were generated by rewiring $5$ edges and permuting $10$ node attributes, while the graphs of the second class were generated by rewiring $10$ edges and permuting $5$ node attributes.
After the generation of all graphs, noise from $\mathcal{N}(0,0.45^2)$ was also added to every node attribute in every graph \shortcite{feragen2013scalable}.
\paragraph{Synthie.} This dataset consists of $400$ synthetic graphs, subdivided into four classes, with $15$ real-valued node attributes.
Two types of graphs and two types of attributes were generated, and each combination of those gave rise to a class (four classes in total).
All graphs were generated by randomly adding edges between $10$ perturbed instances of two Erd{\"o}s R{\'e}nyi graphs.
To generate graphs of the first type, perturbed instances of the first Erd{\"o}s R{\'e}nyi graph were choosen with probability $0.8$, while perturbed instances of the second Erd{\"o}s R{\'e}nyi graph were choosen with probability $0.2$.
To generate graphs of the second type, the two probabilities were reversed.
The vertices of each graph were then annotated by attributes drawn either from the first or from the second type of attributes \shortcite{morris2016faster}.

\subsubsection{Experimental Setup}
We evaluated the performance of the graph kernels on the datasets presented above. 
Specifically, we made use of the GraKeL library which contains implementations of a large number of graph kernels \shortcite{siglidis2020grakel}.
We used the following $20$ kernels in our experimental evaluation: ($1$) vertex histogram kernel (VH), ($2$) random walk kernel (RW), ($3$) shortest path kernel (SP), ($4$) graphlet kernel (GR), ($5$) Weisfeiler-Lehman subtree kernel (WL-VH), ($6$) Weisfeiler-Lehman shortest path kernel (WL-SP), ($7$) Weisfeiler-Lehman pyramid match kernel (WL-PM), ($8$) Weisfeiler-Lehman optimal assignment kernel (WL-OA), ($9$) neighborhood hash kernel (NH), ($10$) neighborhood subgraph pairwise distance kernel (NSPDK), ($11$) Lov\'asz $\vartheta$ kernel (Lo-$\vartheta$), ($12$) SVM-$\vartheta$ kernel (SVM-$\vartheta$), ($13$) ordered decompositional DAGs with subtree kernel (ODD-STh), ($14$) pyramid match kernel (PM), ($15$) GraphHopper kernel (GH), ($16$) subgraph matching kernel (SM), ($17$) propagation kernel (PK), ($18$) multiscale Laplacian kernel (ML), ($19$) core Weisfeiler-Lehman subtree kernel (CORE-WL-VH), and ($20$) core shortest path kernel (CORE-SP).
Note that some of the kernels (\eg WL-SP, CORE-SP) correspond to frameworks applied to graph kernels.
Furthermore, since some kernels can handle different types of graphs than others, we conduct three distinct experiments.
The three experiments are characterized by the types of graphs contained in the employed datasets: ($1$) datasets with unlabeled graphs, ($2$) datasets with node-labeled graphs, and ($3$) datasets with node-attributed graphs.
It is important to mention that kernels that are designed for node-labeled graphs can also be applied to unlabaled graphs by initializing the node labels of all vertices of the unlabaled graphs to the same value.
Hence, we evaluate these kernels on datasets that contain node-labeled graphs, but also on datasets that contain unlabeled graphs.
Moreover, kernels that are designed for node-attributed graphs can be applied to unlabeled graphs and to graphs that contain discrete node labels.
To achieve that, in the case of unlabeled graphs, all the vertices of all graphs are assigned the same attribute, while in the case of node-labeled graphs, node labels are transformed into feature vectors (\eg using a ``one-hot'' encoding scheme).
Hence, we evaluated these kernels on all three experimental scenarios.

We also compare the above graph kernels against the following $4$ state-of-the-art GNNs: ($1$) DGCNN \shortcite{zhang2018end}, ($2$) GraphSAGE \shortcite{hamilton2017inductive}, ($3$) DiffPool \shortcite{ying2018hierarchical}, and ($4$) GIN \shortcite{xu2019powerful}.
Note that GNNs iteratively update the feature vectors of the vertices of each graph, and thus, they assume node-attributed graphs.
Therefore, for unlabeled graphs and for graphs that contain discrete node labels, we follow the procedure described above.
In the cased of unlabeled graphs, all the vertices of all graphs are assigned the same attribute, while in the case of node-labeled graphs, node labels are transformed into feature vectors.
It should be mentioned that we only compare the performance of GNNs against that of kernels, and not their running time.
The running time of a GNN depends on the values of some hyperparameters (\eg number of epochs, batch size, etc.), while the different models run on a GPU instead of a CPU.
Hence, the running time of a GNN is not directly comparable to that of a graph kernel, and thus we refrain from reporting those results.

In the case of graph kernels, to perform graph classification, we employed a Support Vector Machine (SVM) classifier and in particular, the LIB-SVM implementation \shortcite{chang2011libsvm}.
To evaluate the performance of the different kernels and GNNs, we employ the framework proposed by \shortciteA{errica2020fair}.
Therefore, we perform $10$-fold cross-validation to obtain an estimate of the generalization performance of each method.
For the common datasets, we use the splits (and results) provided by \shortciteA{errica2020fair}.
For the remaining datasets, we use the code provided by \shortciteA{errica2020fair} to evaluate the $4$ GNNs.
Within each fold, the parameter $C$ of the SVM and the hyperparameters of the kernels (see below) and GNNs were chosen based on a validation experiment on a single $90\%-10\%$ split of the training data.
We chose the value of parameter $C$ from $\{10^{-7},10^{-5},\ldots,10^5,10^7\}$.
Moreover, we normalized all kernel values as follows $\hat{k}(G_i, G_j) = \nicefrac{k(G_i,G_j)}{\sqrt{k(G_i, G_i) \, k(G_j, G_j)}}$ for any graphs $G_i, G_j$.
All experiments were performed on a cluster of $80$ Intel\textsuperscript{\textcopyright} Xeon\textsuperscript{\textcopyright} CPU E$7-4860$ @ $2.27$GHz with $1$TB RAM.
Note that each kernel was computed on a single thread of the cluster.
We set a time limit of $24$ hours for each kernel to compute the kernel matrix.
Hence, we denote by \texttt{TIMEOUT} kernel computations that did not finish within one day.
We also set a memory limit of $64$GB, and we denote by \texttt{OUT-OF-MEM} computations that exceeded this limit.

\begin{table}[t]
\centering
\scriptsize
\def\arraystretch{1.2}
\begin{tabular}{|l|c|c|} \hline
\multirow{2}{*}{Kernels} & \multicolumn{2}{c|}{Hyperparameters} \\ \cline{2-3} 
                         & Fixed          & Chosen based on validation set performance          \\ \hline
VH                       &   --                & --                 \\ \hline
RW                      &$\lambda = 10^{\lceil \log_{10}(\frac{1}{deg^{2}_{\max}}) \rceil} $                 &   $k \in \{2, \dots 10, \infty \}$              \\ \hline
SP                       &--                   &--                  \\ \hline
GR                       &$k=5$                   &$n_{\text{samples}}=\{200, 500, 1000, 2000, 5000\}$                  \\ \hline
WL                       &  --                 & $h \in \{4, \dots, 8\} $                 \\ \hline
WL-OA                    &  --                 & $h \in \{4, \dots, 8\} $                 \\ \hline
NH                       &Count-sensitive neighborhood hash             & $h\in\{1, \dots, 6\}$                 \\ \hline
NSPDK                    &--                   &  $r^* \in\{1, \dots, 6\}, \ d^* \in\{3, \dots, 7\}$               \\ \hline
Lo-$\vartheta$      &$2\leq|S|\leq 8$                &  $n_{\text{samples}}=\{100, 200, 500, 1000\}$             \\ \hline
SVM-$\vartheta$  &$2\leq|S|\leq 8$  &  $n_{\text{samples}}=\{100, 200, 500, 1000\}$                \\ \hline
ODD-STh                  &--                   & $h \in\{1, \dots, 11\}$                \\ \hline
PM                       &--                   &$L \in \{2, 4, 6\}, \ d \in \{4, 6, 8, 10\}$                  \\ \hline
GH                       &--                   &linear kernel/gaussian kernel                 \\ \hline
SM                       &$k=3$                   &    --              \\ \hline
PK                      &$w=10^{-5}$                   & $T \in \{1, \dots, 6\}$                \\ \hline
ML                       &$\gamma=0.01, \ \eta=0.01, \ \hat{p}=10$                &  $l_{max} \in \{0, \dots, 5\}, \ \tilde{c} \in\{50, 100, 200, 300\} $                \\ \hline
CORE                     &  --                 & --                \\ \hline
\end{tabular}
\caption{Values of the hyperparameters of the graphs kernels and frameworks included in our experimental comparison. Note that for some kernels, only a subset of the hyperparameters was optimized, while the rest of the hyperparameters were kept fixed.}
\label{tab:kernel_parametrization}
\end{table}

As mentioned above, to choose the hyperparameters of the kernels, we performed a validation experiment on a single $90\%-10\%$ split of the training set.
Hence, given a kernel, for each combination of hyperparameter values, we generated a seperate kernel matrix.
The hyperparameter values that result into the classifier with the best performance on the validation set are the ones selected for the final model learning.
The values of the different hyperparameters of the kernels are shown in Table~\ref{tab:kernel_parametrization}.
It is interesting to mention that the number of hyperparameters ranges significantly across kernels.
For instance, some kernels such as the vertex historgram kernel (VH) lack hyperparameters, while other kernels such as the multiscale Laplacian kernel (ML) contain a large number of hyperparameters.
Hence, for the vertex historgram (VH) kernel, we compute only a single kernel matrix in each experiment, while for the multiscale Laplacian (ML) kernel, we compute $24$ different kernel matrices in each experiment.
Note also that instead of performing cross-validation to identify the best combination of hyperparameter values, we could have applied multiple kernel learning to the generated kernel matrices \shortcite{massimo2016hyper}.

For each experiment, we report the average accuracy over the $10$ runs of the cross-validation procedure.
Furthermore, we report running times averaged over the $10$ independent runs.
For each run, we compute running times as follows: for each fold of a $10$-fold cross-validation experiment, the running time of the kernel corresponds to the running time for the computation of the kernel matrix that performed best on the validation experiment.

\subsubsection{Experimental Results}
We next present our experimental results.
As mentioned above, we evaluate the graph kernels by performing graph classification on unlabeled, node-labeled and node-attributed benchmark datasets.

\paragraph{Node-Labeled Graphs.}

\begin{table}[!t]
\centering
\scriptsize
\def\arraystretch{1.05}
\begin{tabular}{llcccc} \toprule
& \multirow{3}{*}{Methods} & \multicolumn{4}{c}{DATASETS} \\ \cline{3-6}
& & \multirow{2}{*}{MUTAG} & \multirow{2}{*}{ENZYMES} & \multirow{2}{*}{NCI1} & \multirow{2}{*}{PTC-MR} \\
& & & & & \\ 
\midrule
\parbox[t]{2mm}{\multirow{17}{*}{\rotatebox[origin=c]{90}{Kernels}}} & VH & 69.1 {\tiny ($\pm$ 4.1)} & 20.0 {\tiny ($\pm$ 4.8)} & 55.7 {\tiny ($\pm$ 2.0)} & 57.1 {\tiny ($\pm$ 9.6)} \\ 
& RW & 81.4 {\tiny ($\pm$ 8.9)} & 16.7 {\tiny ($\pm$ 1.8)} & \texttt{TIMEOUT} & 54.4 {\tiny ($\pm$ 9.8)} \\ 
& SP & 82.4 {\tiny ($\pm$ 5.5)} & 37.3 {\tiny ($\pm$ 8.7)} & 72.5 {\tiny ($\pm$ 2.0)} & 60.2 {\tiny ($\pm$ 9.4)} \\ 
& WL-VH & 86.7 {\tiny ($\pm$ 7.3)} & 50.7 {\tiny ($\pm$ 7.3)} & 85.2 {\tiny ($\pm$ 2.2)} & 64.9 {\tiny ($\pm$ 6.4)} \\ 
& WL-SP & 81.4 {\tiny ($\pm$ 8.7)} & 27.3 {\tiny ($\pm$ 7.4)} & 60.8 {\tiny ($\pm$ 2.4)} & 54.5 {\tiny ($\pm$ 9.8)} \\ 
& WL-PM & 88.3 {\tiny ($\pm$ 7.1)} & 57.5 {\tiny ($\pm$ 6.8)} & 85.6 {\tiny ($\pm$ 1.7)} & 65.1 {\tiny ($\pm$ 7.5)} \\ 
& WL-OA & 87.2 {\tiny ($\pm$ 5.4)} & 58.0 {\tiny ($\pm$ 5.0)} & 86.3 {\tiny ($\pm$ 1.6)} & 65.7 {\tiny ($\pm$ 9.6)} \\ 
& NH & 88.3 {\tiny ($\pm$ 6.3)} & 54.5 {\tiny ($\pm$ 3.6)} & 84.7 {\tiny ($\pm$ 1.9)} & 63.4 {\tiny ($\pm$ 9.2)} \\ 
& NSPDK & 85.6 {\tiny ($\pm$ 8.9)} & 42.2 {\tiny ($\pm$ 8.0)} & 74.3 {\tiny ($\pm$ 2.1)} & 59.1 {\tiny ($\pm$ 7.3)} \\ 
& ODD-STh & 80.4 {\tiny ($\pm$ 8.8)} & 32.3 {\tiny ($\pm$ 4.8)} & 75.2 {\tiny ($\pm$ 2.0)} & 59.4 {\tiny ($\pm$ 9.8)} \\ 
& PM & 85.1 {\tiny ($\pm$ 5.8)} & 43.2 {\tiny ($\pm$ 5.3)} & 73.5 {\tiny ($\pm$ 1.9)} & 60.2 {\tiny ($\pm$ 8.2)} \\ 
& GH & 82.5 {\tiny ($\pm$ 5.8)} & 37.2 {\tiny ($\pm$ 6.6)} & 71.0 {\tiny ($\pm$ 2.3)} & 60.2 {\tiny ($\pm$ 9.4)} \\ 
& SM & 85.7 {\tiny ($\pm$ 5.8)} & 35.7 {\tiny ($\pm$ 5.5)} & \texttt{TIMEOUT} & 60.2 {\tiny ($\pm$ 6.8)} \\ 
& PK & 76.6 {\tiny ($\pm$ 5.2)} & 44.0 {\tiny ($\pm$ 6.3)} & 82.1 {\tiny ($\pm$ 2.1)} & 65.1 {\tiny ($\pm$ 5.6)} \\ 
& ML & 87.2 {\tiny ($\pm$ 7.5)} & 48.5 {\tiny ($\pm$ 7.8)} & 79.7 {\tiny ($\pm$ 1.8)} & 64.5 {\tiny ($\pm$ 5.8)} \\ 
& CORE-WL-VH & 85.6 {\tiny ($\pm$ 6.5)} & 51.7 {\tiny ($\pm$ 7.0)} & 85.2 {\tiny ($\pm$ 2.2)} & 65.5 {\tiny ($\pm$ 5.6)} \\ 
& CORE-SP & 85.1 {\tiny ($\pm$ 6.8)} & 39.5 {\tiny ($\pm$ 9.3)} & 73.8 {\tiny ($\pm$ 1.4)} & 57.3 {\tiny ($\pm$ 9.7)} \\
\midrule
\parbox[t]{2mm}{\multirow{4}{*}{\rotatebox[origin=c]{90}{GNNs}}} & DGCNN & 84.0 {\tiny ($\pm$ 7.1)} & 46.3 {\tiny ($\pm$ 6.3)} & 76.4 {\tiny ($\pm$ 1.7)} & 59.5 {\tiny ($\pm$ 6.9)} \\ 
& GraphSAGE & 83.6 {\tiny ($\pm$ 9.6)} & 46.1 {\tiny ($\pm$ 5.4)} & 76.0 {\tiny ($\pm$ 1.8)} & 61.7 {\tiny ($\pm$ 4.9)} \\ 
& DiffPool & 79.8 {\tiny ($\pm$ 6.7)} & 50.7 {\tiny ($\pm$ 8.7)} & 76.9 {\tiny ($\pm$ 1.9)} & 61.1 {\tiny ($\pm$ 5.6)} \\ 
& GIN & 84.7 {\tiny ($\pm$ 6.7)} & 44.5 {\tiny ($\pm$ 4.1)} & 80.0 {\tiny ($\pm$ 1.4)} & 59.1 {\tiny ($\pm$ 7.0)} \\
\bottomrule
\end{tabular}
\vspace{.1cm}
\\
\begin{tabular}{llcccc} \toprule
& \multirow{3}{*}{Methods} & \multicolumn{3}{c}{DATASETS} & \multicolumn{1}{c}{\multirow{2}{*}{Avg.}} \\ \cline{3-5}
& & \multirow{2}{*}{D\&D} & \multirow{2}{*}{PROTEINS} & \multirow{2}{*}{AIDS} & \multicolumn{1}{c}{\multirow{2}{*}{Rank}} \\ 
& & & & \\ 
\midrule
\parbox[t]{2mm}{\multirow{17}{*}{\rotatebox[origin=c]{90}{Kernels}}} & VH & 74.8 {\tiny ($\pm$ 3.7)} & 71.1 {\tiny ($\pm$ 4.4)} & 80.0 {\tiny ($\pm$ 2.3)} & \multicolumn{1}{c}{18.7} \\ 
& RW & \texttt{OUT-OF-MEM} & 69.5 {\tiny ($\pm$ 5.1)} & 79.7 {\tiny ($\pm$ 2.3)} & \multicolumn{1}{c}{19.9} \\ 
& SP & 77.9 {\tiny ($\pm$ 4.5)} & 74.9 {\tiny ($\pm$ 3.6)} & 99.3 {\tiny ($\pm$ 0.4)} & \multicolumn{1}{c}{10.9} \\ 
& WL-VH & 78.7 {\tiny ($\pm$ 2.3)} & 76.2 {\tiny ($\pm$ 3.5)} & 98.3 {\tiny ($\pm$ 0.8)} & \multicolumn{1}{c}{5.8} \\ 
& WL-SP & 76.0 {\tiny ($\pm$ 3.5)} & 72.1 {\tiny ($\pm$ 3.1)} & 99.0 {\tiny ($\pm$ 0.6)} & \multicolumn{1}{c}{15.9} \\ 
& WL-PM & \texttt{OUT-OF-MEM} & 75.9 {\tiny ($\pm$ 3.8)} & 99.4 {\tiny ($\pm$ 0.2)} & \multicolumn{1}{c}{5.3} \\ 
& WL-OA & 77.6 {\tiny ($\pm$ 3.0)} & 76.2 {\tiny ($\pm$ 3.9)} & 99.2 {\tiny ($\pm$ 0.3)} & \multicolumn{1}{c}{3.7} \\ 
& NH & 74.6 {\tiny ($\pm$ 3.5)} & 75.0 {\tiny ($\pm$ 4.2)} & 99.2 {\tiny ($\pm$ 0.5)} & \multicolumn{1}{c}{6.5} \\ 
& NSPDK & 78.9 {\tiny ($\pm$ 4.7)} & 72.5 {\tiny ($\pm$ 2.9)} & 97.8 {\tiny ($\pm$ 1.1)} & \multicolumn{1}{c}{11.8} \\ 
& ODD-STh & 76.4 {\tiny ($\pm$ 4.5)} & 70.9 {\tiny ($\pm$ 4.1)} & 90.4 {\tiny ($\pm$ 2.0)} & \multicolumn{1}{c}{15.9} \\ 
& PM & 77.9 {\tiny ($\pm$ 3.7)} & 70.9 {\tiny ($\pm$ 4.4)} & 99.7 {\tiny ($\pm$ 0.3)} & \multicolumn{1}{c}{10.6} \\ 
& GH & \texttt{TIMEOUT} & 74.8 {\tiny ($\pm$ 2.4)} & 99.4 {\tiny ($\pm$ 0.3)} & \multicolumn{1}{c}{12.8} \\ 
& SM & \texttt{OUT-OF-MEM} & \texttt{OUT-OF-MEM} & 92.2 {\tiny ($\pm$ 1.8)} & \multicolumn{1}{c}{16.2} \\ 
& PK & 77.7 {\tiny ($\pm$ 4.2)} & 73.1 {\tiny ($\pm$ 4.7)} & 96.3 {\tiny ($\pm$ 1.2)} & \multicolumn{1}{c}{11.1} \\ 
& ML & 78.6 {\tiny ($\pm$ 4.0)} & 74.2 {\tiny ($\pm$ 4.4)} & 98.5 {\tiny ($\pm$ 0.5)} & \multicolumn{1}{c}{7.5} \\ 
& CORE-WL-VH & 79.5 {\tiny ($\pm$ 3.2)} & 76.5 {\tiny ($\pm$ 4.4)} & 98.8 {\tiny ($\pm$ 0.5)} & \multicolumn{1}{c}{4.4} \\ 
& CORE-SP & 79.3 {\tiny ($\pm$ 3.8)} & 76.5 {\tiny ($\pm$ 3.9)} & 99.5 {\tiny ($\pm$ 0.3)} & \multicolumn{1}{c}{8.7} \\ 
\midrule
\parbox[t]{2mm}{\multirow{4}{*}{\rotatebox[origin=c]{90}{GNNs}}} & DGCNN & 76.6 {\tiny ($\pm$ 4.3)} & 73.2 {\tiny ($\pm$ 3.2)} & 99.1 {\tiny ($\pm$ 1.4)} & \multicolumn{1}{c}{10.6} \\ 
& GraphSAGE & 72.9 {\tiny ($\pm$ 2.0)} & 74.3 {\tiny ($\pm$ 3.8)} & 97.7 {\tiny ($\pm$ 0.6)} & \multicolumn{1}{c}{11.9} \\ 
& DiffPool & 75.0 {\tiny ($\pm$ 3.5)} & 72.5 {\tiny ($\pm$ 3.5)} & 99.2 {\tiny ($\pm$ 0.3)} & \multicolumn{1}{c}{11.1} \\ 
& GIN & 75.3 {\tiny ($\pm$ 2.9)} & 72.8 {\tiny ($\pm$ 3.6)} & 98.8 {\tiny ($\pm$ 0.6)} & \multicolumn{1}{c}{11.7} \\
\bottomrule
\end{tabular}
\caption{Average classification accuracy ($\pm$ standard deviation) on the $7$ classification datasets containing node-labeled graphs. The ``Avg. Rank'' column illustrates the average rank of each kernel/GNN. The lower the average rank, the better the overall performance of the kernel/GNN.}
\label{tab:results_labeled}
\end{table}

\begin{table}[t]
\centering
\scriptsize
\def\arraystretch{1.05}
\begin{tabular}{lrrrr} \toprule
\multirow{3}{*}{Kernels} & \multicolumn{4}{c}{DATASETS} \\ \cline{2-5}
& \multicolumn{1}{c}{\multirow{2}{*}{MUTAG}} & \multicolumn{1}{c}{\multirow{2}{*}{ENZYMES}} & \multicolumn{1}{c}{\multirow{2}{*}{NCI1}} & \multicolumn{1}{c}{\multirow{2}{*}{PTC-MR}} \\
& & & & \\
\midrule
VH & 0.01s & 0.04s & 0.84s & 0.02s \\ 
RW & 1m 31.24s & 3h 36m 7.01s & \texttt{TIMEOUT} & 9m 9.27s \\ 
SP & 0.92s & 11.03s & 1m 9.69s & 1.52s \\ 
WL-VH & 0.2s & 3.54s & 7m 5.11s & 0.42s \\ 
WL-SP & 6.3s & 1m 15.06s & 10m 55.37s & 10.78s \\ 
WL-PM & 2m 0.87s & 1h 10m 28.36s & 13h 28m 58.53s & 11m 53.67s \\ 
WL-OA & 0.65s & 21.89s & 2h 27m 25.05s & 3.29s \\ 
NH & 0.98s & 12.65s & 13m 56.45s & 4.35s \\ 
NSPDK & 3.94s & 25.77s & 4m 29.99s & 7.81s \\ 
ODD-STh & 1.49s & 1m 2.86s & 49m 2.76s & 4.2s \\ 
PM & 3.17s & 30.86s & 41m 51.78s & 13.14s \\ 
GH & 24.79s & 15m 35.62s & 3h 43m 7.2s & 1m 33.9s \\ 
SM & 1m 57.25s & 3h 25m 43.59s & \texttt{TIMEOUT} & 4m 19.8s \\ 
PK & 0.53s & 11.71s & 10m 30.02s & 1.79s \\ 
ML & 8m 16.84s & 58m 40.97s & 7h 18m 35.72s & 22m 9.56s \\ 
CORE-WL-VH & 0.63s & 7.95s & 12m 36.28s & 1.01s \\ 
CORE-SP & 2.69s & 48.02s & 3m 16.54s & 3.97s \\
\bottomrule
\end{tabular}
\vspace{.2cm}
\\
\begin{tabular}{lrrrm{1.48cm}} \toprule
\multirow{3}{*}{Kernels} & \multicolumn{3}{c}{DATASETS} & \multicolumn{1}{c}{\multirow{2}{*}{Avg.}} \\ \cline{2-4}
& \multicolumn{1}{c}{\multirow{2}{*}{D\&D}} & \multicolumn{1}{c}{\multirow{2}{*}{PROTEINS}} & \multicolumn{1}{c}{\multirow{2}{*}{AIDS}} & \multicolumn{1}{c}{\multirow{2}{*}{Rank}} \\ 
& & & & \\
\midrule
VH & 0.24s & 0.1s & 0.25s & \multicolumn{1}{c}{1.0} \\ 
RW & \texttt{OUT-OF-MEM} & 51m 10.11s & 1h 51m 56.47s & \multicolumn{1}{c}{15.1} \\ 
SP & 55m 58.79s & 1m 18.91s & 13.93s & \multicolumn{1}{c}{4.7} \\ 
WL-VH & 4m 42.13s & 25.34s & 28.89s & \multicolumn{1}{c}{2.7} \\ 
WL-SP & 6h 42m 57.36s & 6m 55.52s & 1m 22.62s & \multicolumn{1}{c}{10.7} \\ 
WL-PM & \texttt{OUT-OF-MEM} & 6h 20m 51.01s & 6h 44m 21.01s & \multicolumn{1}{c}{15.8} \\ 
WL-OA & 1h 22m 26.27s & 3m 48.99s & 4m 36.18s & \multicolumn{1}{c}{8.6} \\ 
NH & 11m 31.88s & 1m 4.52s & 1m 16.5s & \multicolumn{1}{c}{6.9} \\ 
NSPDK & 5h 15m 23.52s & 6m 35.72s & 56.01s & \multicolumn{1}{c}{8.7} \\ 
ODD-STh & 30m 39.54s & 2m 6.26s & 1m 57.88s & \multicolumn{1}{c}{9.0} \\ 
PM & 4m 55.53s & 1m 15.8s & 5m 33.7s & \multicolumn{1}{c}{9.0} \\ 
GH & \texttt{TIMEOUT} & 3h 44m 19.99s & 38m 48.57s & \multicolumn{1}{c}{13.8} \\ 
SM & \texttt{OUT-OF-MEM} & \texttt{OUT-OF-MEM} & 4h 26m 46.71s & \multicolumn{1}{c}{15.7} \\ 
PK & 7m 29.57s & 45.6s & 1m 46.21s & \multicolumn{1}{c}{5.1} \\ 
ML & 1h 26m 59.75s & 1h 35m 36.4s & 33m 16.63s & \multicolumn{1}{c}{14.1} \\ 
CORE-WL-VH & 1m 53.21s & 1m 11.44s & 1m 15.03s & \multicolumn{1}{c}{4.4} \\ 
CORE-SP & 5h 2m 39.71s & 3m 31.97s & 40.11s & \multicolumn{1}{c}{7.6} \\ 
\bottomrule
\end{tabular}
\caption{Average CPU running time for kernel matrix computation on the $7$ classification datasets containing node-labeled graphs. The ``Avg. Rank'' column illustrates the average rank of each kernel. The lower the average rank, the lower the overall running time of the kernel.}
\label{tab:runtimes_labeled}
\end{table}

Tables~\ref{tab:results_labeled} and~\ref{tab:runtimes_labeled} illustrate average prediction accuracies and average running times of the compared kernels and the GNNs on the datasets that contain node-labeled graphs.
We observe that the kernels that employ some neighborhood aggregation mechanism (\eg the Weisfeiler-Lehman framework) yield very good performance.
Specifically, the WL-OA kernel outperforms all the other kernels on $3$ out of the $7$ datasets (ENZYMES, NCI1, and PTC-MR), while the CORE-WL-VH kernel is the best-performing approach on $2$ out of the remaining $4$ datasets (D\&D and PROTEINS).
Moreover, the WL-VH, WL-PM and NH kernels also achive high accuracies on most datasets.
Surprisingly, WL-SP, although equipped with a neighborhood aggregation scheme, performs much worse than the other kernels which employ the same framework and also much worse than the SP and CORE-SP kernels.
The core framework leads to performance improvements on most datasets.
For instance, in the case of the SP kernel, it leads to better accuracies on all but one dataset.
It is worth mentioning that CORE-SP provides the highest accuracy on the PROTEINS dataset (along with CORE-WL-VH) and second best accuracy on the D\&D and AIDS datasets.
As regards the kernels for graphs with continuous attributes (GH, SM, PK, and ML), most of them failed to produce results comparable to the best-performing kernels.
The only exception is the ML kernel which yielded good results on most datasets.
Moreover, the GH kernel reached the highest accuracy on the AIDS dataset.
It is also interesting to mention that the VH and RW kernels achieved very low accuracy levels.
The $4$ GNN models also failed to yield performance competitive with that of the kernels that employ neighborhood aggregation mechanisms.
Specifically, all GNNs were outperformed by some graph kernel on all $7$ datasets.
DGCNN performed better than the rest of the GNNs, but in most cases, all GNN models achieved similar accuracies to each other.

On most datasets, the variability in the performance of the different kernels is low.
The ENZYMES dataset is an exception to that, since the average accuracy of the best-performing kernel is equal to $58.0\%$, while that of the worst-performing kernel is equal to $16.7\%$.
Furthermore, the AIDS dataset is almost perfectly classified by several kernels, and this raises some concerns about the value of this dataset for graph kernel comparison.

In terms of running time, as expected, VH is the fastest kernel on all datasets.
This kernel computes the dot product on vertex label histograms, hence, its complexity is linear to the number of vertices.
The running time of WL-VH, CORE-WL-VH, SP, PK, and NH is also low compared to the other kernels on most datasets.
Note also that while the worst-case complexity of SP is very high, by employing an explicit computation scheme, the running time of the kernel in real scenarios is very attractive.
We also observe that the ML, RW, SM and WL-PM kernels are very expensive in terms of runtime.
Specifically, the SM kernel failed to compute the kernel matrix on NCI1 within one day, while it exceeded the maximum available memory on two other datasets (D\&D and PROTEINS).
It should be mentioned that the size of the graphs (\ie number of vertices) and the size of the dataset (\ie number of graphs) have a different impact on the running time of the kernels.
For instance, the average running time of the PM kernel is relatively high on datasets that contain small graphs.
However, this kernel is much more competitive on datasets which contain large graphs such as the D\&D dataset on which it was the third fastest kernel.

Overall, when dealing with tasks that involve node-labeled graphs, we suggest to use a kernel that utilizes some neighborhood aggregation mechanism.
For instance, the WL-VH and NH kernels achieve high accuracies and are very efficient even when the size of the graphs and/or the dataset is large.
The WL-OA kernel can potentially outperform the above two kernels, however, it is also more expensive to compute.
From the above experimental evaluation, it is also clear that graph kernels are more effective than GNNs in classifying graphs whose vertices are annotated with discrete labels.
Still, we need to stress that graph kernels do not scale to large datasets (\eg datasets that contain hunderds of thousands of graphs), and this is a limitation inherent to kernel methods in general.
In such scenarios, GNNs should be preferred over graph kernels.

\paragraph{Unabeled Graphs}

\begin{table}[t]
\centering
\scriptsize
\def\arraystretch{1.05}
\resizebox{\textwidth}{!} {
\begin{tabular}{llccccccc} \toprule
& \multirow{3}{*}{Methods} & \multicolumn{6}{c}{DATASETS} & \multirow{2}{*}{Avg.} \\ \cline{3-8}
& & IMDB & IMDB & REDDIT & REDDIT & REDDIT & \multirow{2}{*}{COLLAB} & \multirow{2}{*}{Rank}\\
& & BINARY & MULTI & BINARY & MULTI-5K & MULTI-12K & & \\ 
\midrule
\parbox[t]{2mm}{\multirow{20}{*}{\rotatebox[origin=c]{90}{Kernels}}} & VH & 50.0 {\tiny ($\pm$ 0.0)} & 33.3 {\tiny ($\pm$ 0.0)} & 50.0 {\tiny ($\pm$ 0.0)} & 20.0 {\tiny ($\pm$ 0.0)} & 21.7 {\tiny ($\pm$ 1.5)} & 52.0 {\tiny ($\pm$ 0.1)} & 18.3 \\ 
& RW & 64.1 {\tiny ($\pm$ 4.5)} & 44.6 {\tiny ($\pm$ 4.1)} & \texttt{TIMEOUT} & \texttt{TIMEOUT} & \texttt{TIMEOUT} & 68.0 {\tiny ($\pm$ 1.7)} & 17.2 \\ 
& SP & 58.2 {\tiny ($\pm$ 4.7)} & 39.2 {\tiny ($\pm$ 2.3)} & 81.7 {\tiny ($\pm$ 2.5)} & 47.9 {\tiny ($\pm$ 1.9)} & \texttt{TIMEOUT} & 58.8 {\tiny ($\pm$ 1.2)} & 15.2 \\ 
& GR & 66.1 {\tiny ($\pm$ 2.7)} & 39.5 {\tiny ($\pm$ 2.7)} & 76.1 {\tiny ($\pm$ 2.6)} & 34.7 {\tiny ($\pm$ 2.0)} & 23.0 {\tiny ($\pm$ 1.4)} & 73.0 {\tiny ($\pm$ 2.0)} & 12.8 \\ 
& WL-VH & 70.7 {\tiny ($\pm$ 6.8)} & 51.3 {\tiny ($\pm$ 4.4)} & 67.8 {\tiny ($\pm$ 3.5)} & 50.5 {\tiny ($\pm$ 1.6)} & 38.7 {\tiny ($\pm$ 1.7)} & 78.3 {\tiny ($\pm$ 2.1)} & 6.5 \\ 
& WL-SP & 58.2 {\tiny ($\pm$ 4.7)} & 39.2 {\tiny ($\pm$ 2.3)} & \texttt{TIMEOUT} & \texttt{TIMEOUT} & \texttt{TIMEOUT} & 58.8 {\tiny ($\pm$ 1.2)} & 19.0 \\ 
& WL-PM & 73.6 {\tiny ($\pm$ 3.4)} & 49.1 {\tiny ($\pm$ 5.5)} & \texttt{OUT-OF-MEM} & \texttt{OUT-OF-MEM} & \texttt{OUT-OF-MEM} & \texttt{OUT-OF-MEM} & 14.9 \\ 
& WL-OA & 72.6 {\tiny ($\pm$ 5.5)} & 51.1 {\tiny ($\pm$ 4.3)} & 89.0 {\tiny ($\pm$ 1.3)} & 54.0 {\tiny ($\pm$ 1.2)} & \texttt{TIMEOUT} & 80.5 {\tiny ($\pm$ 2.0)} & 5.8 \\ 
& NH & 71.6 {\tiny ($\pm$ 4.5)} & 50.5 {\tiny ($\pm$ 5.0)} & 81.2 {\tiny ($\pm$ 2.0)} & 49.9 {\tiny ($\pm$ 2.4)} & 39.6 {\tiny ($\pm$ 1.4)} & 81.1 {\tiny ($\pm$ 2.4)} & 5.8 \\ 
& NSPDK & 67.4 {\tiny ($\pm$ 3.3)} & 44.6 {\tiny ($\pm$ 3.8)} & \texttt{TIMEOUT} & \texttt{TIMEOUT} & \texttt{TIMEOUT} & \texttt{TIMEOUT} & 18.2 \\ 
& Lo-$\vartheta$ & 51.0 {\tiny ($\pm$ 4.2)} & 39.8 {\tiny ($\pm$ 2.6)} & \texttt{TIMEOUT} & \texttt{TIMEOUT} & \texttt{TIMEOUT} & \texttt{TIMEOUT} & 20.1 \\ 
& SVM-$\vartheta$ & 52.3 {\tiny ($\pm$ 4.0)} & 39.5 {\tiny ($\pm$ 2.7)} & 74.8 {\tiny ($\pm$ 2.6)} & 31.4 {\tiny ($\pm$ 1.1)} & 22.9 {\tiny ($\pm$ 0.9)} & 52.0 {\tiny ($\pm$ 0.1)} & 15.8 \\ 
& ODD-STh & 65.0 {\tiny ($\pm$ 4.0)} & 46.7 {\tiny ($\pm$ 3.4)} & 52.1 {\tiny ($\pm$ 3.2)} & 43.1 {\tiny ($\pm$ 1.8)} & 30.0 {\tiny ($\pm$ 1.6)} & 52.0 {\tiny ($\pm$ 0.1)} & 13.2 \\ 
& PM & 66.3 {\tiny ($\pm$ 4.2)} & 46.1 {\tiny ($\pm$ 3.8)} & 86.5 {\tiny ($\pm$ 2.1)} & 48.3 {\tiny ($\pm$ 2.5)} & 41.1 {\tiny ($\pm$ 0.6)} & 74.0 {\tiny ($\pm$ 2.4)} & 8.7 \\ 
& GH & 59.4 {\tiny ($\pm$ 3.4)} & 39.5 {\tiny ($\pm$ 2.6)} & \texttt{TIMEOUT} & \texttt{TIMEOUT} & \texttt{TIMEOUT} & 60.0 {\tiny ($\pm$ 1.4)} & 18.1 \\ 
& SM & \texttt{TIMEOUT} & \texttt{TIMEOUT} & \texttt{OUT-OF-MEM} & \texttt{OUT-OF-MEM} & \texttt{OUT-OF-MEM} & \texttt{TIMEOUT} & -- \\ 
& PK & 51.7 {\tiny ($\pm$ 3.7)} & 34.5 {\tiny ($\pm$ 3.0)} & 63.9 {\tiny ($\pm$ 3.0)} & 34.9 {\tiny ($\pm$ 1.7)} & 23.9 {\tiny ($\pm$ 1.2)} & 57.0 {\tiny ($\pm$ 1.2)} & 16.2 \\ 
& ML & 69.9 {\tiny ($\pm$ 4.8)} & 47.7 {\tiny ($\pm$ 3.2)} & 89.4 {\tiny ($\pm$ 2.1)} & 35.4 {\tiny ($\pm$ 2.0)} & \texttt{OUT-OF-MEM} & 75.6 {\tiny ($\pm$ 1.6)} & 9.1 \\ 
& CORE-WL-VH & 73.5 {\tiny ($\pm$ 6.1)} & 51.7 {\tiny ($\pm$ 4.1)} & 73.0 {\tiny ($\pm$ 4.5)} & 51.1 {\tiny ($\pm$ 1.6)} & 40.2 {\tiny ($\pm$ 1.8)} & 84.5 {\tiny ($\pm$ 2.0)} & 4.5 \\ 
& CORE-SP & 68.5 {\tiny ($\pm$ 3.9)} & 51.0 {\tiny ($\pm$ 3.5)} & 91.0 {\tiny ($\pm$ 1.8)} & \texttt{TIMEOUT} & \texttt{OUT-OF-MEM} & \texttt{TIMEOUT} & 12.8 \\ 
\midrule
\parbox[t]{2mm}{\multirow{4}{*}{\rotatebox[origin=c]{90}{GNNs}}} & DGCNN & 69.2 {\tiny ($\pm$ 3.0)} & 45.6 {\tiny ($\pm$ 3.4)} & 87.8 {\tiny ($\pm$ 2.5)} & 49.2 {\tiny ($\pm$ 1.2)} & 43.9 {\tiny ($\pm$ 1.0)} & 71.2 {\tiny ($\pm$ 1.9)} & 7.9 \\ 
& GraphSAGE & 68.8 {\tiny ($\pm$ 4.5)} & 47.6 {\tiny ($\pm$ 3.5)} & 84.3 {\tiny ($\pm$ 1.9)} & 50.0 {\tiny ($\pm$ 1.3)} & 43.5 {\tiny ($\pm$ 1.0)} & 73.9 {\tiny ($\pm$ 1.7)} & 7.3 \\ 
& DiffPool & 68.4 {\tiny ($\pm$ 3.3)} & 45.6 {\tiny ($\pm$ 3.4)} & 89.1 {\tiny ($\pm$ 1.6)} & 53.8 {\tiny ($\pm$ 1.4)} & 44.4 {\tiny ($\pm$ 1.4)} & 68.9 {\tiny ($\pm$ 2.0)} & 7.2 \\ 
& GIN & 71.2 {\tiny ($\pm$ 3.9)} & 48.5 {\tiny ($\pm$ 3.3)} & 89.9 {\tiny ($\pm$ 1.9)} & 56.1 {\tiny ($\pm$ 1.7)} & 48.3 {\tiny ($\pm$ 1.6)} & 75.6 {\tiny ($\pm$ 2.3)} & 3.6 \\ 
\bottomrule
\end{tabular}
}
\caption{Average classification accuracy ($\pm$ standard deviation) on the $6$ classification datasets containing unlabeled graphs. The ``Avg. Rank'' column illustrates the average rank of each kernel/GNN. The lower the average rank, the better the overall performance of the kernel/GNN.}
\label{tab:results_unlabeled}
\end{table}

\begin{table}[t]
\centering
\scriptsize
\def\arraystretch{1.05}
\resizebox{\textwidth}{!} {
\begin{tabular}{lrrrrrrc} \toprule
\multirow{3}{*}{Kernels} & \multicolumn{6}{c}{DATASETS} & \multirow{2}{*}{Avg.} \\ \cline{2-7}
& IMDB & IMDB & REDDIT & REDDIT & REDDIT & \multirow{2}{*}{COLLAB} & \multirow{2}{*}{Rank}\\
& BINARY & MULTI & BINARY & MULTI-5K & MULTI-12K & & \\ 
\midrule
VH & 0.07s & 0.15s & 0.67s & 2.2s & 6.37s & 0.24s & 1.0 \\ 
RW & 10m 26.54s & 12m 8.7s & \texttt{TIMEOUT} & \texttt{TIMEOUT} & \texttt{TIMEOUT} & 9h 41m 21.24s & 15.6 \\ 
SP & 11.51s & 7.92s & 4h 48m 11.19s & 12h 40m 19.5s & \texttt{TIMEOUT} & 24m 14.94s & 8.3 \\ 
GR & 30m 6.66s & 16m 13.44s & 24m 32.66s & 1h 13m 7.1s & 30m 57.41s & 49m 22.03s & 10.8 \\ 
WL-VH & 12.21s & 12.0s & 13m 27.98s & 6m 2.02s & 5m 11.78s & 24m 41.07s & 4.7 \\ 
WL-SP & 1m 21.33s & 1m 24.83s & \texttt{TIMEOUT} & \texttt{TIMEOUT} & \texttt{TIMEOUT} & 2h 53m 9.92s & 13.2 \\ 
WL-PM & 2h 3m 26.51s & 2h 25m 58.31s & \texttt{OUT-OF-MEM} & \texttt{OUT-OF-MEM} & \texttt{OUT-OF-MEM} & \texttt{OUT-OF-MEM} & 17.1 \\ 
WL-OA & 23.84s & 40.96s & 1h 37m 38.26s & 15h 58m 53.54s & \texttt{TIMEOUT} & 7h 2m 16.42s & 10.7 \\ 
NH & 25.08s & 33.73s & 20m 7.15s & 3h 29m 27.55s & 11h 44m 40.96s & 15m 28.61s & 7.5 \\ 
NSPDK & 2m 10.12s & 3m 8.3s & \texttt{TIMEOUT} & \texttt{TIMEOUT} & \texttt{TIMEOUT} & \texttt{TIMEOUT} & 15.2 \\ 
Lo-$\vartheta$  & 6h 4m 8.55s & 5h 35m 19.86s & \texttt{TIMEOUT} & \texttt{TIMEOUT} & \texttt{TIMEOUT} & \texttt{TIMEOUT} & 17.4 \\ 
SVM-$\vartheta$ & 30.37s & 51.3s & 19m 29.55s & 24m 40.57s & 47m 39.6s & 1m 41.97s & 6.3 \\ 
ODD-STh & 3.94s & 4.55s & 1m 53.5s & 4m 48.92s & 8m 20.66s & 26m 9.55s & 3.3 \\ 
PM & 1m 31.37s & 3m 1.25s & 10m 12.88s & 51m 45.1s & 3h 50m 38.6s & 10m 22.45s & 8.0 \\ 
GH & 2m 11.15s & 2m 3.71s & \texttt{TIMEOUT} & \texttt{TIMEOUT} & \texttt{TIMEOUT} & 2h 19m 30.0s & 13.8 \\ 
SM & \texttt{TIMEOUT} & \texttt{TIMEOUT} & \texttt{OUT-OF-MEM} & \texttt{OUT-OF-MEM} & \texttt{OUT-OF-MEM} & \texttt{TIMEOUT} & -- \\
PK & 7.09s & 13.43s & 1m 26.03s & 5m 52.84s & 20m 22.64s & 1m 11.76s & 3.3 \\ 
ML & 1h 40m 28.88s & 1h 50m 36.11s & 8h 21m 18.76s & 47m 21.37s & \texttt{OUT-OF-MEM} & 4h 28m 12.65s & 13.3 \\ 
CORE-WL-VH & 55.99s & 1m 13.89s & 9m 52.79s & 25m 1.53s & 17m 37.71s & 5h 4m 10.52s & 7.5 \\ 
CORE-SP & 3m 58.29s & 4m 29.55s & 10h 37m 3.94s & \texttt{TIMEOUT} & \texttt{OUT-OF-MEM} & \texttt{TIMEOUT} & 15.1 \\
\bottomrule
\end{tabular}
}
\caption{Average CPU running time for kernel matrix computation on the $6$ classification datasets containing unlabeled graphs. The ``Avg. Rank'' column illustrates the average rank of each kernel. The lower the average rank, the lower the overall running time of the kernel.}
\label{tab:runtimes_unlabeled}
\end{table}

Tables~\ref{tab:results_unlabeled} and~\ref{tab:runtimes_unlabeled} illustrate average prediction accuracies and average running times of the compared kernels and GNNs on the $6$ datasets that contain unlabeled graphs.
We observe that the GIN model is the best-performing method.
It outperforms all the other methods on $2$ out of the $6$ datasets (REDDIT-MULTI-5K and REDDIT-MULTI-12K).
The CORE-WL-VH kernel achieves the second best performance.
Indeed, the CORE-WL-VH kernel outperforms all the other approaches on $3$ out of the $6$ datasets (IMDB-BINARY, IMDB-MULTI and COLLAB).
The WL-OA, NH and WL-VH kernels also yield high performance on most datasets.
In fact, these kernels along with CORE-WL-VH outperform the remaining $3$ GNN models, that is DGCNN, GraphSAGE and DiffPool.
Furthermore, on the remaining datasets, it reached the second, the second and the eighth best accuracy levels among all methods considered.
We should note that the core framework improved significantly the performance of the SP kernel on several datasets, while CORE-SP achieved the highest average accuracy on the REDDIT-BINARY dataset.
The Lo-$\vartheta$ kernel was the worst-performing kernel, followed by VH, WL-SP, VH, NSPDK and GH in that order.
It is interesting to mention that most of the kernels that reached the highest accuracies can also handle graphs with dicrete node labels.
For those kernels, the label of each vertex was set equal to its degree.
The kernels that can handle only unlabeled graphs (GR, Lo-$\vartheta$, SVM-$\vartheta$) failed to achieve accuracies competitive to the best-performing kernels.
With regards to the kernels that can handle graphs with continuous attributes (GH, SM, PK, and ML), as mentioned above, ML yielded the best results.
GH and PK achieved low accuracy levels, while SM failed to generate even a single kernel matrix due to running time or memory issues.
In this set of experiments, the $4$ GNN models provided very good performance results.
Before applying the GNNs to a dataset, the vertices of all graphs were annotated with a single feature that was set equal to the degree of the vertex.
As already mentioned, GIN is the best-performing method.
The remaining $3$ GNNs yielded similar performance to each other, but were still outperformed by a few graph kernels.

On most datasets, the variability in the performance of the different kernels is low.
The kernels achieve higher performance on binary classification tasks (IMDB-BINARY and REDDIT-BINARY) than on multi-class classification tasks.
For instance, on IMDB-MULTI, REDDIT-MULTI-5K and REDDIT-MULTI-12K, the highest average accuracies obtained by the considered approaches are $51.7\%$, $56.1\%$ and $48.3\%$, respectively.
Hence, it is clear that these three datasets are very challenging even for state-of-the-art methods.

In terms of running time, similar to the labeled case, VH is again the fastest kernel on all datasets.
The running time of PK, ODD-STh, and WL-VH is also low compared to the other kernels on most datasets.
The SVM-$\vartheta$, NH, PM, SP and CORE-WL kernels were also competitive in terms of running time.
Besides achieving low accuracy levels, the Lo-$\vartheta$ kernel is also very computationally expensive.
The WL-PM, RW, NSPDK, CORE-SP, WL-SP and GH are also very expensive in terms of running time.
The above $7$ kernels did not manage to calculate any kernel matrix on REDDIT-MULTI-5K and REDDIT-MULTI-12K within one day.
It should be mentioned that these two datasets contain several thousands of graphs, while the size of the graphs is also large (\ie several hundreds of vertices on average).
The SM kernel failed to compute the kernel matrix on IMDB-BINARY, IMDB-MULTI and COLLAB within one day, while it exceeded the maximum available memory on the remaining three datasets.

When dealing with tasks that involve unlabeled graphs, we suggest to assign discrete node labels to the vertices of the graphs (\eg set the label of each vertex equal to its degree), and then to again employ kernels that utilize some neighborhood aggregation mechanism.
For instance, the CORE-WL-VH, WL-OA, NH, and WL-VH kernels achieve high accuracies, while their computational complexity is realtively low.
Altenatively, a GNN model could be employed, especially in the case of large datasets.
Note, however, that even though GIN was found to be the best-performing approach in this set of experiments, the highest performance on $4$ out of the $6$ datasets was achieved by some graph kernel and not by a GNN.
Therefore, kernels still seem to be well-suited for such kind of datasets.

\paragraph{Node-Attributed Graphs}

\begin{table}[t]
\centering
\scriptsize
\def\arraystretch{1.05}
\begin{tabular}{llcccccc} \toprule
& \multirow{3}{*}{Methods} & \multicolumn{5}{c}{DATASETS} & \multirow{2}{*}{Avg.} \\ \cline{3-7}
& & \multirow{2}{*}{ENZYMES} & \multirow{2}{*}{PROTEINS\_full} & \multirow{2}{*}{SYNTHETICnew} & \multirow{2}{*}{Synthie} & \multirow{2}{*}{BZR} & \multirow{2}{*}{Rank} \\ 
& & & & & \\ 
\midrule
\parbox[t]{2mm}{\multirow{5}{*}{\rotatebox[origin=c]{90}{Kernels}}} & SP & \texttt{TIMEOUT} & \texttt{TIMEOUT} & \texttt{TIMEOUT} & \texttt{TIMEOUT} & \texttt{TIMEOUT} & -- \\ 
& GH & 67.7 {\tiny ($\pm$ 6.5)} & 72.6 {\tiny ($\pm$ 1.9)} & 74.3 {\tiny ($\pm$ 5.6)} & 73.8 {\tiny ($\pm$ 7.3)} & 82.3 {\tiny ($\pm$ 7.2)} & 3.2 \\ 
& SM & \texttt{TIMEOUT} & \texttt{OUT-OF-MEM} & \texttt{TIMEOUT} & \texttt{TIMEOUT} & 79.5 {\tiny ($\pm$ 5.6)} & 8.2 \\ 
& PK & 21.5 {\tiny ($\pm$ 3.4)} & 59.6 {\tiny ($\pm$ 0.2)} & 47.7 {\tiny ($\pm$ 7.5)} & 46.2 {\tiny ($\pm$ 3.6)} & 78.8 {\tiny ($\pm$ 5.5)} & 7.1 \\ 
& ML & 33.2 {\tiny ($\pm$ 5.8)} & 71.1 {\tiny ($\pm$ 4.6)} & 47.7 {\tiny ($\pm$ 7.3)} & 49.0 {\tiny ($\pm$ 8.3)} & 81.3 {\tiny ($\pm$ 6.2)} & 5.9 \\
\midrule
\parbox[t]{2mm}{\multirow{4}{*}{\rotatebox[origin=c]{90}{GNNs}}} & DGCNN & 38.9 {\tiny ($\pm$ 5.7)} & 72.9 {\tiny ($\pm$ 3.5)} & 53.7 {\tiny ($\pm$ 3.1)} & 80.0 {\tiny ($\pm$ 3.4)} & 81.8 {\tiny ($\pm$ 4.4)} & 4.2 \\ 
& GraphSAGE & 58.2 {\tiny ($\pm$ 6.0)} & 73.0 {\tiny ($\pm$ 4.5)} & 88.0 {\tiny ($\pm$ 7.3)} & 51.3 {\tiny ($\pm$ 9.9)} & 81.2 {\tiny ($\pm$ 4.2)} & 3.8 \\ 
& DiffPool & 59.5 {\tiny ($\pm$ 5.6)} & 73.7 {\tiny ($\pm$ 3.5)} & 72.0 {\tiny ($\pm$ 6.7)} & 84.5 {\tiny ($\pm$ 3.9)} & 84.5 {\tiny ($\pm$ 4.2)} & 2.4 \\ 
& GIN & 59.6 {\tiny ($\pm$ 4.5)} & 73.3 {\tiny ($\pm$ 4.0)} & 80.5 {\tiny ($\pm$ 6.6)} & 89.7 {\tiny ($\pm$ 4.6)} & 85.4 {\tiny ($\pm$ 5.1)} & 1.6 \\
\bottomrule
\end{tabular}
\caption{Average classification accuracy ($\pm$ standard deviation) on the $5$ classification datasets containing node-attributed graphs. The ``Avg. Rank'' column illustrates the average rank of each kernel/GNN. The lower the average rank, the better the overall performance of the kernel/GNN.}
\label{tab:results_attributed}
\end{table}

\begin{table}[t]
\centering
\scriptsize
\def\arraystretch{1.1}
\begin{tabular}{lrrrrrc} \toprule
\multirow{3}{*}{Kernels} & \multicolumn{5}{c}{DATASETS} & \multirow{2}{*}{Avg.} \\ \cline{2-6}
& \multicolumn{1}{c}{\multirow{2}{*}{ENZYMES}} & \multicolumn{1}{c}{\multirow{2}{*}{PROTEINS\_full}} & \multicolumn{1}{c}{\multirow{2}{*}{SYNTHETICnew}} & \multicolumn{1}{c}{\multirow{2}{*}{Synthie}} & \multicolumn{1}{c}{\multirow{2}{*}{BZR}} & \multirow{2}{*}{Rank} \\ 
& & & & & \\ 
\midrule
SP & \texttt{TIMEOUT} & \texttt{TIMEOUT} & \texttt{TIMEOUT} & \texttt{TIMEOUT} & \texttt{TIMEOUT} & -- \\
GH & 16m 36.48s & 3h 15m 5.36s & 12m 37.05s & 17m 44.28s & 4m 11.74s & 2.2 \\ 
SM & \texttt{TIMEOUT} & \texttt{OUT-OF-MEM} & \texttt{TIMEOUT} & \texttt{TIMEOUT} & 6h 15m 59.36s & 4.0 \\ 
PK & 14.93s & 1m 10.25s & 14.43s & 11.37s & 7.24s & 1.0 \\ 
ML & 1h 4m 52.54s & 2h 48m 24.59s & 2h 49m 35.19s & 1h 44m 10.48s & 47m 44.62s & 2.8 \\ 
\bottomrule
\end{tabular}
\caption{Average CPU running time for kernel matrix computation on the $5$ classification datasets containing node-attributed graphs. The ``Avg. Rank'' column illustrates the average rank of each kernel. The lower the average rank, the lower the overall running time of the kernel.}
\label{tab:runtimes_attributed}
\end{table}

As mentioned above, the majority of graph kernels can handle graphs that are either unlabeled or contain discrete node labels.
On the other hand, the number of graph kernels that can handle graphs that contain continuous vertex attributes is limited.
Moreover, most of these kernels do not scale even to relatively small datasets.
Tables~\ref{tab:results_attributed} and~\ref{tab:runtimes_attributed} illustrate average prediction accuracies and average runtimes of graph kernels and GNNs on datasets that contain node-attributed graphs.
Note that although the graphs of some of these datasets contain discrete node labels, we did not take these discrete labels into account since our main aim was to evaluate the ability of the kernels to properly handle continuous node attributes.
GIN is the best-performing approach also in this set of experiments, while GNNs generally outperform graph kernels.
GH is the best-performing kernel since it outperforms all the other kernels on all datasets.
Furthermore, GH outperforms $2$ out of the $4$ GNNs (DGCNN and GraphSAGE).
GH is followed by ML and PK in terms of performance in that order.
One of the most striking findings of this set of experiments is that the SP kernel did not manage to compute the kernel matrix even on a single dataset within one day, while the SM kernel finished its computations within one day only on the BZR dataset on which it was outperformed by GH and ML.
The $4$ GNNs yielded in most cases high levels of accuracy.
However, some GNNs failed to produce competitive results on some datasets.
For instance, DGCNN achieved an average accuracy of $53.7\%$ on SYNTHETICnew, while GraphSAGE yielded an average accuracy of $51.3\%$ on Synthie.
This might be due to the neighborhood aggregation mechanisms or readout functions employed by these models.

In terms of running time, PK is the most efficient kernel since it handled all datasets in less than two minutes.
GH and ML are much slower than PK on all datasets.
For instance, the average computation time of ML and GH was greater than $2$ hours and $3$ hours on PROTEINS\_full, respectively.
The SP and SM kernels, as already discussed, are very expensive in terms of running time, and hence, their usefulness in real-world problems is limited.

To sum up, it is clear that the running time of most kernels for node-attributed graphs is prohibitive, especially considering the relatively small size of the datasets.
Although the running time of PK is attractive, it achieved low accuracies on almost all datasets.
An open challenge in the field of graph kernels is thus to develop scalable kernels for graphs with continuous vertex attributes.
On the other hand, GNNs can naturally handle continuous node features, while they have also outperformed graph kernels in the experimental evaluation.
Therefore, when dealing with graphs whose vertices are annotated with continuous attributes, we recommend using a GNN model instead of a graph kernel.

\subsection{Expessiveness of Graph Kernels}
Over the past years, the expessive power of graph kernels was assessed almost exclusively from empirical studies.
So far, there are only a few theoretical findings related to the expressiveness of graph kernels.
For instance, as already mentioned, it has been shown that the mapping induced by kernels that are computable in polynomial time is not injective (and thus these kernels cannot solve the graph isomorphism problem) \shortcite{gartner2003graph}.
Recently, \shortciteA{kriege2018property} proposed a framework to measure the expressiveness of graph kernels based on ideas from property testing, and showed that some well-established graph kernels such as the shortest path kernel, the random walk kernel, and the Weisfeiler-Lehman subtree kernel cannot identify fundamental graph properties such as triangle-freeness and bipartitness.
It is thus clear that there are several interesting questions about the expressiveness of graph kernels which are far from being answered.
An example of such a question is whether a specific graph kernel captures graph similarity better than others for some specific application.

In what follows, we conduct an experiment to empirically answer the above question.
Specifically, we build a dataset that contains instances of different families of graphs.
Then, we compare the similarities (\ie kernel values) produced by graph kernels against those generated by an intractable graph similarity function which we consider to be an oracle function that outputs the true similarity between graphs.
Formally, for any two graphs $G_1 = (V_1, E_1)$ and $G_2 = (V_2, E_2)$ on $n$ vertices with respective $n \times n$ adjacency matrices $A_1$ and $A_2$, we define a function $f : \mathcal{G} \times \mathcal{G} \rightarrow \mathbb{R}$ where $\mathcal{G}$ is the space of graphs which quantifies the similarity of $G_1$ and $G_2$.
The function can be expressed as the following maximization problem:
\begin{equation}
    f(G_1, G_2) = \max_{P \in \Pi} \frac{\sum_{i=1}^n \sum_{j=1}^n \left[ A_1 \odot P A_2 P^\top \right]_{ij}}{||A_1||_F \, ||A_2||_F}
    \label{eq:sim_function}
\end{equation}
where $\Pi$ denotes the set of $n \times n$ permutation matrices, $\odot$ denotes the elementwise product, and $||\cdot||_F$ is the Froebenius matrix norm.
For clarity of presentation we assume $n$ to be fixed (\ie both graphs consist of $n$ vertices).
In order to apply the function to graphs of different cardinality, one can append zero rows and columns to the adjacency matrix of the smaller graph to make its number of rows and columns equal to $n$.
Therefore, the problem of graph comparison can be reformulated as the problem of maximizing the above function over the set of permutation matrices.
A permutation matrix $P$ gives rise to a bijection $\pi : V_1 \rightarrow V_2$.
The function defined above seeks for a bijection such that the number of common edges $|\{ (u,v) \in E_1 : \big(\pi(u),\pi(v)\big) \in E_2 \}|$ is maximized.
Then, the number of common edges is normalized such that it takes values between $0$ and $1$.
Observe that the above definition is symmetric in $G_1$ and $G_2$.
The two graphs are isomorphic to each other if and only if there exists a permutation matrix $P$ for which the above function is equal to $1$.
Therefore, a value equal to $0$ denotes maximal dissimilarity, while a value equal to $1$ denotes that the two graphs are isomorphic to each other.
Note that if the compared graphs are not empty (\ie they contain at least one edge), the function will take some value greater than $0$.
Solving the above optimization problem for large graphs is clearly intractable since there are $n!$ permutation matrices of size $n$.
In fact, the above function is related to the well-studied Frobenius distance between graphs which is known to be an NP-complete problem \shortcite{grohe2018graph}.

\subsubsection{Dataset}
\begin{figure}[t]
    \begin{minipage}{0.4\linewidth}
        \centering
        \begin{tabular}{l|c} \hline
            \multicolumn{2}{c}{Synthetic Dataset} \\ \hline\hline
            Max \# vertices & $9$ \\
            Min \# vertices & $2$ \\
            Average \# vertices & $7.29$ \\ \hline
            Max \# edges & $36$ \\
            Min \# edges & $1$ \\
            Average \# edges & $11.34$ \\ \hline
            \# graphs & $191$ \\ \hline
        \end{tabular}
        \captionof{table}{Summary of the synthetic dataset that we used in our experiment.}
        \label{tab:statistics_synthetic}
    \end{minipage}
    \hfill
    \begin{minipage}{0.5\linewidth}
        \centering
        \includegraphics[width=0.8\textwidth]{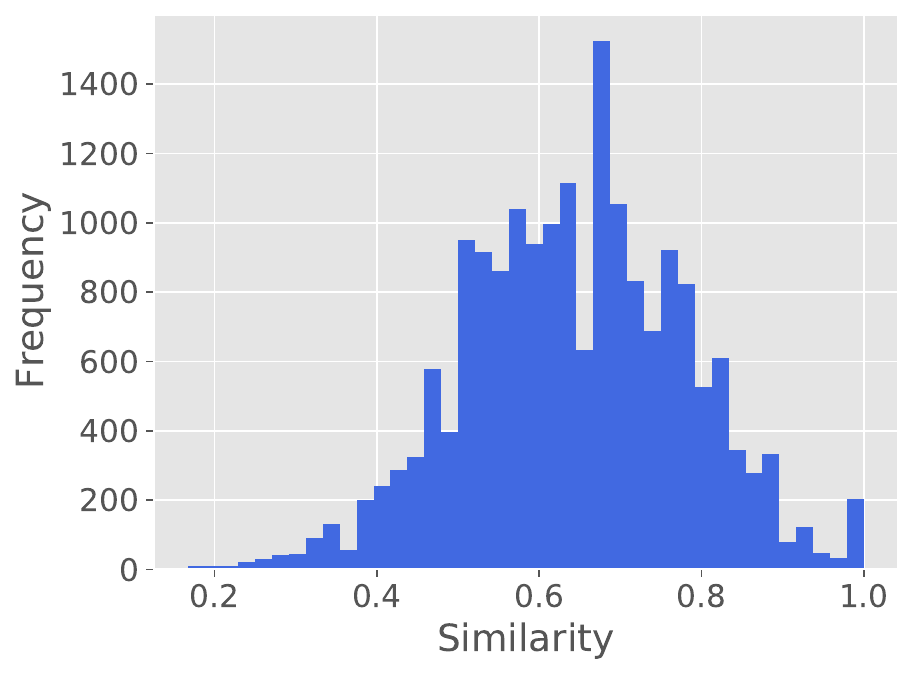}
        \captionof{figure}{Distribution of similarities between the synthetic graphs.}
        \label{fig:sim_distribution}
    \end{minipage}
    \vspace{-.6cm}
\end{figure}

Since the function defined in Equation~\eqref{eq:sim_function} is intractable for large graphs, we generated graphs consisting of at most $9$ vertices.
Furthermore, each graph is connected and contains at least $1$ edge.
We generated $191$ pairwise non-isomorphic graphs.
The dataset consists of different types of synthetic graphs.
These include simple structures such as cycle graphs, path graphs, grid graphs, complete graphs and star graphs, but also randomly-generated graphs such as Erd{\H{o}}s-R{\'e}nyi graphs, Barab{\'a}si-Albert graphs and Watts-Strogatz graphs.
Table~\ref{tab:statistics_synthetic} shows statistics of the synthetic dataset that we used in our experiments.
Figure~\ref{fig:sim_distribution} illustrates the distribution of the similarities of the generated graphs as computed by the proposed measure.
There are $\nicefrac{191*192}{2} = 18,336$ pairs of graphs in total (including pairs consisting of a graph and itself).
Interestingly, most of the similarities take values between $0.5$ and $0.8$.

\subsubsection{Experimental Settings}
The set of kernels for this experiment contains the same $20$ kernels that were evaluated in the context of the graph classification experiments.
Once again, we use the implementations of the kernels contained in the GraKeL library \shortcite{siglidis2020grakel}.
Note that the synthetic graphs are unlabeled.
Therefore, for kernels that assume node-labeled graphs, all vertices of all graphs are annotated with a single label, while for kernels that assume node-attributed graphs, all the vertices of all graphs are assigned the same attribute.

As discussed above, the function defined in Equation~\eqref{eq:sim_function} gives an output in the range $[0, 1]$.
We normalize the obtained kernel values as follows such that they also take values in the range $[0,1]$: $\hat{k}(G_i, G_j) = \nicefrac{k(G_i,G_j)}{\sqrt{k(G_i, G_i) \, k(G_j, G_j)}}$ for any graphs $G_i, G_j$.
We should stress that the normalized kernel value can take a value equal to $1$ even if the compared graphs are mapped to different representations.
Indeed, if the angle between the vector representations of two graphs is $0^\circ$, then their normalized kernel value is equal to $1$.
To avoid such a scenario, we could define a distance function between graphs and accordingly compute the Euclidean distance between the graph representations generated by the different approaches.
However, it turns out that most widely-used learning algorithms compute the inner products between the input objects or some transformations of these objects. 
In fact, when learning with kernels, we usually normalize the kernel matrices using the equation defined above before feeding to a kernel method such as the SVM classifier.
Therefore, we believe that evaluating the ``similarity'' of the obtained representations is more natural than evaluating their ``distance''.
With regards to the values of the hyperparameters of the $20$ kernels, we experiment with the same values as in the case of graph classification.
Specifically, we choose the hyperparameter values that result into the highest correlation between the kernel values generated by a given kernel and the similarities produced by the function of Equation~\eqref{eq:sim_function}.

To assess how well the different approaches approximate the similarity function, we employed two evaluation metrics: the Pearson correlation coefficient and the mean squared error (MSE).
In our setting, a high value of correlation would mean that the approach under consideration captures the relationships between the similarities (\eg whether the similarity of a pair of graphs is greater or lower than that of another pair).
On the other hand, a very small value of MSE denotes that the derived similarities are very close to those produced by the function defined in Equation~\eqref{eq:sim_function}.
A credible graph representation learning/similarity approach would yield both a high correlation and a small MSE.
The former would ensure that similar/dissimilar graphs are indeed deemed similar/dissimilar by the considered approach, while the latter would verify that the similarity values are on par with those produced by the similarity function of Equation~\eqref{eq:sim_function}.

\begin{figure}[t]
    \centering
    \subfloat
    {\includegraphics[width=\linewidth]{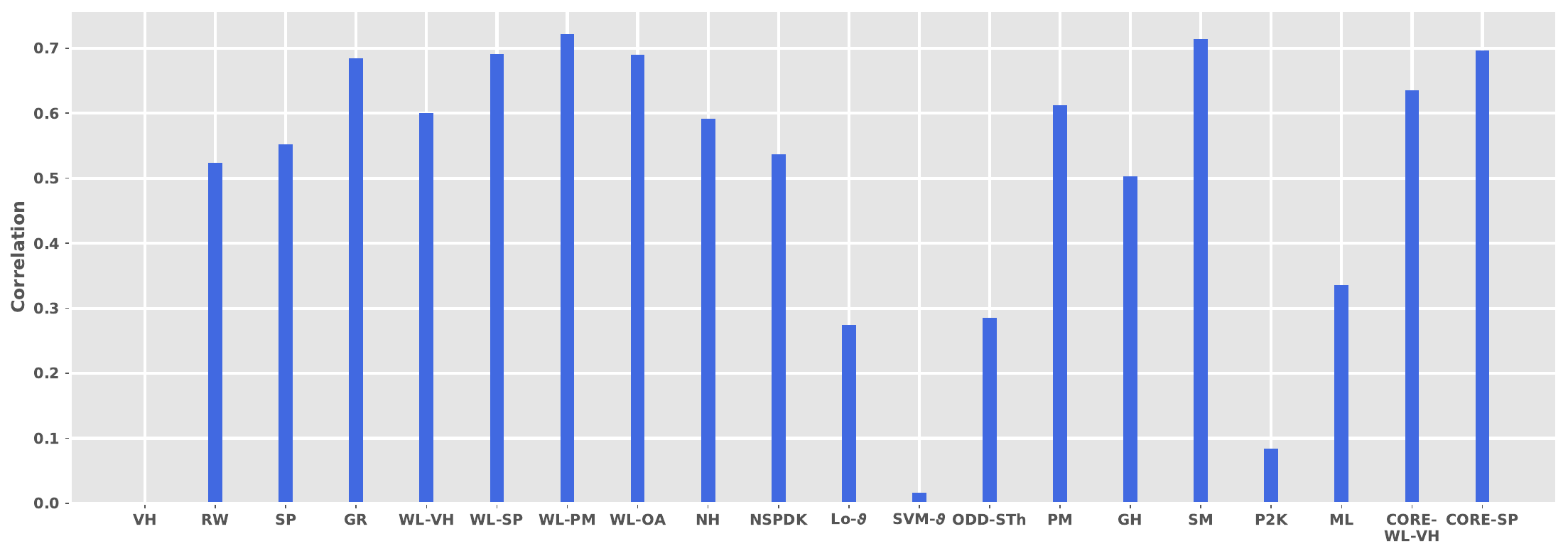}}\\
    \subfloat
    {\includegraphics[width=\linewidth]{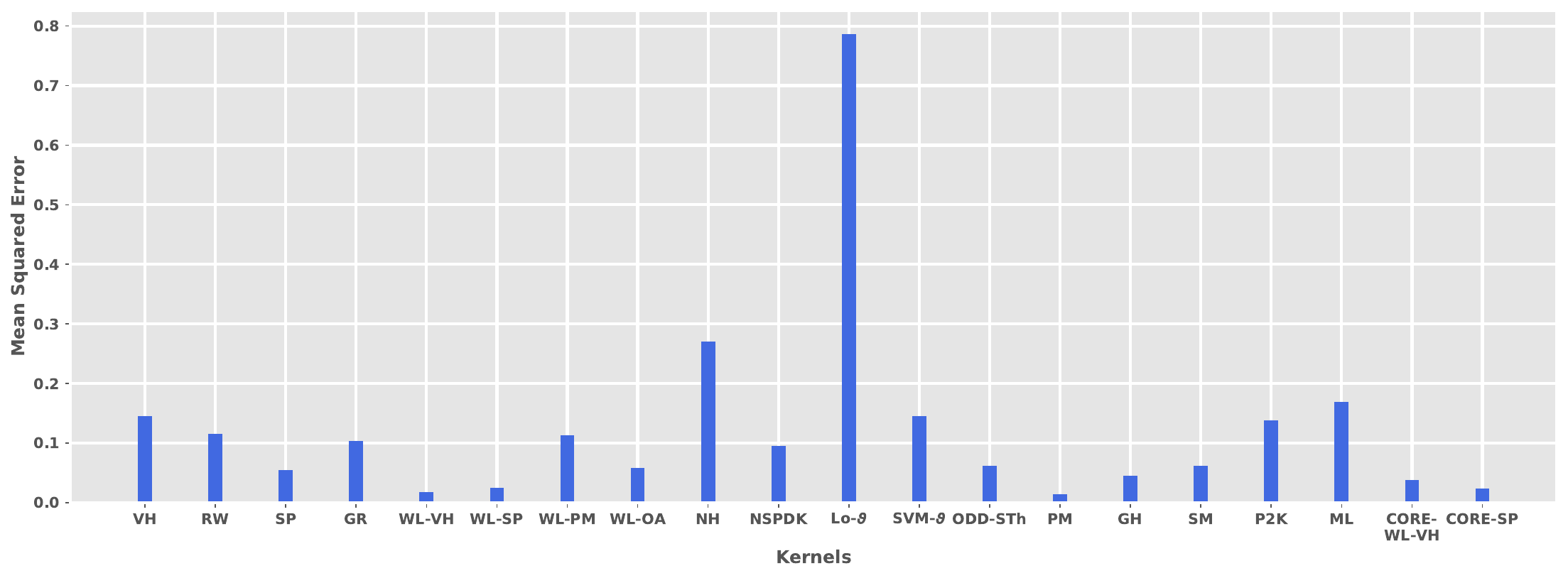}}
    \caption{Correlation and mean squared error between the kernel values produced by the $20$ kernels and the similarities produced by the function defined in Equation~\eqref{eq:sim_function}.}
    \label{fig:cor_mse}
\end{figure}

\begin{figure}[t]
    \centering
    \includegraphics[width=.95\textwidth]{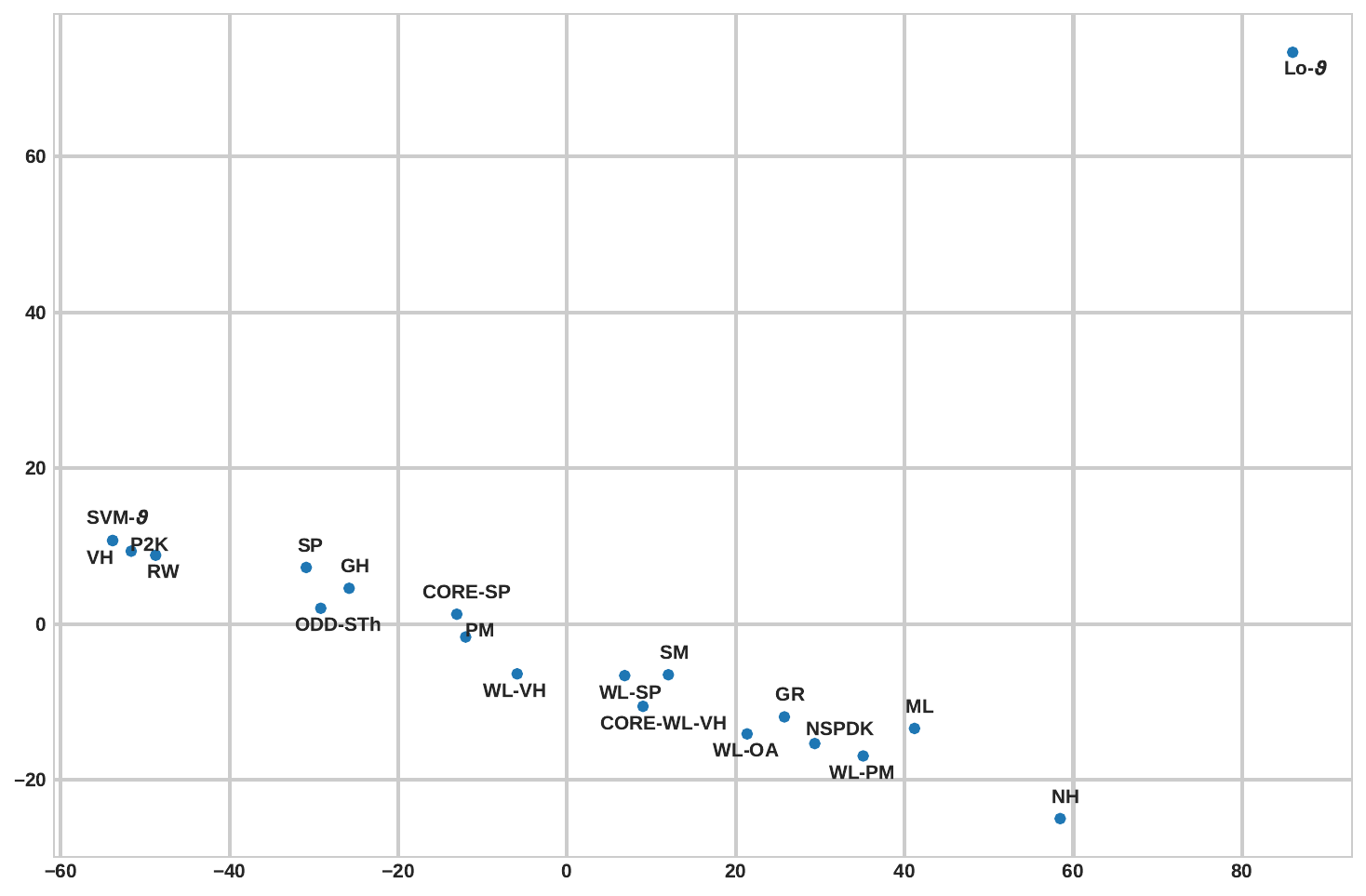}
    \caption{Projection of the representations of the $20$ kernels in $\mathbb{R}^2$ (using PCA). Each kernel is represented as a vector of kernel values between the graphs of the synthetic dataset.}
    \label{fig:pca_projection}
\end{figure}

\subsubsection{Results}
Figure~\ref{fig:cor_mse} illustrates the correlation and MSE between the kernel values produced by the $20$ kernels and the similarities produced by the function of Equation~\eqref{eq:sim_function}.
In terms of correlation, WL-PM and SM are the best-performing approaches followed by CORE-SP, WL-SP, WL-OA and GR.
The correlation between the first two kernels and the function of Equation~\eqref{eq:sim_function} is greater than $0.7$, while the rest of the above kernels achieve a correlation slightly lower than $0.7$.
On the other hand, VH, SVM-$\vartheta$ and PK yield very low levels of correlation (smaller than $0.1$) followed by Lo-$\vartheta$, ODD-STh and ML.
Note that for unlabeled graphs, the normalized kernel values of the VH kernel are always equal to $1$, and thus correlation is not defined since the values produced by the constant function have a variance equal to zero.
Overall, the majority of correlations is greater than $0.5$ which demonstrates that most kernels indeed capture some notion of similarity between graphs.
In terms of MSE, the PM, WL-VH, WL-SP and CORE-SP kernels are the best-performing approaches.
Notably, these graph kernels achieve very low values of MSE which indicates that the produced kernel values are very close to the similarities that emerge from the function of Equation~\ref{eq:sim_function}.
It is interesting to mention that most kernels achieve an MSE smaller than $0.1$.
The Lo-$\vartheta$ kernel yields an MSE value much greater than those of the other kernels, while the NH, ML, VH, PK and SVM-$\vartheta$ kernels also fail to achieve low levels of MSE.
As already mentioned, most graph kernels are generally motivated by runtime considerations.
They are computable in polynomial time, which usually has an impact on their expressive power.
Our results indicate that even though kernels do not provide any guarantees on how well they can approximate the above function, empirically, they seem to capture several aspects of graph similarity to a large extent. 

We next study whether there exist groups of kernels that are more similar to each other than to other kernels.
To discover such groups of kernels, we utilize the kernel values produced by the different kernels.
When a kernel is applied to the synthetic dataset that was introduced above, $\nicefrac{191*192}{2} = 18,336$ kernel values are computed in total.
We thus represent each kernel as a vector in a common space (of dimension $18,336$) based on the emerging kernel values.
We then project the representations of the kernels to the $2$-dimensional space using PCA.
The results are shown in Figure~\ref{fig:pca_projection}.
The position of each dot represents a projection of the kernel values generated by a single kernel.
We observe that VH, RW, PK and SVM-$\vartheta$ form a cluster, while NH and Lo-$\vartheta$ are isolated and are thus far from the other kernels.
All the remaining kernels are close to each other in the low dimensional space, but they do not form well-defined clusters.
Interestingly, the SP and GH kernels which both extract shortest paths are very close to each other, while all the Weisfeiler-Lehman kernels (\ie WL-VH, WL-SP, CORE-WL-VH, WL-OA and WL-PM) are also relatively close to each other in the $2$-dimensional space.

\section{Conclusion}\label{sec:conclusion}
Recent years have witnessed a tremendous increase in the availability of graph-structured data.
Graphs arise in many different contexts where it is necessary to represent relationships between entities.
Specifically, graphs are the commonly employed structure for representing data in various domains including bioinformatics, chemoinformatics, social networks and information networks.
The abundance of graph-structured data and the need to perform machine learning tasks on this kind of data led to the development of several sophisticated approaches such as graph kernels.
In this survey, we provided a detailed overview of graph kernels.
Furthermore, we empirically evaluated the effectiveness of several graph kernels, and measured their running time.
We hope that this survey will provide a better understanding of the current progress on graph kernels and graph classification, and offer some guidelines on how to apply these approaches in order to solve real-world problems.

Although graph kernels have achieved remarkable results in many tasks, there are still some challenges to be addressed, while there is also still some room for improvement.
For example, the majority of the kernels that can handle graphs with continuous attributes are either very expensive in terms of computational complexity or fail to produce competitive results.
Hence, we believe that an important direction of research is the development of scalable graph kernels for graphs annotated with continuous attributes which will also provide improvements over the state-of-the-art approaches.
Another useful direction of research is to capitalize on the framework for designing valid assignment kernels presented above, and to develop new kernels which compute an optimal assignment between substructures extracted from graphs.
In general, the complexity of the assignment kernels is more attractive than that of kernels that belong to the $R$-convolution framework, and hence, it is our belief that this framework can pave the way for the development of more efficient graph kernels.

\vskip 0.2in
\bibliographystyle{theapa}
\bibliography{sample}

\end{document}